\documentclass{article}

\usepackage{newtxtext}
\usepackage[left=1.25in, top=1in, bottom=1in, right=1.25in]{geometry}
\usepackage[authoryear]{natbib}
\usepackage{graphicx} % provides \includegraphics
\usepackage[utf8]{inputenc} % allow utf-8 input
\usepackage[T1]{fontenc}    % use 8-bit T1 fonts
\usepackage[colorlinks=true,citecolor=blue,linkcolor=blue,urlcolor=blue]{hyperref}
\usepackage{url}            % simple URL typesetting
\usepackage{booktabs}       % professional-quality tables
\usepackage{amsfonts}       % blackboard math symbols
\usepackage{nicefrac}       % compact symbols for 1/2, etc.
\usepackage{microtype}      % microtypography
\usepackage{xcolor}         % colors
\usepackage{amsmath}
\usepackage{tcolorbox}
\usepackage{cleveref}
\usepackage{wrapfig}
\usepackage{pifont}

\newcommand{\gradnorm}{g_{\mathrm{norm}}}
\usepackage{algorithm}
\usepackage{algorithmic}
\usepackage{enumitem}
\usepackage{wrapfig}    % for wrapped figures
\usepackage{lipsum}     % for dummy text; remove in your document

\title{Fantastic Pretraining Optimizers and \\ Where to Find Them}

\usepackage{float}

\author{
Kaiyue Wen \\
Stanford University\\
\texttt{kaiyuew@stanford.edu} 
\and
David Hall  \\
Stanford University\\
\texttt{dlwh@cs.stanford.edu} \\
\and
Tengyu Ma  \\
Stanford University\\
\texttt{tengyuma@stanford.edu} \\
\and
Percy Liang  \\
Stanford University\\
\texttt{pliang@cs.stanford.edu} \\
}

\begin{document}

\maketitle
\begin{abstract}
AdamW has long been the dominant optimizer in language model pretraining, despite numerous claims that alternative optimizers offer 1.4 to 2$\times$ speedup. We posit that two methodological shortcomings have obscured fair comparisons and hindered practical adoption: (i) unequal hyperparameter tuning and (ii) limited or misleading evaluation setups.
To address these two issues, we conduct a systematic study of ten deep learning optimizers across four model scales (0.1B–1.2B parameters) and data-to-model ratios (1--8$\times$ the Chinchilla optimum). We find that fair and informative comparisons require rigorous hyperparameter tuning and evaluations across a range of model scales and data-to-model ratios, performed at the end of training.
% \pl{at the end of training run - this means not using intermediate checkpoints? I'd either cut or make this clearer} \kaiyue{To discuss, perhaps too long: We find that fair and informative comparisons require rigorous hyperparameter tuning and evaluations across a range of model scales and data-to-model ratios, conducted at the end of the training runs rather than using intermediate checkpoints.}
% % \tnote{I think this sentence and the construct here suggests strongly that 
% "rigorous hyperparameter tuning is essential for fair comparisons" is a summary of the three points below, which doesn't seem to be true (or maybe it's true but it gives people the feeling that it only covers the first point). I think let's try to see if we can find a one-liner that covers the three points below? }\kaiyue{I see, this makes sense. Let me try to think whether there are some alternatives}\tnote{how about something plain and basic like "We found that fair and informative comparisons require proper hyperparameter tuning and comparing at a large scale and the end of the training trajectory" } \kaiyue{perhaps compares at multiple scaling regimes? The end of trajectory one feels slightly too detailed here.} \tnote{what multiple scale regimes mean? is it the same as at a large scale? I think the detail (compared at the end of the trajecotry) is fine because that's the only thing that matters in the third point?}
First, optimal hyperparameters for one optimizer may be suboptimal for another, making blind hyperparameter transfer unfair. 
% \tnote{A nitpicking--I felt that it's equally fine to say "..for one optimizer may be suboptimal for another"; anyways "far from optimal" can have many different interpretations on how far is far} \kaiyue{agree, will change}
Second, the actual speedup of many proposed optimizers over well-tuned baselines is lower than claimed and decreases with model size to only 1.1$\times$ for 1.2B parameter models. Thirdly, comparing intermediate checkpoints before reaching the target training budgets can be misleading, as rankings between two optimizers can flip during training due to learning rate decay.
Through our thorough investigation, we find that all the fastest optimizers such as Muon and Soap, use matrices as preconditioners --- multiplying gradients with matrices rather than entry-wise scalars. However, the speedup of matrix-based optimizers is inversely proportional to model scale, decreasing from 1.4$\times$ over AdamW for 0.1B parameter models to merely 1.1$\times$ for 1.2B parameter models.
\end{abstract}

\section{Introduction}

\begin{figure}[t]
    \begin{minipage}{0.48\textwidth}
        \centering
\includegraphics[width=\linewidth]{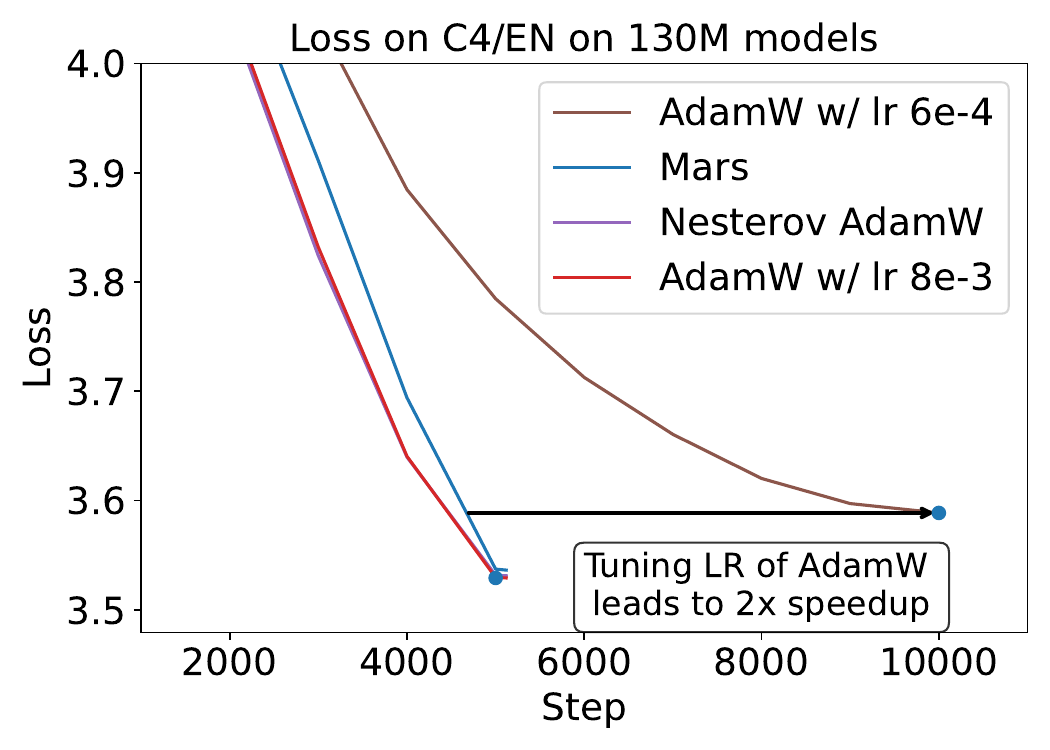}
    \end{minipage}    
    \begin{minipage}{0.46\textwidth}
        \centering \includegraphics[width=\linewidth]{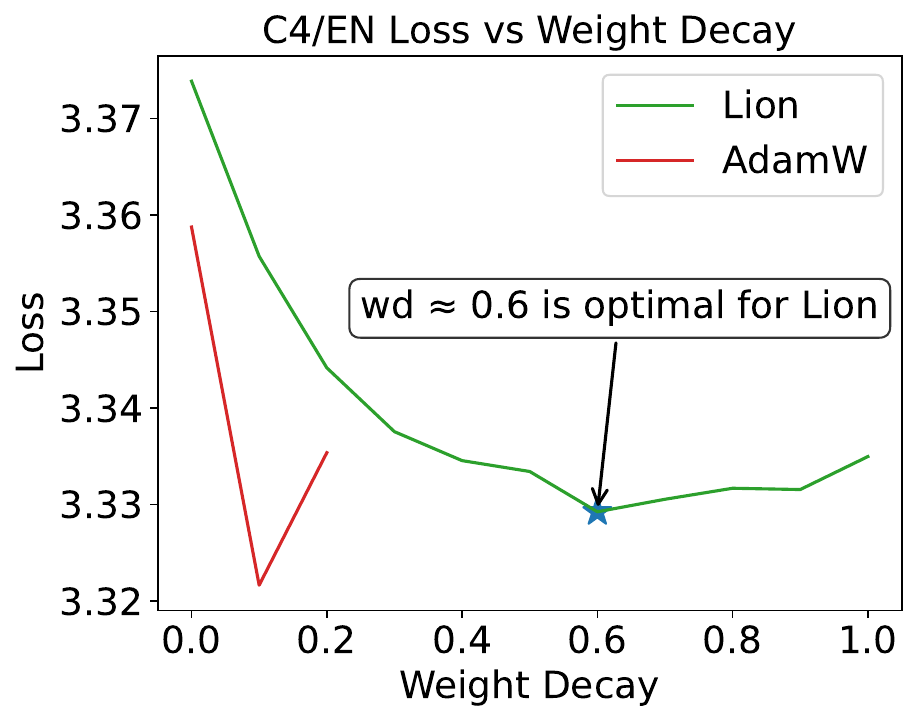}
    \end{minipage}
    \vfill
    \begin{minipage}
        {0.46\textwidth}
            \centering
    \includegraphics[width=\linewidth]{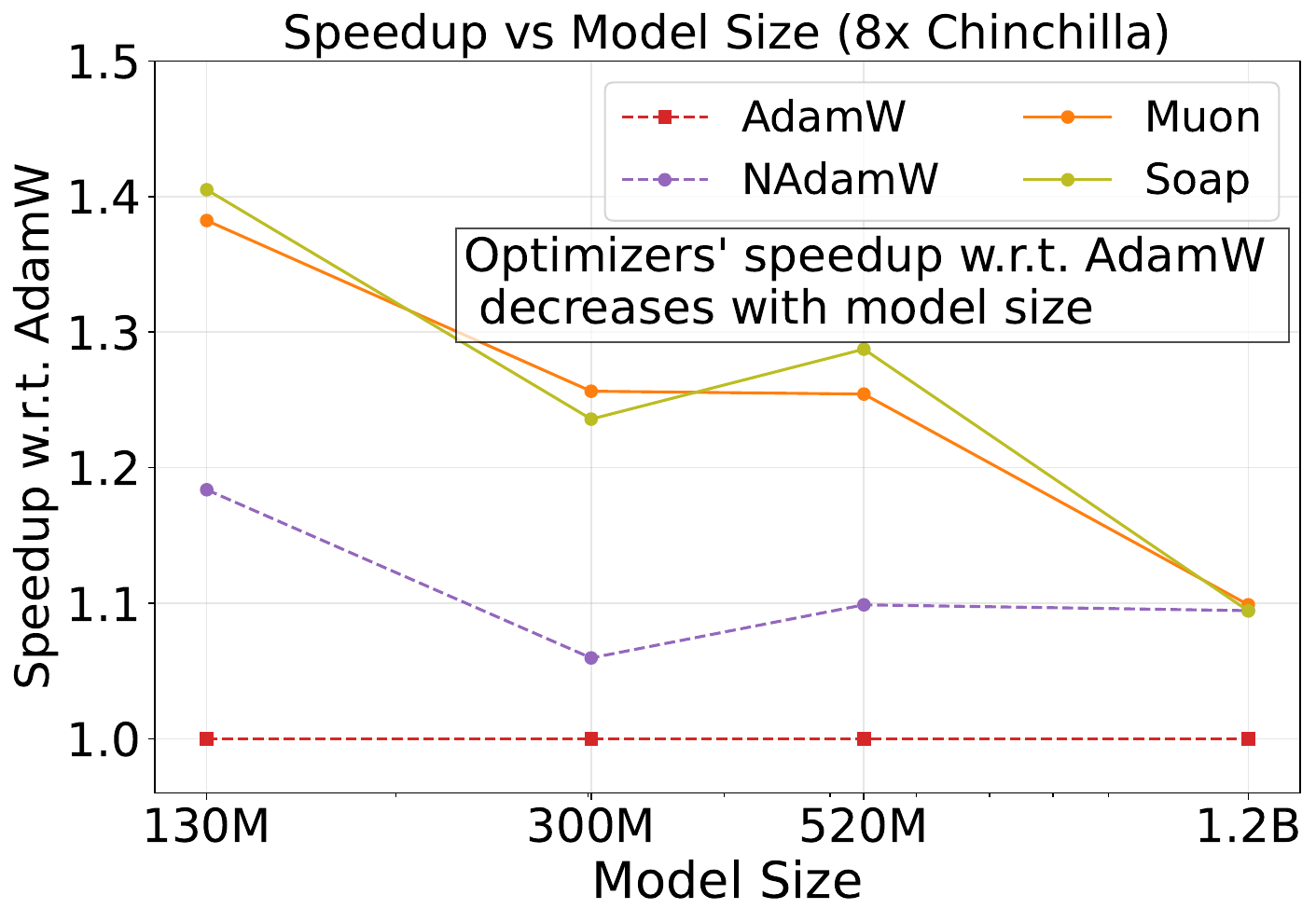}
        \end{minipage}
    \begin{minipage}
    {0.48\textwidth}
        \centering
\includegraphics[width=\linewidth]{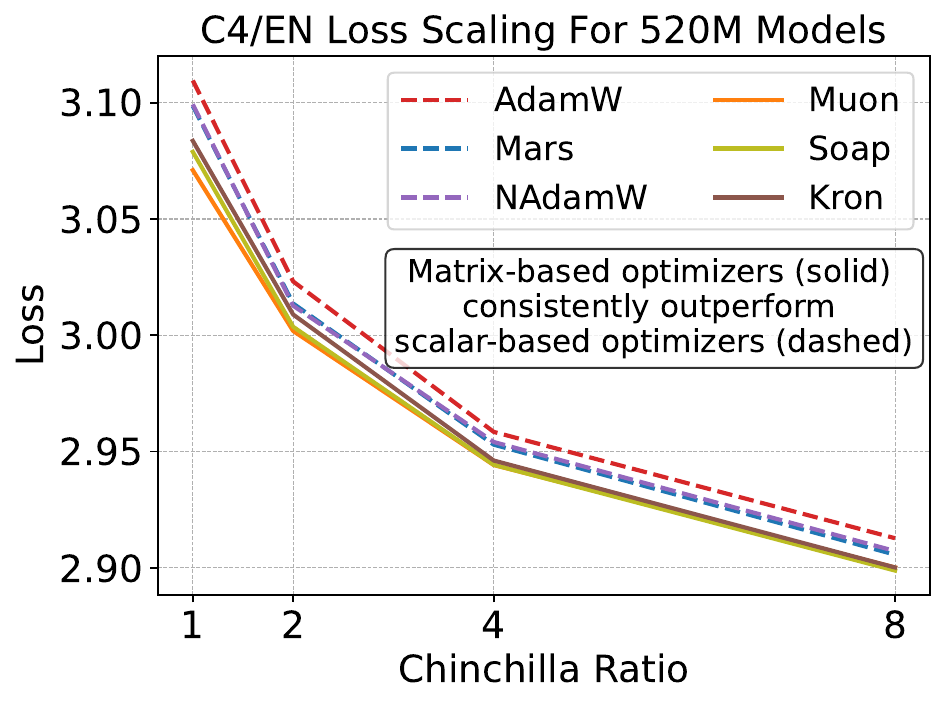}
    \end{minipage}
    \caption{
    \textbf{Top Left}: {
    The commonly used AdamW baseline for optimizer design is under-tuned.} Up to a 2$\times$ speedup is achievable by tuning a single hyperparameter (learning rate) in the GPT-3 recipe~\cite{brown2020languagemodelsfewshotlearners} for a 100M model (adopted in~\cite{liu2024sophiascalablestochasticsecondorder,wen2024understandingwarmupstabledecaylearningrates,yuan2025mars,liang2025cautious,wang2025sharpnessdisparityprincipletransformers}), highlighting the importance of proper hyperparameter optimization. 
    \textbf{Top Right}: {Fixing hyperparameters across optimizers does not guarantee fair comparison.} Shared hyperparameters such as learning rate and weight decay are commonly set to a constant in previous studies. However, even conceptually similar optimizers may correspond to very different optimal hyperparameters. 
    \textbf{Bottom Left}: {Speedup decays with model size.} While some optimizers show high (1.3-1.4$\times$) speedup over AdamW on models under 1B parameters, the speedup decays with model size to only 1.1$\times$ for 1.2B parameters.  
    \textbf{Bottom Right}: {Matrix-based optimizers consistently outperform scalar-based optimizers.}
    The loss curves for three scalar-based optimizers (AdamW, Nesterov AdamW, Mars) and three matrix-based optimizers (Kron, Soap, Muon) trained with different Chinchilla ratios of data are shown. Matrix-based optimizers achieve a consistent speedup over scalar-based optimizers. Furthermore, the three matrix-based optimizers converge to a similar loss in an overtrained setting.}
    \label{fig:motivation}

    \vspace{-0.2in}
\end{figure}

Pretraining has been the most computationally expensive component in the training pipeline for large language models, accounting for over 95\% of the cost in DeepSeek V3~\cite{deepseekai2025deepseekv3technicalreport}, and the additional RL training cost in DeepSeek R1~\cite{deepseekai2025deepseekr1incentivizingreasoningcapability} is also comparatively much smaller.
Until recently, AdamW has been the standard optimizer. Recent studies have introduced novel optimizers that claim to accelerate pretraining by 1.4$\times$ to 2$\times$ compared to AdamW~\cite{liu2024sophiascalablestochasticsecondorder,vyas2025soapimprovingstabilizingshampoo,liu2025muonscalablellmtraining,yuan2025mars,liang2025cautious,wang2025sharpnessdisparityprincipletransformers,liu2025focusorderconcentratedupdating,pethick2025trainingdeeplearningmodels,ma2025swansgdnormalizationwhitening}, yet these optimizers have not yet been widely adopted in real-world pretraining~\cite{deepseekai2024deepseekllmscalingopensource,yang2025qwen3technicalreport,grattafiori2024llama3herdmodels}, with the notable exception of Kimi K2~\cite{kimiteam2025kimik2openagentic}, which uses the Muon-clip optimizer (a variant of Muon~\cite{jordan2024muon}).

We pinpoint two problems in optimizer evaluation in the evaluation process of these optimizers that both undermine confidence in new methods and limit their practical adoption. First, some baselines suffer from improperly tuned hyperparameters. Second, many experiments are confined to smaller-scale settings, leaving unanswered questions about how these optimizers perform in broader, more realistic scenarios.

To address these issues, we benchmark eleven optimizers, including AdamW, in a rigorously controlled setup, focusing on two key questions:
\begin{enumerate}[leftmargin=*]
  \item \textbf{How to ensure hyperparameter optimality?} Previous works typically rely on manual hyperparameter selection and keep common hyperparameters such as learning rate and weight decay fixed across optimizers (e.g. \cite{liu2025muonscalablellmtraining}). This tuning process may result in a weak baseline, as the hyperparameters chosen may favor the proposed optimizers rather than for AdamW. We validate this concern by showing that tuning just one hyperparameter in the widely adopted GPT-3 recipe (introduced in~\cite{brown2020languagemodelsfewshotlearners} and used in~\cite{liu2024sophiascalablestochasticsecondorder,wen2024understandingwarmupstabledecaylearningrates,yuan2025mars,liu2025physicsskilllearning,liu2025focusorderconcentratedupdating,liang2025cautious,wang2025sharpnessdisparityprincipletransformers}) can yield a 2$\times$ speedup for pretraining (\Cref{fig:motivation}, top left). To address, this we perform coordinate descent in the hyperparameter space on N hyperparameters for all eleven optimizers, iterating until convergence across for each of six different settings for models with up to 0.5B parameters to ensure that the hyperparameters are near-optimal for each setting.
  \item \textbf{How do speedups differ across different scaling regimes?} Large language models are trained in different regimes: depending on the model size, the data-to-model ratio (the number of tokens over the number of parameters) can range from $1$ to more than $50$ times the Chinchilla optimal (about $20$ based on~\cite{hoffmann2022trainingcomputeoptimallargelanguage}). Typical optimizer experiments in $1\times$ Chinchilla optimal regimes raise concerns about a new optimizer's effectiveness in high data-to-model ratio. To address this, we benchmark the optimizers in four distinct data-to-model ratios (1$\times$, 2$\times$, 4$\times$ and 8$\times$ the Chinchilla optimal regime) and scale up to  1.2B parameter models (following~\cite{everett2024scalingexponentsparameterizationsoptimizers,zhang2025doescriticalbatchsize}).
\end{enumerate}

Concretely, we employ the Llama 2 architecture \cite{touvron2023llama2openfoundation,grattafiori2024llama3herdmodels} 
(ranging from 0.1B to 1.2B parameters) and a data mixture similar to OLMo 2's~\cite{olmo20252olmo2furious}. We focus on the final validation loss on the C4‐EN mixture as a known proxy for downstream performance~\cite{bhagia2024establishingtaskscalinglaws}, while also tracking exact downstream performance on various benchmarks. As previous works~\cite{vyas2025soapimprovingstabilizingshampoo,liu2025muonscalablellmtraining} show that the step-wise computation overhead of matrix-based optimizers can be reduced to under $10\%$ through proper implementation, we primarily compare algorithms by the number of tokens needed to reach a given loss.

Our empirical results show the necessity of careful hyperparameter tuning and end-of-training evaluations across a range of model scales and data-to-model ratios:
\begin{enumerate}
[leftmargin=*]
  \item \textbf{Hyperparameter transfer between optimizers is non-trivial.} Even similar optimizers may need very different hyperparameters (e.g., Lion's optimal weight decay $\approx$0.6 vs.\ AdamW's $\approx$0.1, \Cref{fig:motivation}, top right), so fixing hyperparameters across optimizers can lead to unfair comparisons.
    \item \textbf{The speedup of new optimizers is lower than claimed and diminishes with model size.} Many reported speedups of 2$\times$ simply reflect a weak baseline. Against our well-tuned AdamW baseline, the speedup of alternative optimizers does not exceed 1.4$\times$(\Cref{fig:speedup}). Furthermore, while new optimizers such as Muon and Soap show 1.3$\times$ speedups for small models (0.1B), the speedups diminish to around 1.1$\times$ for 1.2B parameter models at 8$\times$ Chinchilla Ratio (\Cref{fig:motivation}, bottom left), a regime that is not tested in previous works studying the scaling law of these optimizers \footnote{The original Soap paper~\cite{vyas2025soapimprovingstabilizingshampoo} investigate model sizes up to 0.6B parameters. and Kimi's paper on Muon~\cite{liu2025muonscalablellmtraining} only considers $1\times$ Chinchilla regime.}.
    \item \textbf{Early-stage loss curves can mislead significantly.} During learning rate decay, loss curves of different optimizers may cross multiple times (\Cref{fig:necessity}), so judging optimizers using intermediate checkpoints may result in a different ranking than comparing models at the target training budget.
\end{enumerate}

Our benchmarking also reveals new insights about optimizer design: 

\begin{enumerate}[leftmargin=*]
    \item \textbf{Matrix-based optimizers consistently outperform scalar-based optimizers for small models.} 
    \emph{Scalar-based optimizers} (e.g., AdamW, Lion, Mars, etc.) update each parameter individually using scalar operations. After proper tuning, all scalar-based optimizers achieve similar optimization speeds to AdamW, with an average speedup ratio of less than 1.2$\times$. \emph{Matrix-based optimizers} (e.g., Kron, Muon, Soap, etc.) leverage the inherent matrix structure of neural network parameters and precondition gradients using matrix multiplication. Despite their diverse update rules, matrix-based optimizers all deliver approximately a 1.3$\times$ speedup over AdamW (\Cref{fig:motivation}, bottom right) for models under 520M parameters.

    \item \textbf{Optimal choice of optimizer shifts depends on data-to-model ratios.} A winner in the $1\times$ Chinchilla regime may be suboptimal when data-to-model ratio increases. For example, while Muon is consistently the best optimizer in smaller Chinchilla ratio regimes, it is outperformed by Kron and Soap when the data-to-model ratio increases to 8x or larger (\Cref{fig:speedup,fig:case}). 
\end{enumerate}

\section{Related Works}

\textbf{Optimizers for Deep Learning.} A long line of work has studied optimization
for deep learning, incorporating insights from classical optimization literature~\cite{robbins1951stochastic,nesterov1983method,duchi2011adaptive} and domain knowledge about deep neural networks~\cite{pmlr-v28-sutskever13}. (i) Early insights motivated the rise of optimizers that use adaptive learning rates based on second-order momentum~\cite{tieleman2012lecture,zeiler2012adadeltaadaptivelearningrate,kingma2017adammethodstochasticoptimization}. Adam~\cite{kingma2017adammethodstochasticoptimization} later became the default baseline for optimizers with adaptive learning rates. Since then, improvements over Adam and SGD have been proposed, with notable examples including Nesterov Adam~\cite{dozat2016incorporating} and AdamW with decoupled weight decay~\cite{loshchilov2019decoupled}. Other improvements include addressing the convergence of Adam on convex loss~\cite{j.2018on,zaheer2018adaptive,taniguchi2024adoptmodifiedadamconverge}, considering interpolation between Adam and SGD to improve generalization~\cite{luo2019adaptivegradientmethodsdynamic,xie2022adaptiveinertiadisentanglingeffects}, performing further variance reductions on optimizer updates~\cite{liu2021varianceadaptivelearningrate,zhang2019lookaheadoptimizerksteps,yuan2025mars,xie2024adanadaptivenesterovmomentum,pagliardini2024ademamixoptimizerbetterfaster,zhuang2020adabeliefoptimizeradaptingstepsizes}, incorporating momentum on weights~\cite{ivgi2023dogsgdsbestfriend,defazio2024roadscheduled}, allowing easier hyperparameter tuning~\cite{pmlr-v202-defazio23a,mishchenko2024prodigyexpeditiouslyadaptiveparameterfree,defazio2024optimallineardecaylearning}, reducing memory usage by incorporating the structure of 
neural networks~\cite{shazeer2018adafactoradaptivelearningrates,zhang2025adamminiusefewerlearning,zhu2025apollosgdlikememoryadamwlevel,luo2023cameconfidenceguidedadaptivememory,modoranu2024microadam,zhao2024galorememoryefficientllmtraining}, and modifying the algorithm to allow larger batch sizes~\cite{you2017largebatchtrainingconvolutional,you2020largebatchoptimizationdeep}.
(ii) Starting from Preconditioned SGD~\cite{Li_2018} and Shampoo~\cite{gupta2018shampoopreconditionedstochastictensor}, another line of optimizer design~\cite{morwani2024newperspectiveshampoospreconditioner,eschenhagen2023kronecker,martens2020optimizingneuralnetworkskroneckerfactored,li2018preconditionermatrixliegroup,eschenhagen2025purifyingshampooinvestigatingshampoos} began to incorporate matrix preconditioners instead of simple scalar preconditioners. Techniques including learning rate grafting~\cite{agarwal2020disentanglingadaptivegradientmethods}, blocking, and distributed methodology~\cite{anil2021scalablesecondorderoptimization} have since been proposed. These matrix-based approaches later led to the theory of modular duality in deep learning optimization~\cite{bernstein2024modulardualitydeeplearning,bernstein2024oldoptimizernewnorm,large2024scalableoptimizationmodularnorm} and new optimizers including Muon~\cite{jordan2024muon} and Scion~\cite{pethick2025trainingdeeplearningmodels}.
(iii) Motivated by Newton's algorithm, there has been a line of work that tries to incorporate Hessian information~\cite{becker:improving,Yao_Gholami_Shen_Mustafa_Keutzer_Mahoney_2021,schaul2013peskylearningrates,Yao_Gholami_Shen_Mustafa_Keutzer_Mahoney_2021}. (iv) Symbolic discovery of optimizers~\cite{chen2023symbolic} has discovered a memory-efficient SignGD~\cite{pmlr-v80-bernstein18a} variant called Lion, which claims to outperform Adam on a wide range of tasks.

\textbf{Optimization for Pretraining.} Since~\cite{brown2020languagemodelsfewshotlearners}, the cost of pretraining has increased dramatically. One of the challenges is how to choose hyperparameters with minimal cost.  
A pivotal line of work in this direction is the tensor program series that allows for extrapolating some hyperparameters across scales~\cite{yang2020tensor,yang2020tensor2,yang2020scalinglimitswideneural,yang2021tensorprogramsiwide,yang2023tensorprogramsivbadaptive,yang2022tensorprogramsvtuning,yang2024spectralconditionfeaturelearning}. Empirical results suggest that fitting a power law to scale hyperparameters is now common practice for large models~\cite{li2025predictablescalei,everett2024scalingexponentsparameterizationsoptimizers,liu2025muonscalablellmtraining,deepseekai2024deepseekllmscalingopensource,zhang2025doescriticalbatchsize}. We have also incorporated this approach in our paper. 
Another popular line of research is designing better optimizers specifically for pretraining. These optimizers are our main objects of study. An (incomplete) list of optimizers and their claimed speedups over AdamW includes Sophia~\cite{liu2024sophiascalablestochasticsecondorder} (2$\times$), Soap~\cite{vyas2025soapimprovingstabilizingshampoo} (1.4$\times$), Muon~\cite{jordan2024muon,liu2025muonscalablellmtraining} (2$\times$), MARS~\cite{yuan2025mars} (2$\times$), Cautious AdamW~\cite{liang2025cautious} (2$\times$), Block-wise Learning Rate Adam~\cite{wang2025sharpnessdisparityprincipletransformers} (2$\times$), FOCUS~\cite{liu2025focusorderconcentratedupdating} (2$\times$), SWAN~\cite{ma2025swansgdnormalizationwhitening} (2$\times$), DION~\cite{ahn2025diondistributedorthonormalizedupdates} (3$\times$), and SPlus~\cite{frans2025stablewhiteningoptimizerefficient} (2$\times$). We present a comparison of setups with these prior works in ~\Cref{sec:comparison-with-prior}.

\textbf{Re-evaluation Methodology.} Our work is a rigorous evaluation of optimizers for pretraining. Rigorous evaluation has been an important part of deep learning research to clarify the current status of research and move the community forward. \cite{jiang2019fantasticgeneralizationmeasures} critically examines metrics for predicting LLMs' generalization capability and has facilitated research on understanding loss landscape sharpness and generalization and new optimizers such as SAM~\cite{foret2021sharpnessawareminimizationefficientlyimproving}. \cite{schmidt2021descendingcrowdedvalley} re-evaluated optimizers at that time and showed that (i) which optimizer is optimal is problem-specific, and (ii) rigorous hyperparameter tuning is required and unequal tuning can account for most of the claimed speedup. Unlike~\cite{schmidt2021descendingcrowdedvalley}, we evaluate the LLM pretraining task and include modern optimizers they did not test. However, we arrive at a similar conclusion: rigorous and fair hyperparameter tuning is still not the norm, but rather the exception in optimizer design research.~\cite{zhao2025deconstructingmakesgoodoptimizer} also examines how different optimizers perform on the pretraining tasks. However, the focus of~\cite{zhao2025deconstructingmakesgoodoptimizer} is on understanding loss structure, and the optimizers tested in the paper are shown to have slower convergence compared to AdamW, whereas the optimizers tested in this paper show small but significant improvements in convergence speed. The Algoperf competition~\cite{kasimbeg2025acceleratingneuralnetworktraining} evaluates different optimizers across different settings and arrives at a similar conclusion that (i) matrix-based optimizers and (ii) denoising methods such as Nesterov momentum lead to speedup over AdamW. Our works focus on the pretraining setting and investigates the effect of scaling dataset and model sizes.~\citet{kaddour2023traingainrevisitingefficient} compares optimizers including Sophia, as well as some other algorithmic improvements, in the setting of training Bert and T5-like models. They also observe the ordering of optimizers can flip during learning rate decay, and AdamW remains competitive when properly tuned.

\paragraph{Comparison with concurrent work~\citet{semenov2025benchmarkingoptimizerslargelanguage}} \citet{semenov2025benchmarkingoptimizerslargelanguage} investigates the same topic of benchmarking optimizers for pre-training. These two works agree on many high-level points, e.g., (i) non-zero weight decay and decaying to a small learning rate are essential for pretraining; (ii) variance-reduced AdamW variants such as Mars show non-trivial speedups over vanilla AdamW. 

However, our results differ in the relative performance of matrix-level optimizers. Our paper shows that matrix-level optimizers such as Muon can achieve significant speedups over variance-reduced versions of AdamW, such as Mars, their study finds that AdEMaMix and Mars outperform Muon.  Our initial investigation suggests this discrepancy largely stems from the differences in the batch sizes used in experiments. Their most extensively tuned experiments on 130M models use a batch size of only 0.1M and 0.02M tokens, whereas our experiments operate with tuned batch sizes that are not smaller than 0.4M tokens. This difference is likely a result of different hardware regimes. We leverage 128 TPU-v5lite chips (which are approximately equivalent to 12 H100 GPUs), where only larger batches (larger than 0.4M) can fully utilize the parallelism in the compute. On the other side, their experiments appear to be primarily conducted on 1-8 H100 GPUs, where smaller batches might be preferable. Since Mars and AdEMAMix both perform gradient averaging and variance reduction, these methods are advantangeous in their noise-dominated small-batch regime, whereas in our larger-batch setting these benefits diminish and matrix-level optimizers become more competitive. 

In their larger scale experiments,~\citet{semenov2025benchmarkingoptimizerslargelanguage} increase both model size to 720M parameters and batch size to 1M tokens, respectively, which is closer to our setting. However, we differ in hyperparameter tuning methodology: (i) we conduct more extensive sweeps over learning rates on our 520M scales experiments and generally find higher values (4e-3 to 8e-3) compared to their choices (1e-3 to 2e-3), which are transferred from smaller scale and for some optimizers are not tuned; and (ii) for Muon, we separately tune the learning rates for the embedding layers and the matrix weights, which also improves its performance. These differences highlight the sensitivity of optimizer benchmarking to hardware and tuning strategies, underscoring the importance of carefully controlled experimental design when comparing optimizer performance.

\section{Methodology}

In this section, we detail the experimental design and evaluation protocol that underpin our empirical investigation. In \Cref{sec:experimental-setup}, we specify the general setup for all subsequent studies. We then describe our three-phase hyperparameter-tuning framework: Phase I (in \Cref{sec:phase1}) performs fine-grained coordinate-descent sweeps across multiple model sizes and data-to-model ratios to identify scaling-sensitive parameters; Phase II (in \Cref{sec:phase2}) refines these sensitive parameters on mid-scale settings and selects the most promising optimizers; and Phase III (in \Cref{sec:phase3}) extrapolates hyperparameter scaling laws to the 1.2 billion-parameter regime. Together, these protocols ensure principled, fair, and reproducible comparisons across different optimizers. We present the optimal configurations found and how loss changes with respect to each hyperparameter in~\Cref{app:experiments} and hope that this can facilitate future research. We also open-source the code (\url{https://github.com/marin-community/marin/tree/kaiyue/optimizers}) and the corresponding WandB runs\ (\url{https://wandb.ai/marin-community/optimizer-scaling}).

\subsection{General Experimental Setup}
\label{sec:experimental-setup}

Following~\cite{olmo20252olmo2furious}, we conduct all our experiments on a large-scale pretraining corpus composed of three publicly available datasets, tokenized with the Llama3 tokenizer: DCLM-baseline (3.8 trillion tokens,~\cite{li2025datacomplmsearchgenerationtraining}), StarCoder V2 Data (0.25 trillion tokens,~\cite{lozhkov2024starcoder2stackv2}), and ProofPile 2 (55 billion tokens,~\cite{azerbayev2024llemmaopenlanguagemodel}). 

\begin{table}[h]
  \centering
  \begin{tabular}{lll}
    \toprule
    \textbf{Optimizer}           & \textbf{References}             & \textbf{Algorithm} \\
        \midrule
    \multicolumn{3}{l}{\textbf{Baseline}} \\
    \midrule
    AdamW                    &  \cite{kingma2017adammethodstochasticoptimization,loshchilov2019decoupled} & \Cref{alg:adamw}  \\
    \midrule
  
    \multicolumn{3}{l}{\textbf{Variance-reduced AdamW Variants}} \\
    \midrule
    NadamW                    &\cite{dozat2016incorporating}        & \Cref{alg:nadamw}           \\
    Mars                      &\cite{yuan2025mars}                  & \Cref{alg:mars}              \\
    Cautious                  &\cite{liang2025cautious,wang2024cadamconfidencebasedoptimizationonline} & \Cref{alg:cautious}  \\
    \midrule
    \multicolumn{3}{l}{\textbf{Memory-efficient Optimizers}} \\
    \midrule
    Lion                      &\cite{chen2023symbolic}              & \Cref{alg:lion}              \\
    Adam-mini             &\cite{zhang2025adamminiusefewerlearning} & \Cref{alg:adammini}        \\
    \midrule
    \multicolumn{3}{l}{\textbf{Matrix-based Optimizers}} \\
    \midrule
    Muon                     &\cite{jordan2024muon}                & \Cref{alg:muon}              \\
    Scion                    &\cite{pethick2025trainingdeeplearningmodels} & \Cref{alg:scion}       \\
    Kron (PSGD)              &\cite{Li_2018,li2022blackboxliegroup} & \Cref{alg:kron}            \\
    Soap                     &\cite{vyas2025soapimprovingstabilizingshampoo} & \Cref{alg:soap}       \\
    \midrule
    \multicolumn{3}{l}{\textbf{Hessian-Approximation Optimizers}} \\
    \midrule
    Sophia                  &\cite{liu2024sophiascalablestochasticsecondorder} & \Cref{alg:sophia}       \\
    \bottomrule
  \end{tabular}
  \caption{Optimizers under study}
  \label{tab:optimizers}
\end{table}

\begin{table}[t]
  \centering
  \begin{tabular}{lrrrrrr}
    \toprule
    \textbf{Model} & \textbf{Params} & \textbf{Seq Len} & \textbf{Hidden Dim} & \textbf{Inter Dim} & \textbf{\# Layers} & \textbf{\#  Heads} \\
    \midrule
    Llama-130M   & 130M  & 4096 & 512  & 2048  & 32 & 8   \\
    Llama-300M   & 300M  & 4096 & 768  & 3072  & 32 & 12  \\
    Llama-520M   & 520M  & 4096 & 1024 & 4096  & 32 & 16  \\
    Llama-1.2B   & 1.2B  & 4096 & 1536 & 6144  & 32 & 24 \\
    \bottomrule
  \end{tabular}
  \caption{Detailed architecture hyperparameters for each model size we studied.}
  \label{tab:model-configs}
\end{table}

Our benchmarks cover four model sizes derived from the architecture~\cite{touvron2023llamaopenefficientfoundation,touvron2023llama2openfoundation,grattafiori2024llama3herdmodels}, with approximately 130M, 300M, 520M, and 1.2B parameters. Each variant uses a fixed sequence length of 4,096 and 32 transformer layers (following~\cite{liu2024mobilellmoptimizingsubbillionparameter}), differing only in hidden dimension, intermediate dimension, and number of attention heads. Detailed hyperparameters are summarized in \Cref{tab:model-configs}. Training is implemented in JAX and executed on TPU v5 hardware. We employ a mixed-precision scheme (parameters in fp32 and activations in bf16). For each model, we use $20$ times its non-embedding parameter count to compute the Chinchilla optimal data-to-model ratio based on~\cite{hoffmann2022trainingcomputeoptimallargelanguage}. We will use \textbf{$n \times$} Chinchilla to represent training the models for $n$ times the Chinchilla optimal number of tokens.

Our primary evaluation metric for the model is the language modeling loss on the English split of the C4 dataset~\cite{raffel2023exploringlimitstransferlearning}, which has been shown to be a strong proxy for downstream performance~\cite{bhagia2024establishingtaskscalinglaws}. We also track downstream accuracy and bits-per-byte on the following suite of benchmarks: ARC (Easy and Challenge)~\cite{Clark2018ARC}, BoolQ~\cite{Clark2019BoolQ}, COPA~\cite{Gordon2012COPA}, CommonsenseQA~\cite{Talmor2018CommonsenseQA}, HellaSwag~\cite{Zellers2019HellaSwag}, LAMBADA~\cite{Paperno2016LAMBADA}, OpenBookQA~\cite{Mihaylov2018OpenBookQA}, PIQA~\cite{Bisk2020PIQA}, WSC273~\cite{levesque2012winograd}, and Winogrande~\cite{Sakaguchi2020WinoGrande}.

Our study includes a wide range of eleven optimizers listed in~\Cref{tab:optimizers}. Due to the page limit, we defer the exact algorithm descriptions of these optimizers to~\Cref{def:optimizerdesign}. We selected these eleven optimizers according to three guiding principles: (i) include widely adopted baselines such as AdamW and Lion; (ii) cover recently proposed optimizers; and (iii) when multiple methods share a similar update rule, choose a few representative algorithms. These choices ensure both breadth and depth in our comparison. We group the optimizers under study into five general classes:
\begin{enumerate}[leftmargin=*]
\item Our baseline algorithm is AdamW, with $m_t$ and $v_t$ being first- and second-order momentum of gradient. AdamW's update rule is $w_{t+1} = w_t 
    - \eta\,\frac{{m_t}}{\sqrt{v_t} + \epsilon}
    - \eta\,\lambda\,w_t$.
\item Reducing the variance of updates is a shared motivation behind many optimizers. For example, Nesterov AdamW incorporates the Nesterov lookahead technique to estimate gradients more accurately, with the following update rule:
$w_{t+1} = w_t 
    - \eta\,\frac{\mathbf{\beta_1 m_t + \color{red}( 1 - \beta_1)g_t}}{\sqrt{v_t} + \epsilon}
    - \eta \,\lambda\,w_t$.
\item Optimizers such as Lion aim to reduce the memory required by AdamW by only keeping the first-order momentum. Lion is observed to perform better than AdamW at a small scale~\cite{liang2025cautious}: $w_{t+1} = w_t 
    - \eta\, \mathrm{sign}(\beta_2 m_t + (1 - \beta_2)g_t)$, where $m_t$ is the first-order momentum.
\item Other optimizers like Muon leverage the matrix structure of neural networks and perform preconditioning of gradients through matrix multiplication instead of scalar multiplication. For Muon, the key operation is called Newton-Schulz: $\mathrm{NS}(M) = M (aM + bM^\top M + c(M^\to pM)^2)$. With appropriate $a,b,c$, one can prove that $\mathrm{NS}^{(5)}(M) \approx \arg \max_{\|O\|_{\mathrm{op}} = 1} \mathrm{Tr}( O^\top M)$ when $\|M\|_{\mathrm{op}} < 1$. Muon has the following update rule: $w_{t+1} = w_t 
    - \eta\, \mathrm{NS^{(5)}}(\beta_2 \tilde m_t + (1 - \beta_2)g_t)$. This is done for all the matrices except the token classification head and the embedding in the network. 
\item Optimizers including Sophia are motivated by the famous Newton's method and use Hessian-vector product to approximate the diagonalized version of the Hessian matrix empirically.\footnote{We defer the result of Sophia to~\Cref{app:sophia_experiments}.}
 \end{enumerate}

\subsection{Phase I: Fine-grained Hyperparameter Coordinate Descent}
\label{sec:phase1}

\begin{table}[b]
  \centering
  \begin{tabular}{lccccccccccc}
    \toprule
    \textbf{Stage} & \textbf{LR} & \textbf{WD} & \textbf{min lr ratio} & \textbf{Warmup} & \textbf{Max Grad Norm} & \textbf{Batch} & \textbf{Val.\ Loss} \\
    \midrule
    Init      & 0.008 & 0.1 & 0 & 1000 &  1  & 256 & 3.298 \\
    Round 1   & 0.008 & 0.1 & 0 & \textcolor{red}{2000}  & 1  & 256 & 3.282 \\
    Round 2   & 0.008 & 0.1 & 0 & 2000 & 1 & \textcolor{red}{128} & 3.263 \\
    Best      & 0.008 & 0.1 & 0 & 2000 & \textcolor{red}{2} & 128 & 3.263 \\
    \bottomrule
  \end{tabular}
  \caption{Illustrative coordinate‐descent steps for AdamW on the 130M 1× Chinchilla regime. Changed hyperparameter values are highlighted in red; We omitted some unchanged hyperparameters ($\beta_1 = 0.9,\beta_2 = 0.98 ,\epsilon = 10^{-10}$).}
  \label{tab:adamw-sweep-example-full}
\end{table}
Fixed or lightly tuned baselines can severely understate an optimizer's capability and lead to overstated speedup claims. For instance, a single tweak to the learning-rate schedule with peak learning rate 6e-4 in the GPT-3 recipe \cite{brown2020languagemodelsfewshotlearners} can produce nearly a 2× speedup (\Cref{fig:motivation}, left). 
To avoid such artifacts, we perform an exhaustive, one-at-a-time sweep over each hyperparameter, identify the set of near-best configurations in each regime, and then determine which knobs truly require re-tuning as scale changes.

For each optimizer, we define a discrete grid for every hyperparameter (e.g., for AdamW, our swept hyperparameters include learning rate, weight decay, warmup steps, $\beta_{1}$, $\beta_{2}$, $\epsilon$, gradient-norm clipping, and batch size). Starting from a initial hyperparameter configuration similar to the hyperparameter configuration provided in the original paper proposing the optimizer, in each iteration, we hold all but one hyperparameter fixed at the current best values, then search the whole grid for that parameter, and accept the new value if the validation loss improves by more than $\Delta_1 = 3 \times 10^{-3}$. We repeat passes until no parameter update yields further significant gain. 
We perform this sweeping for 6 different settings, namely 130M, 300M, 500M at 1× Chinchilla and 130M at 2×, 4×, 8× Chinchilla. 
One exemplary  hyperparameter optimization procedure for AdamW on a model with 300M parameters and 1× Chinchilla is shown in~\Cref{tab:adamw-sweep-example-full}.

\paragraph{Result of Phase I.} By the end of Phase I, we have, for each optimizer and each of the six regimes (130M, 300M, 500M at 1×; 130M at 2×, 4×, 8× Chinchilla), identified a coordinate-wise local optimum of hyperparameters—that is, the single value which, when all other knobs are held fixed, minimizes validation loss. 

\subsection{Phase II: Coordinate Descent on Scaling-Sensitive Hyperparameters}
\label{sec:phase2}

\begin{table}[t]
  \centering
  \begin{tabular}{ll}
    \toprule
    \textbf{Optimizer} & \textbf{Scaling‐Sensitive Hyperparameters} \\
    \midrule
    AdamW             & \texttt{learning rate, warmup, weight decay, batch size} \\
    Nesterov AdamW    & \texttt{learning rate, warmup} \\
    Lion               & \texttt{learning rate, beta2} \\
    Adam-mini          & \texttt{learning rate, weight decay, warmup} \\
    Cautious           & \texttt{learning rate, batch size} \\
    Mars               & \texttt{learning rate, warmup, beta1} \\
    Scion              & \texttt{learning rate, beta1, \# decay steps in WSD \cite{hu2024minicpmunveilingpotentialsmall}} \\
    Muon               & \texttt{learning rate} \\
    Soap            & \texttt{learning rate, warmup, block size} \\
    Kron               & \texttt{learning rate} \\
    \bottomrule
  \end{tabular}
  \caption{Scaling‐sensitive hyperparameters identified in Phase I (\Cref{sec:phase1}). To reduce the memory required by Soap, we apply the parameter blocking as in~\cite{anil2021scalablesecondorderoptimization}} 
  \label{tab:sensitive-hyper}
\end{table}

While the extensive coordinate descent guarantees coordinate-wise optimality, it is too costly to perform on larger-scale experiments.
We empirically observe two crucial properties of the sweeping that allow us to simplify the descent procedure:

\begin{enumerate}
    \item Losses are sensitive to only a subset of hyperparameters; many have little effect on performance when perturbed from their optimal values.
    \item Among the sensitive hyperparameters, the optimal settings for most remain stable across scales, so tuning is needed only at smaller scales.
\end{enumerate}

Building on these observations, we can simplify our hyperparameter search by identifying scaling-sensitive parameters, which (i) crucially influence the final performance, and (ii) change with respect to the model scale.
We define the \textbf{approximate-optimal configuration} for a given regime as the set of all hyperparameter tuples whose final loss lies within $\Delta_2=$6.4e-3 of the regime's best-observed loss $L^*_r$, where $r$ denotes a regime/setting, that is, $r$ is a pair of choice of model size and data budget. Concretely, let $c_h$ denotes a hyperparameter (where $h$ is the index) and $c$ be the tuple of all hyperparameters.
\begin{enumerate}[leftmargin=*]
    \item For each regime $r$, let $\mathcal{C}_r = \{\,c: L(c)\le L^*_r + \Delta_2\}$ be all the hyperparameter configurations in our coordinate descent procedure that yield approximately optimal loss. 
    \item A hyperparameter $c_h$ is called \textbf{scaling-insensitive} if there exists a single value $v_h$ such that there exists  $c\in\mathcal{C}_r$ such that $c_h=v_h$  for every regime $r$ in Phase I. 
    Otherwise, $c_h$ is \textbf{scaling-sensitive}, meaning its optimal value shifts depending on model size or data budget.
\end{enumerate}

We carry forward only the scaling-sensitive hyperparameters (shown in~\Cref{tab:sensitive-hyper}) into Phase II, thereby focusing our next round of coordinate-descent sweeps on the hyperparameters that truly require re-tuning across scaling regimes.
We then perform sweeping for another 6 different settings, namely 300M, 500M at 2×, 4×, 8× Chinchilla. 

\paragraph{Result of Phase II.}  Combined with the results in Phase I, we obtain a set of near-optimal hyperparameters and their corresponding losses for 12 different settings (130M, 300M, 500M and 1×, 2×, 4×, 8× Chinchilla).  

To quantify the speedup of different optimizers over the baseline AdamW, we fit how AdamW's loss scales with data budget $D$ for each model size $N$ with the following functional form: $\hat L_N(D) = \alpha_N D^{-B_N} + \beta_N$. Suppose an optimizer achieves loss $L_{\text{optimizer}}$ at data budget $D_{\text{Optimizer}}$. We calculate the corresponding data budget needed for  AdamW to achieve this loss, denoted by $D_{\text{AdamW}}$ by finding the solution to the equation $\hat L_N(D_{\text{AdamW}}) = L_{\text{optimizer}}$. We then use $D_{\text{AdamW}}/D_{\text{Optimizer}}$ as the estimated speedup ratio.

Through this set of experiments, we observed two phenomena: (i) matrix-based optimizers consistently outperform scalar-based optimizers, but all optimizers' speedup ratios over AdamW do not exceed 1.5$\times$; and (ii) within matrix-based optimizers, Muon performs the best at 1-4× Chinchilla ratio but is overtaken by Soap and Kron when the Chinchilla ratio increases.

\subsection{Phase III: Hyperparameter Scaling Law for Further Extrapolation}
\label{sec:phase3}

Having obtained optimized hyperparameter settings from Phase II (\Cref{sec:phase2}), we now fit a smooth scaling law that predicts the optimal value of each scaling-sensitive hyperparameter as a function of model size $N$ and data budget $D$.
Concretely, we model the optimal value for each scaling-sensitive hyperparameter $h$ as:
$
h(N, D) \;=\; \alpha \,N^{-A}\,D^{-B} \;+\; \beta,
$
where $A$, $B$, $\alpha$, and $\beta$ are learned coefficients. 

We estimate these parameters via non-linear least-squares on the 12 observed $(N,D,h)$ triples for each optimizer, minimizing the squared error between predicted and actual optimal hyperparameter values. To test the quality of our prediction, we ran a full Phase I sweep at $N=1.2\mathrm{B}$ and Chinchilla = 1 for AdamW. 
Comparing the identified optimum against our fitted hyperparameters, we observe that our predicted hyperparameters yield a final loss within 3e-3 of the optimal configuration, showing that our hyperparameter scaling law can effectively predict the optimal hyperparameters. 
We then performed two case studies to further extrapolate our benchmarking:
\begin{enumerate}[leftmargin=*]
    \item To test the effect of scaling up model sizes, we train 1.2B models using AdamW, Nesterov AdamW, and Muon on 1 to 8× Chinchilla ratio. 
    \item To further test the effect of different optimizers when data-to-model ratios are high, we train 130M and 300M models using AdamW, Nesterov AdamW, Muon, and Soap on 16x Chinchilla ratio.
\end{enumerate}

\paragraph{Results of Phase III.} We demonstrate two potential shortcomings of Muon optimizers, which is the best optimizers in Phase I and Phase II, through experiments in this phase: (i) while Muon's speedup persists for models up to 1.2B parameters, the speedup decreases to under 1.2×; (ii) With a 16× Chinchilla ratio, NAdamW and Soap outperform Muon on the 130M model, and Soap also surpasses Muon on the 300M model.

\section{Empirical Findings}

\begin{figure}[t]
    \centering
    \begin{minipage}{0.32\textwidth}
        \centering
        \includegraphics[width=\linewidth]{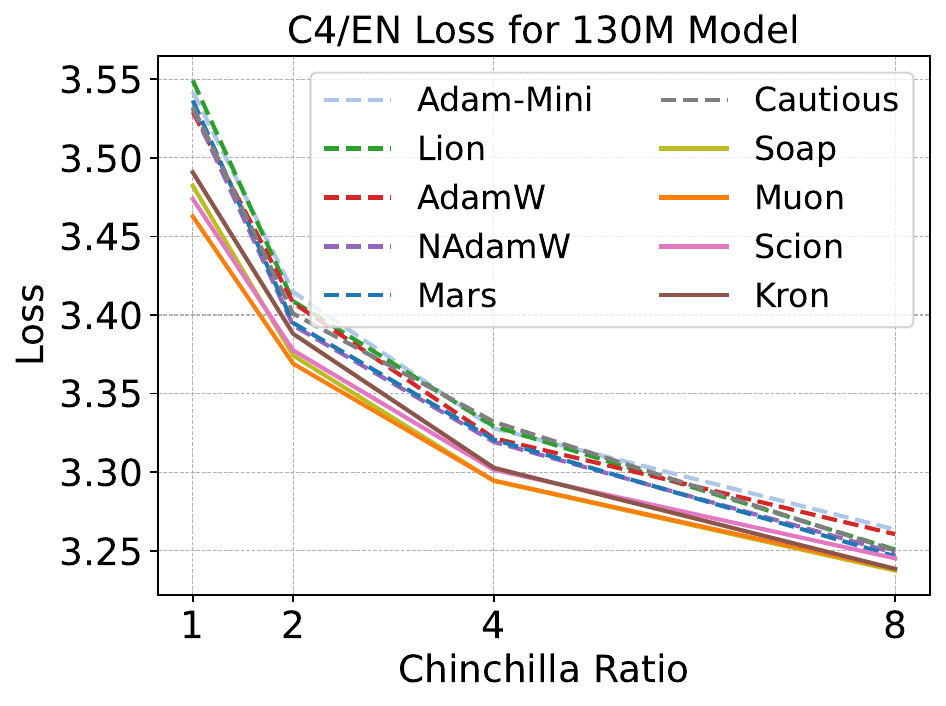}
    \end{minipage}%
    \hfill
    \begin{minipage}{0.32\textwidth}
        \centering
        \includegraphics[width=\linewidth]{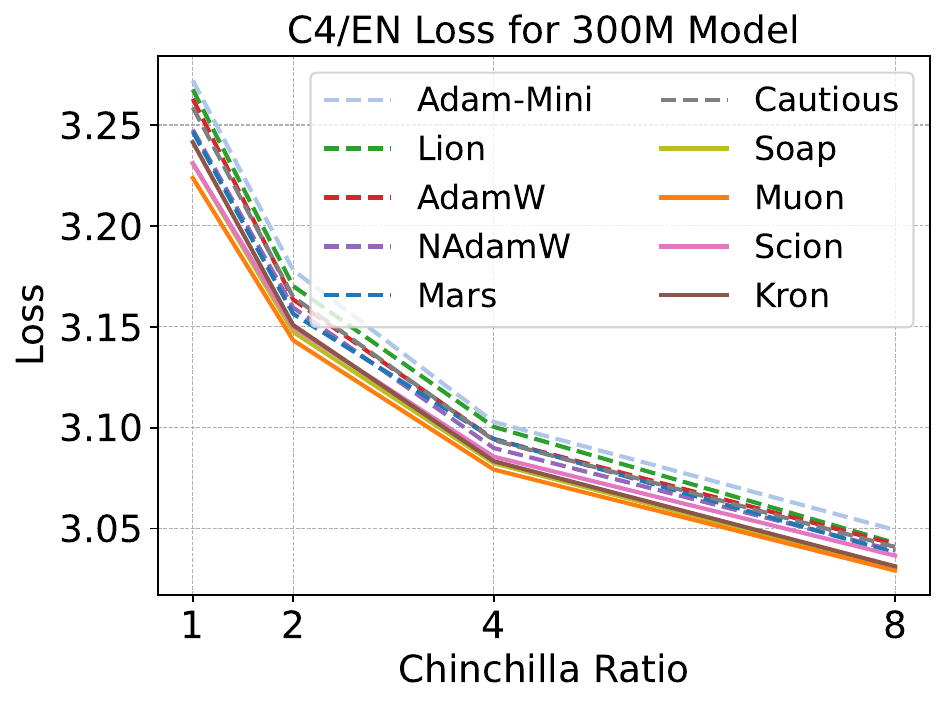}
    \end{minipage}%
    \hfill 
    \begin{minipage}{0.32\textwidth}
        \centering
        \includegraphics[width=\linewidth]{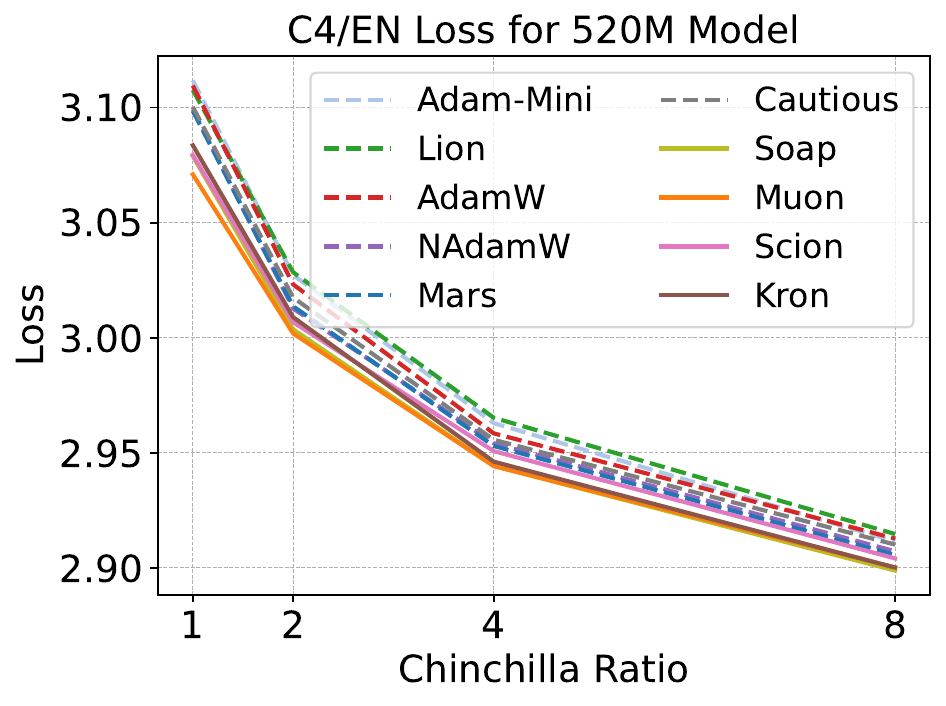}
    \end{minipage}%

    \begin{minipage}{0.32\textwidth}
        \centering
        \includegraphics[width=\linewidth]{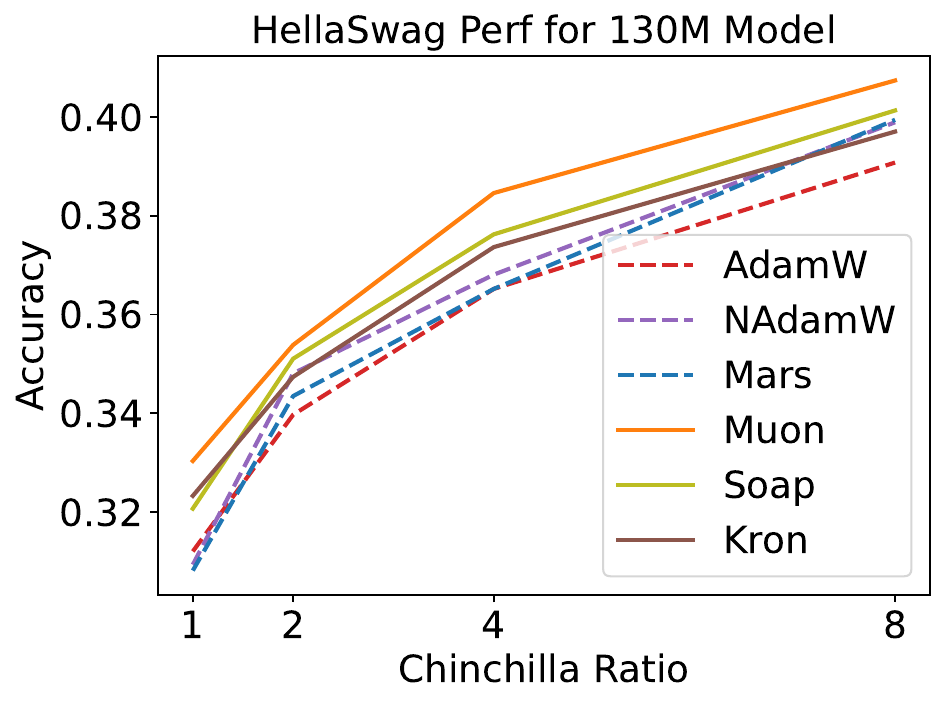}
    \end{minipage}
    \begin{minipage}{0.32\textwidth}
        \centering
        \includegraphics[width=\linewidth]{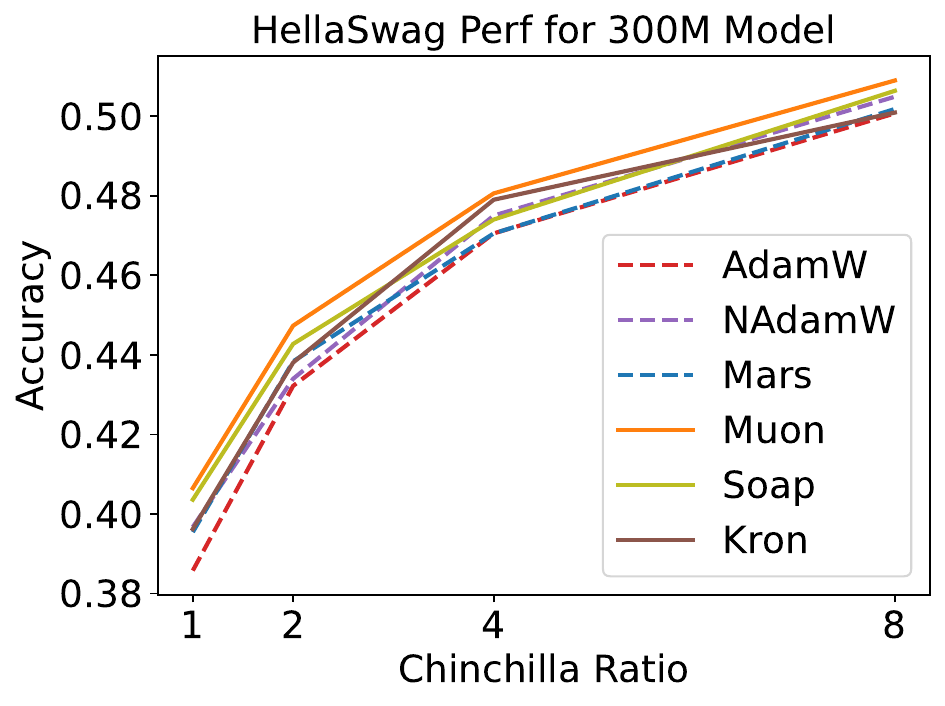}
    \end{minipage}
    \begin{minipage}{0.32\textwidth}
        \centering
        \includegraphics[width=\linewidth]{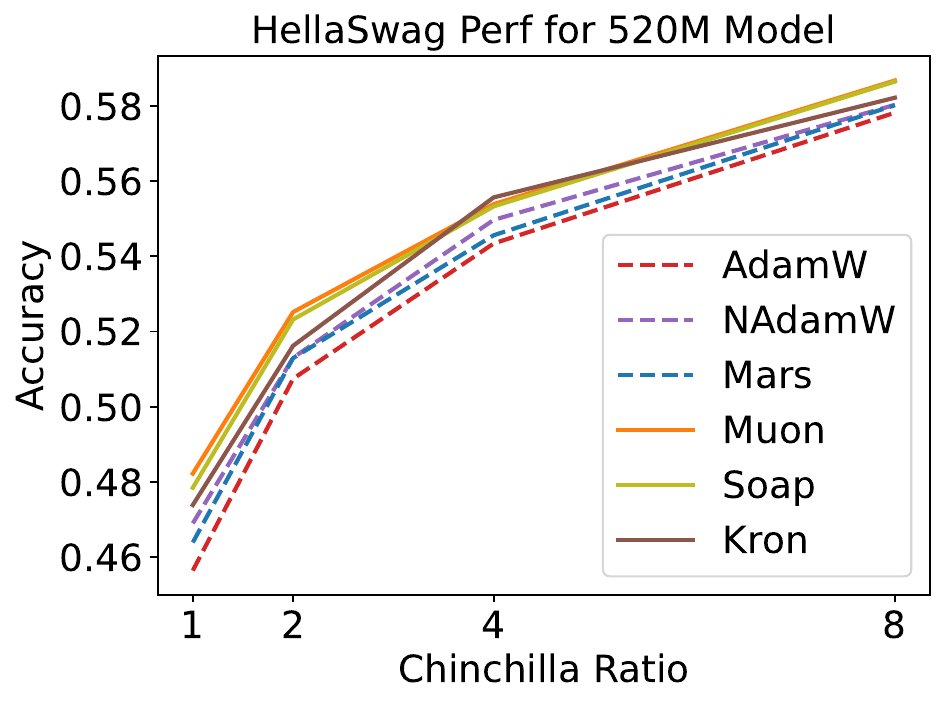}
    \end{minipage}
    \caption{\textbf{Main Results For Phase I \& II.} Top: We plot the validation loss on C4/EN for the experiments in Phase I and Phase II. Every point corresponds to the optimal loss achieved at the corresponding Chinchilla ratio for each optimizer. Bottom: we plot the HellaSwag performance corresponding to the selected run for a subset of optimizers: the AdamW baseline, the top 2 most performant scalar-based optimizers, and the top 3 most performant matrix-based optimizers. Analysis is deferred to~\Cref{sec:opt-comp-gen}.}
    \label{fig:main}
\end{figure}

\begin{figure}[t]
    \begin{minipage}{0.32\textwidth}
        \centering
        \includegraphics[width=\linewidth]{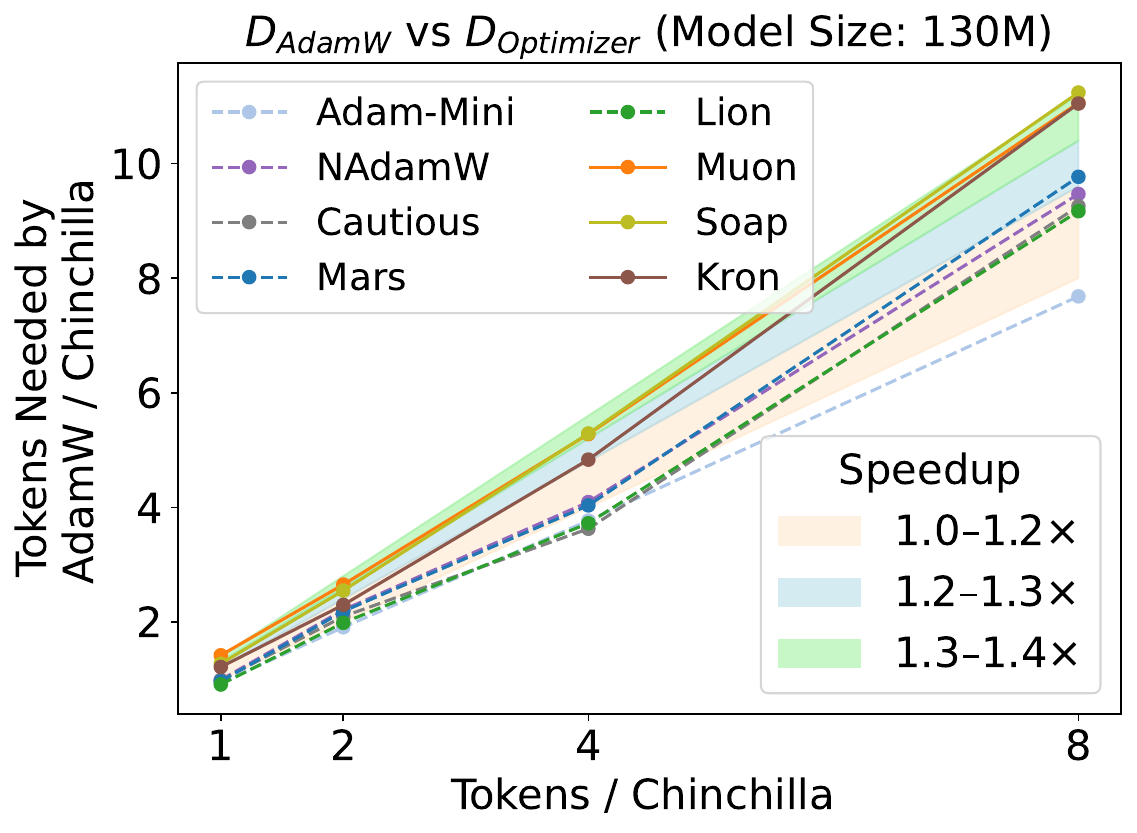}
    \end{minipage}
    \hfill
    \begin{minipage}{0.32\textwidth}
        \centering
        \includegraphics[width=\linewidth]{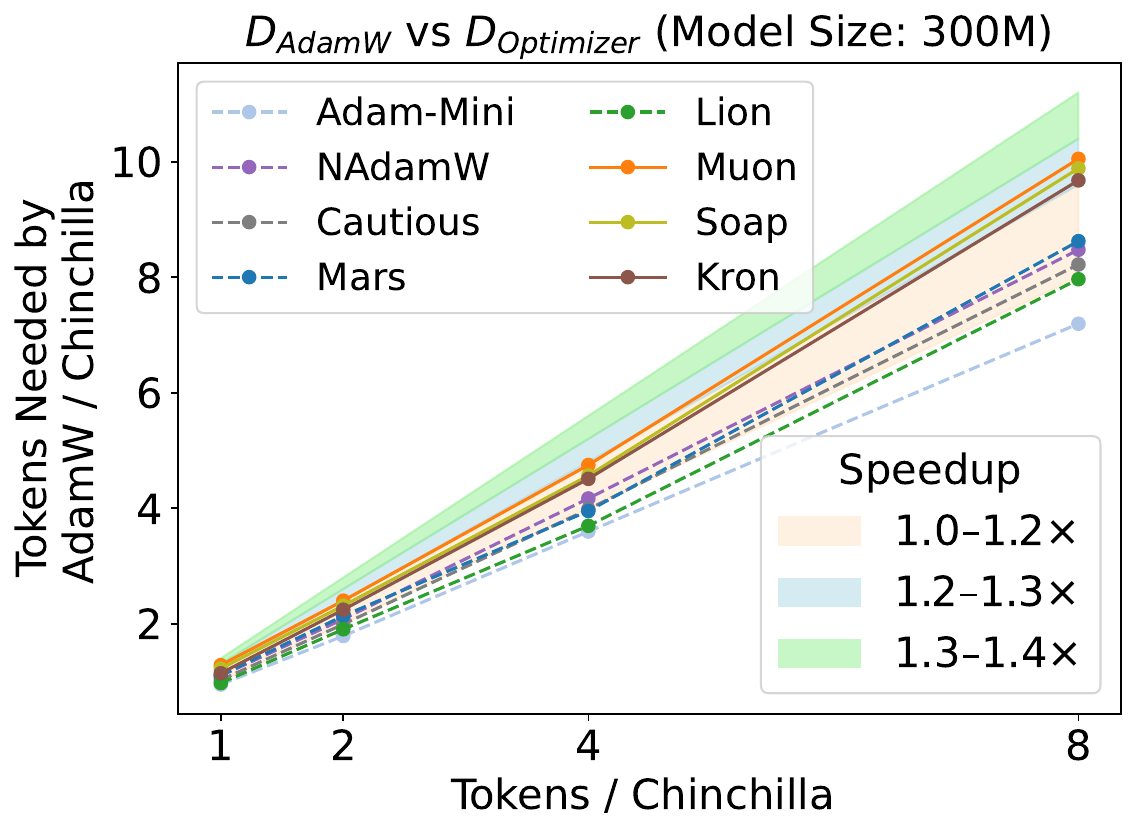}
    \end{minipage}
    \hfill
    \begin{minipage}{0.32\textwidth}
        \centering
        \includegraphics[width=\linewidth]{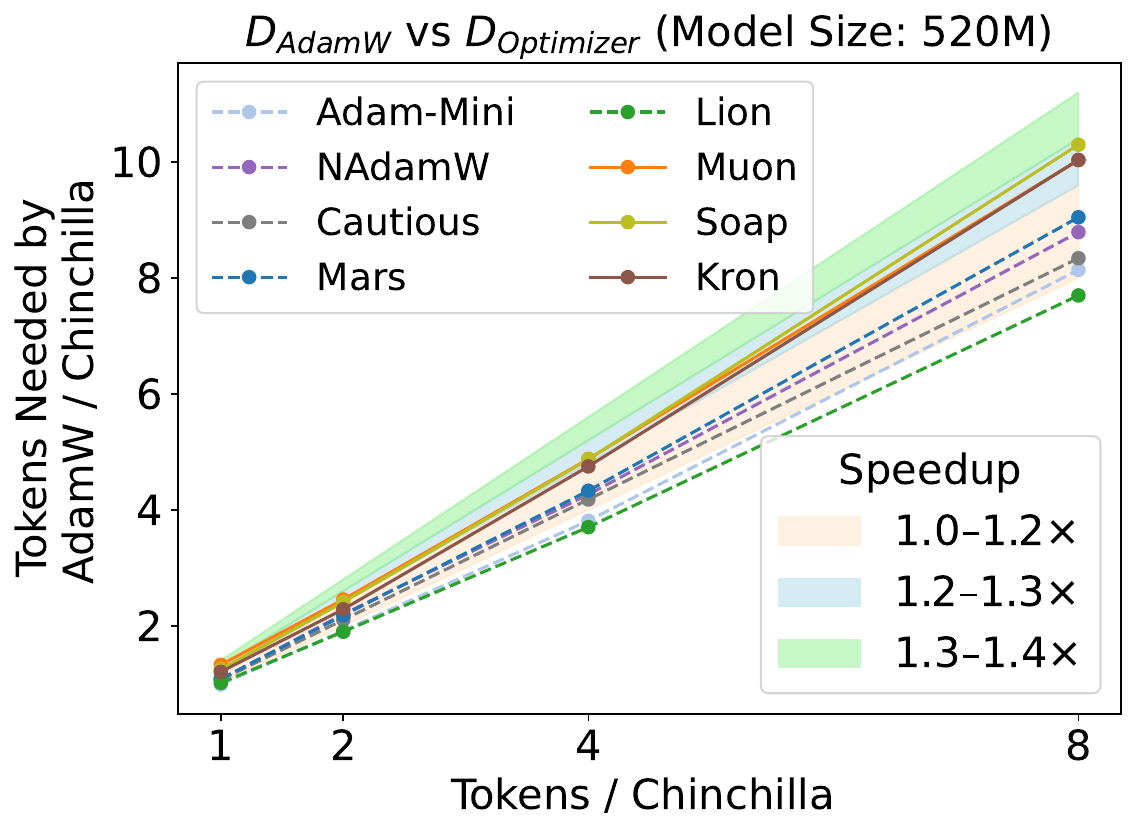}
    \end{minipage}
    \caption{\textbf{Speedup of different optimizers across scale.} We estimate the speedup of different optimizers by fitting a scaling law for AdamW and then map the loss of different optimizers to the corresponding equivalent data budget. We observe that (i) The highest speedup is capped at 1.4$\times$; (ii) matrix-based optimizers consistently outperform scalar-based optimizers and show an increasing speedup with data budget.}
    \label{fig:speedup}
\end{figure}

\subsection{Main Results}
\label{sec:opt-comp-gen}

\paragraph{Results on 0.1B–0.5B-parameter models.} 
Figure~\ref{fig:main} shows the validation loss curves for the 130M, 300M, and 520M models with varying Chinchilla ratios (1 to 8) in our benchmark. We further show that HellaSwag accuracy improvements closely mirror validation-loss gains. This is consistent with prior works that show lower losses translates to better downstream accuracy~\cite{bhagia2024establishingtaskscalinglaws,liu2025regmixdatamixtureregression}.
Across all model scales and compute budgets, both the variance-reduced Adam variants (NAdamW, Mars, Cautious) and the matrix-based optimizers deliver speedups over the AdamW baseline.  However, no method achieves $2\times$ step-wise acceleration claimed in previous literature. We note that Soap~\cite{vyas2025soapimprovingstabilizingshampoo} is one of the few works that conduct independent hyperparameter sweeping for the baseline, and it indeed reports a speedup closest to the actual observed improvement.
Following the methodology defined in~\Cref{sec:phase2}, we calculate the estimated speedup ratio of different optimizers in~\Cref{fig:speedup} and the highest speedup ratio is 1.4$\times$. 
From the measured speedups, three  patterns stand out in this computation regime:

\begin{enumerate}[leftmargin=*]
\item \textit{Matrix-based methods outperform scalar-based methods. The speedup ratio increases with data budget yet decreases with model size.}
For every model size, matrix-based optimizers (Soap, Muon, Kron, Scion—solid curves) consistently drive validation loss below that of their scalar-based counterparts (dashed curves). In the base (1$\times$ Chinchilla) compute regime, Muon performs best, but at 8× Chinchilla compute, the advantage shifts to Soap and Kron. As shown by the super-linear trend in Figure~\ref{fig:speedup}, the speedup ratios of these three optimizers grow with increasing data budget. To the best of authors' knowledge, this dependency on data budget is not noted in prior works, which typically only experiment on one data-to-model ratio.

However, the greatest gains occur on the 130M model, after which the speedup decreases to roughly 1.3× for larger model sizes. We will affirm this observation further in the experiments of 1.2B models, showing that the speedup ratios of matrix-based optimizers decrease to 1.1$\times$ for the 1.2B model.

\item \textit{Variance-reduction techniques provide a small but clear lift.}
Within the scalar-based family, all variance-reduced Adam variants ({NAdamW}, {Mars}, {Cautious}) consistently surpass vanilla AdamW---except for a small lag at the smallest experiment. Notably, {Muon} combines matrix-based updates with Nesterov momentum, illustrating how variance reduction compounds with matrix adaptation can yield greater efficiency. This result is vastly different from the $2\times$ speedup reported in some of the works and we attribute this disparity to the better-tuned baseline.

\item \textit{Memory-efficient variants of AdamW closely track the performance of AdamW.} 
The two memory-efficient AdamW variants ({Lion}, {Adam-mini})—despite their reduced auxiliary state—closely track the performance of AdamW, with a slowdown of at most $5\%$ and sometimes even perform better than AdamW. Interestingly, Lion and Adam-mini show different scaling trends regarding model size: the disadvantage of Lion relative to AdamW widens while the disadvantage of Adam-mini over AdamW narrows. 
\end{enumerate}

\begin{figure}
    \begin{minipage}{0.32\textwidth}
        \centering
        \includegraphics[width=\linewidth]{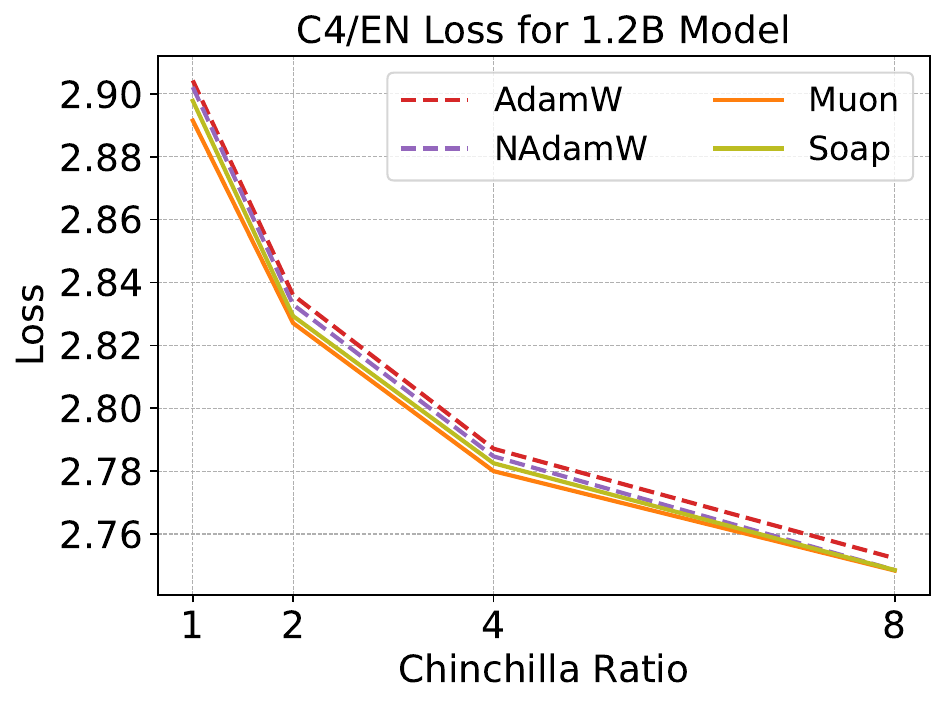}
    \end{minipage}
    \hfill
    \begin{minipage}{0.32\textwidth}
        \centering
        \includegraphics[width=\linewidth]{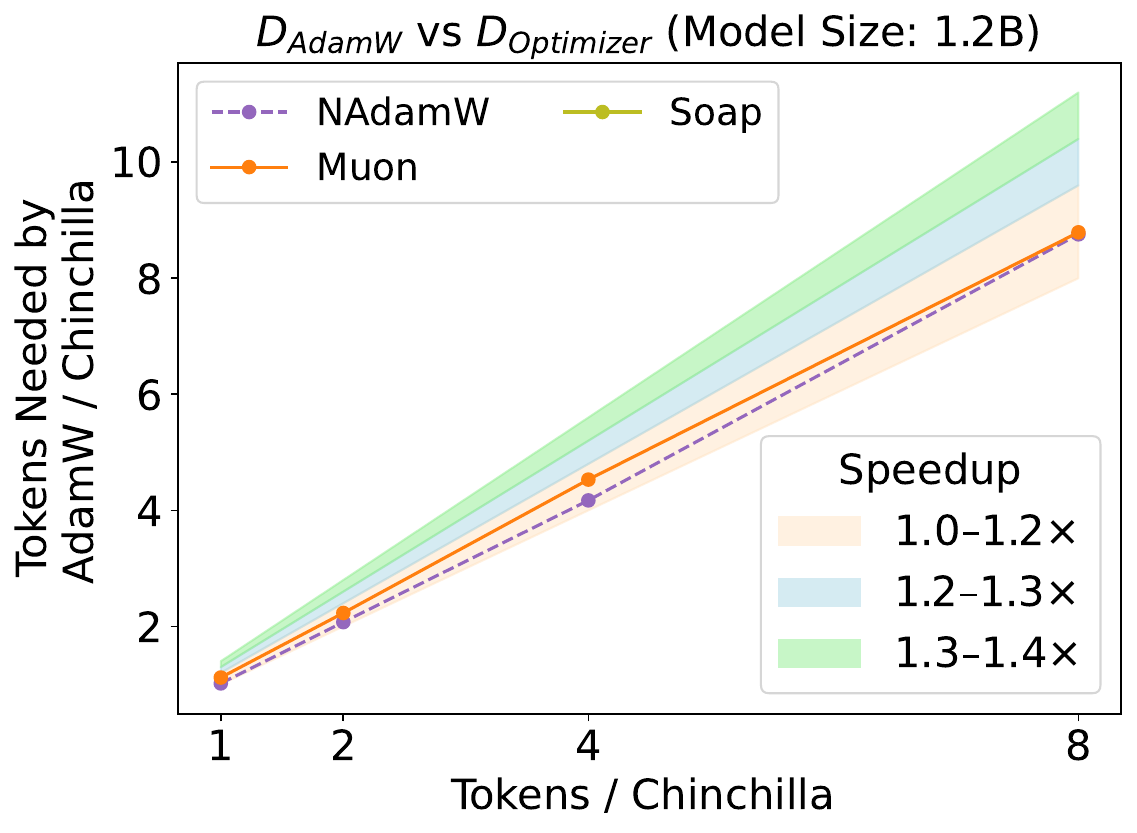}
    \end{minipage}
    \hfill
    \begin{minipage}{0.32\textwidth}
        \centering
        \includegraphics[width=\linewidth]{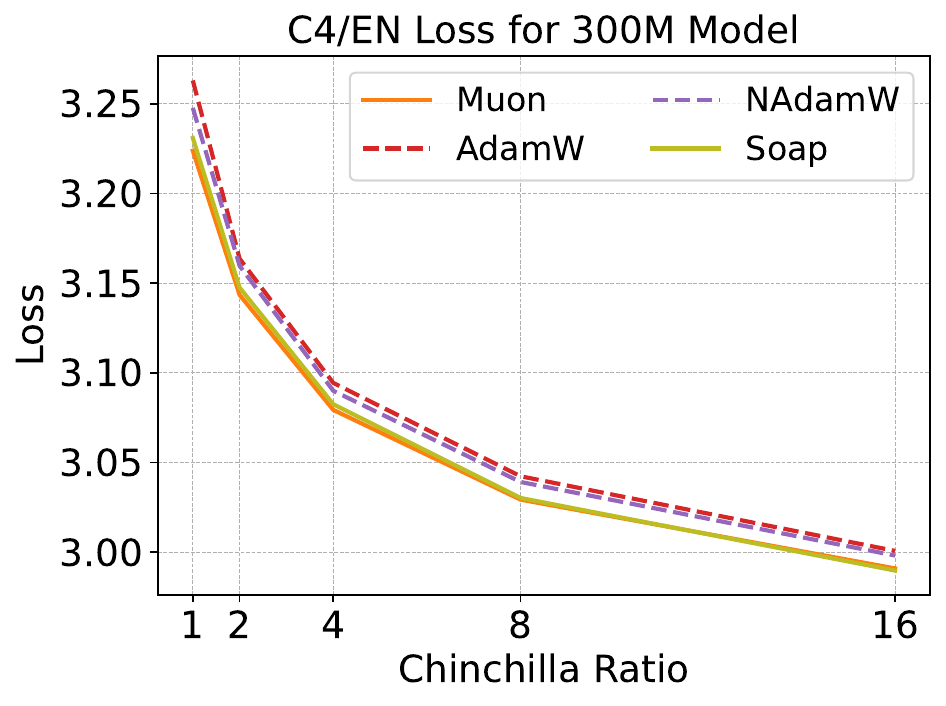}
    \end{minipage}
    \caption{\textbf{Case Studies.} Left: Validation loss scaling on 1.2B model for AdamW, NAdamW, Muon and Soap. Muon and Soap still offer significant speedup over AdamW but no longer significantly speed up over NAdamW. Mid: Estimated speedup ratio with the same methodology~\Cref{fig:speedup}, we observe that Muon and Soap's speedup decays with model size to only $1.1\times$. Last: Experiment with 300M $16 \times$ Chinchilla setting, Soap outperforms Muon when data-to-model ratio further increases.}
    \label{fig:case}
\end{figure}

\paragraph{Results on 1.2B-parameters Models.}
Using the hyperparameter scaling law we fit (\Cref{sec:phase2}), we scale up the model size to 1.2B to examine how the speedup of optimizers scales with model size. We observe that NAdamW, Muon, and Soap still deliver speedup over AdamW, but the speedup deminishes to $\approx$1.1$\times$ for all these optimizers (\Cref{fig:case}, Left and Mid) and no longer leads to downstream improvements (\Cref{tab:benchmark}). Based on this observation, we fit scaling laws for both AdamW and Muon based on the loss of 16 runs, and our scaling law predicts that Muon will result in a slightly higher loss than AdamW in the $7$B and $1\times$ Chinchilla regime. We defer the fitting procedure to~\Cref{app:scaling_law}.

\setlength{\tabcolsep}{3pt}
\begin{table}[h]
  \centering
  \caption{Benchmark performance of 1.2B models with different optimizers and Chinchilla scaling.}
  \label{tab:bench-1p2b}
  \small
  \begin{tabular}{ll*{10}{c}}
    \toprule
    Optimizer 
      & LAMBADA & OpenBook & Wino & PIQA & BoolQ 
      & WSC273 & Hella & ARC-C. & ARC-E & COPA & Avg \\
    \midrule
      \multicolumn{12}{c}{\textbf{8x Chinchilla (193B)}} \\
     AdamW  & 67.16 & 41.40 & 64.96 & 76.12 & 68.59 
      & 82.78 & 67.56 & 43.43 & 74.49 & 85.00 & 67.15 \\
     NAdamW & 67.84 & 40.20 & 64.80 & 77.15 & 68.10 
      & 83.52 & 67.40 & 43.34 & 73.61 & 81.00 & 66.70 \\
    Muon  & 67.53 & 39.80 & 67.09 & 77.09 & 68.81 
            & 80.95 & 67.67 & 43.34 & 73.53 & 84.00 & {66.98} \\
    \bottomrule
  \end{tabular}
  \label{tab:benchmark}
\end{table}

\textbf{High data-to-model Ratio.} In our previous experiments, Muon is outperformed by Soap in the 8$\times$ Chinchilla regime for the 130M and 520M models. To further test this, we train three 300M models to 16$\times$ Chinchilla and verify that Muon is no longer the optimal optimizer when the data-to-model ratio increases (\Cref{fig:case}, right). We conjecture that the second-order momentum maintained by Soap and Kron becomes more effective when the data-to-model ratio increases. In the long run, adaptivity to heterogeneity in parameter directions may lead to a larger speedup. We also perform similar experiments on 130M models and reach the same results (deferred to~\Cref{app:more_overtrained}).

\subsection{Necessity of Rigorous Benchmarking}

Our systematic sweeps uncover both universal optimization principles and surprising optimization-specific nuances, which call for rigorous study when designing future optimizers.

\begin{figure}[t]
    \begin{minipage}{0.48\textwidth}
        \centering
        \includegraphics[width=\linewidth]{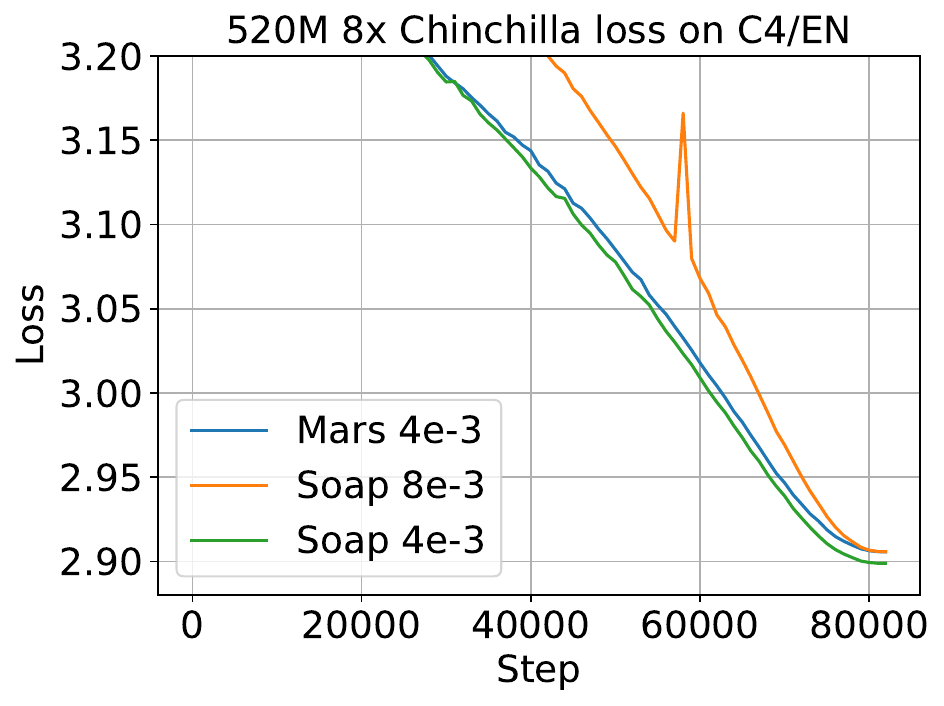}
    \end{minipage}
    \hfill
        \begin{minipage}{0.48\textwidth}
        \centering
        \includegraphics[width=\linewidth]{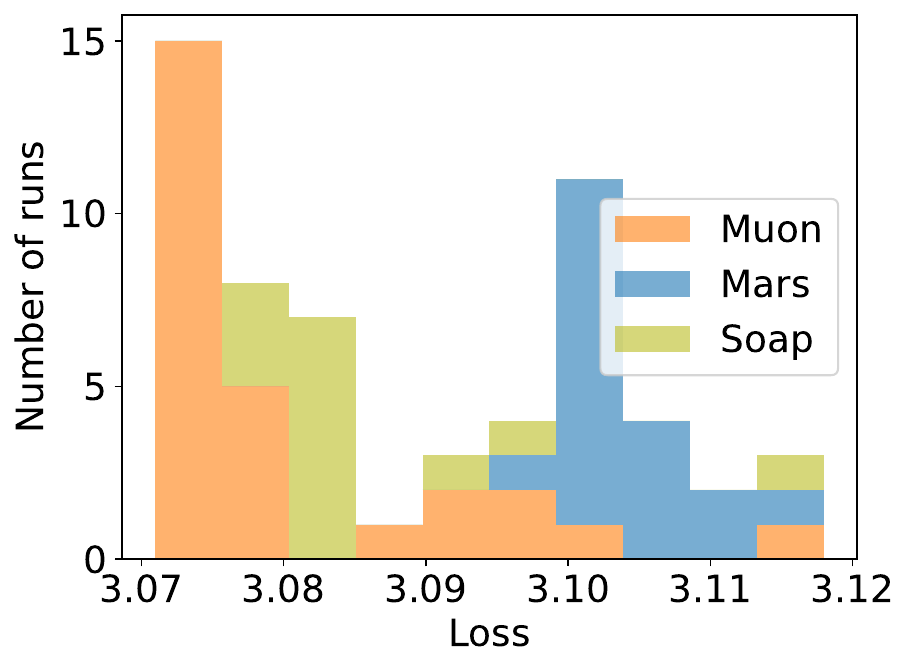}
    \end{minipage}
    \hfill
    \begin{minipage}{0.48\textwidth}
        \centering
        \includegraphics[width=\linewidth]{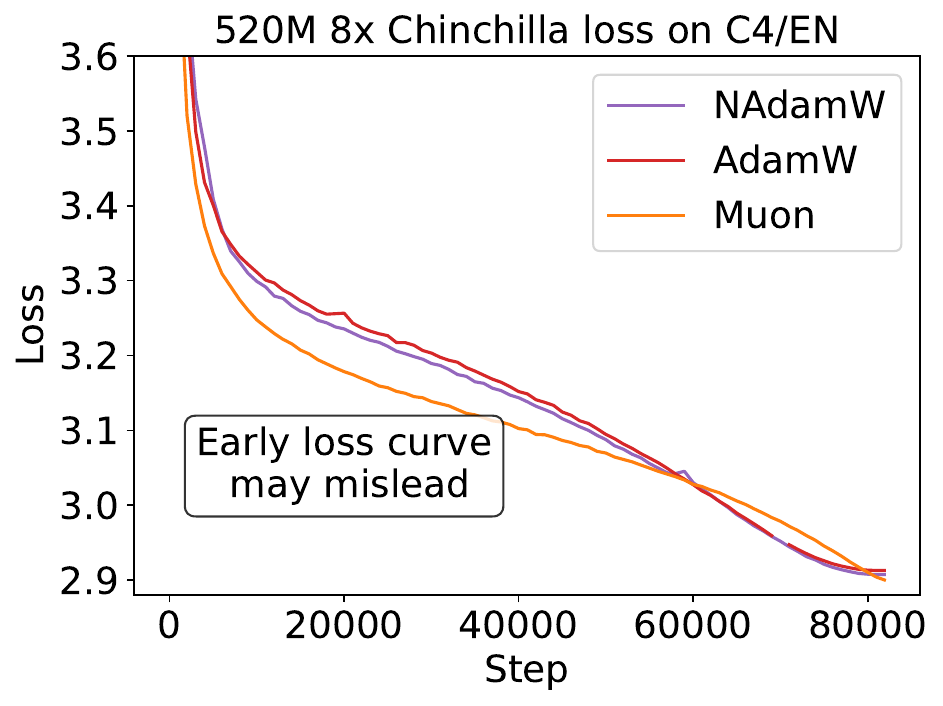}
    \end{minipage}%
    \hfill
    \begin{minipage}{0.48\textwidth}
        \centering
        \includegraphics[width=\linewidth]{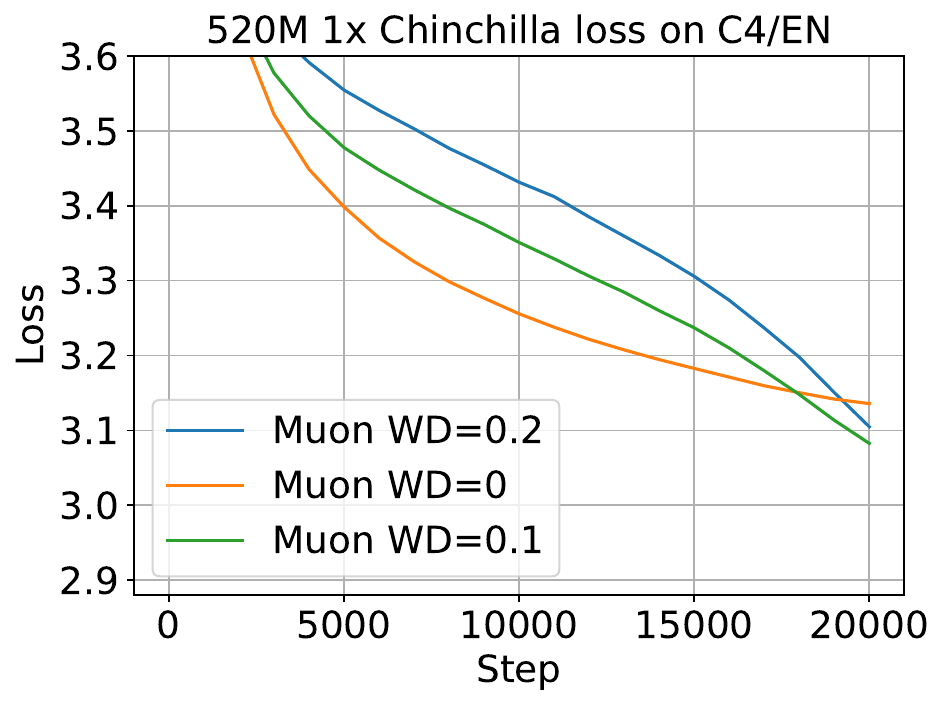}
    \end{minipage}
    
    \caption{\textbf{Necessity of Careful Tuning.} Left: 2$\times$ the optimal learning rate diminishes Soap's loss improvement over Mars on 520M 8x experiment; Mid: Variation of loss when only one hyperparameter differs from optimal learning rate and runs converge to within $0.02$ of optimal. The order of optimizers may flip arbitrarily if rigorous tuning is missing. Right: Changing a single hyperparameter like weight decay may lead to misleading faster loss improvement but plateaus later.}
    \label{fig:necessity}
\end{figure}

First, we find that even a superior optimizer can underperform a less advanced method when its hyperparameters are not precisely tuned. In our exhaustive grid searches, slight deviations from each optimizer's ideal learning rate or other critical hyperparameter often lead to degradation in validation loss that is large enough to flip the ordering (\Cref{fig:necessity}, Left). This hyperparameter sensitivity means manual selection without systematic sweeps will likely produce arbitrary ranking of the optimizers. To further demonstrate this, we plot how validation losses vary  when only \textbf{one} of the hyperparameters deviates from optimal value for {Muon}, {Soap}, {Mars} on 520M model in 1$\times$ Chinchilla regime (\Cref{fig:necessity}, Mid). The ordering of the optimizers can easily flip if one chooses a sub-optimal hyperparameter.

Second, using the same hyperparameters for different optimizers do not guarantee fair comparison between optimizers. For example, while weight decay is essential across all optimizers for optimal performance, the optimal decay strength varies markedly between optimizers. As shown in (\Cref{fig:motivation}, top right), optimal weight decay coefficients differ between the optimizers studied. We also note the optimal weight decay for Kron is larger than the conventional $0.1$ and is approximately $0.5$, which is crucial for Kron to outperform AdamW. 

Third, early training behavior can be highly misleading. Validation-loss curves during this initial phase tend to exaggerate performance gaps (\Cref{fig:motivation}, bottom right) and, in some cases, even reverse the eventual ranking, (\Cref{fig:necessity}, right). Many optimizers exhibit rapid early descent followed by plateauing, meaning that assessments based solely on losses before the end of the training  trajectories fail to predict final outcomes. To avoid such pitfalls, we recommend evaluating optimizers only on the final checkpoints rather than relying on intermediate checkpoints (e.g. in~\cite{liu2025muonscalablellmtraining}). 

We also note that the evaluaton in Sophia~\cite{liu2024sophiascalablestochasticsecondorder} largely follows the correct procedure of the comparisons above --- the peak learning rate was tuned to be optimal for the baseline. However, as the data loader in the codebase used did not fully randomize the order of the data, the optimal peak learning rate in that code base for AdamW is significantly smaller than the optimal peak learning rate for a fully randomized data loader setting. It turns out that Sophia doesn't offer significant speedup over AdamW for models under 0.5B in our setting (\Cref{fig:sophia_experiments}).

\begin{figure}
    \begin{minipage}{0.24\textwidth}
        \centering
        \includegraphics[width=\linewidth]{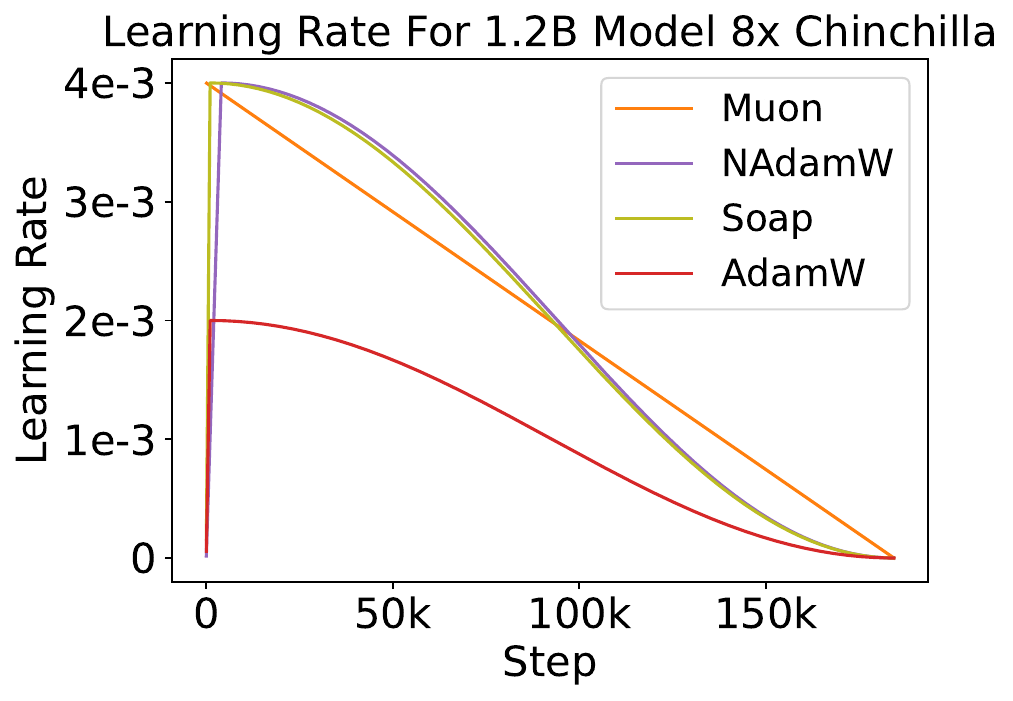}
    \end{minipage}
    \hfill
    \begin{minipage}{0.24\textwidth}
        \centering
        \includegraphics[width=\linewidth]{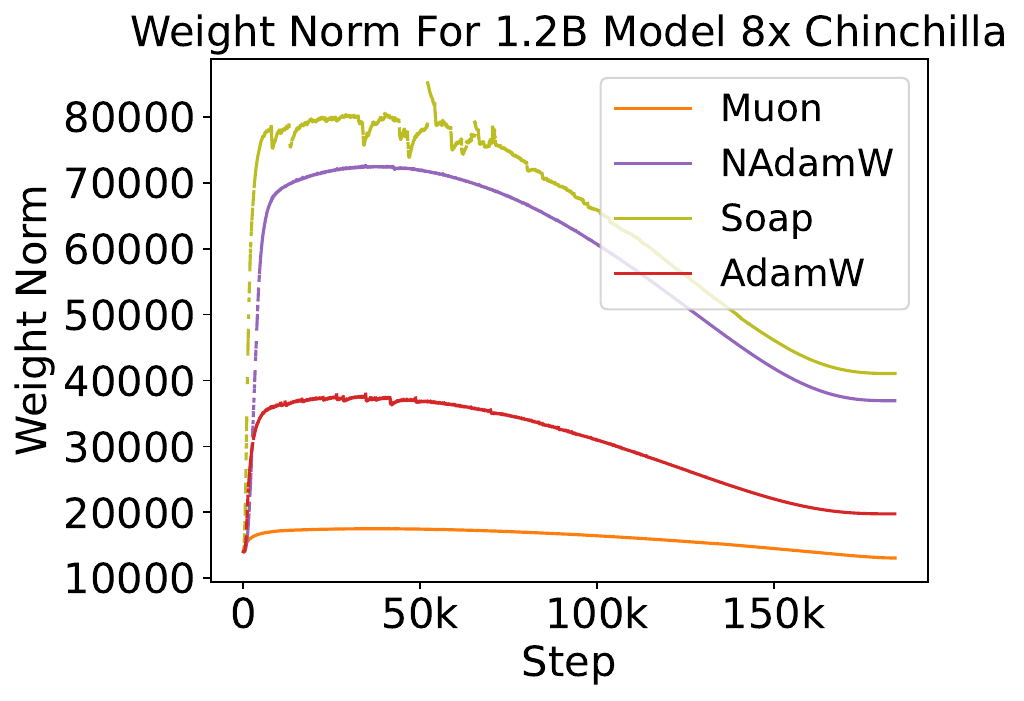}
    \end{minipage}
    \hfill
    \begin{minipage}{0.24\textwidth}
        \centering
        \includegraphics[width=\linewidth]{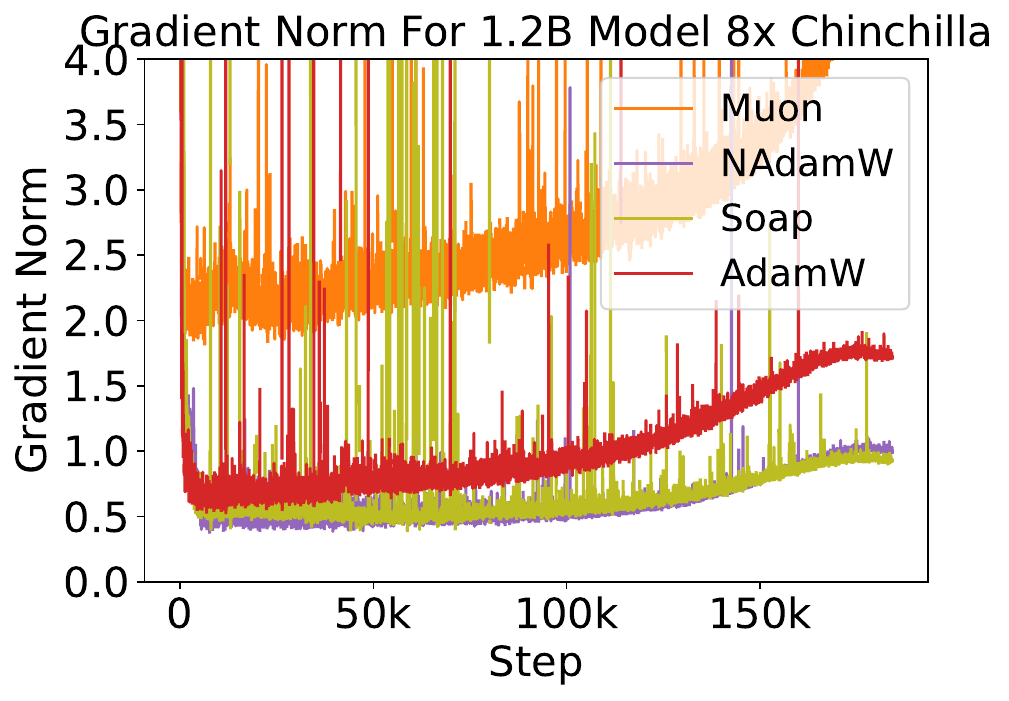}
    \end{minipage}
    \hfill
    \begin{minipage}{0.24\textwidth}
        \centering
        \includegraphics[width=\linewidth]{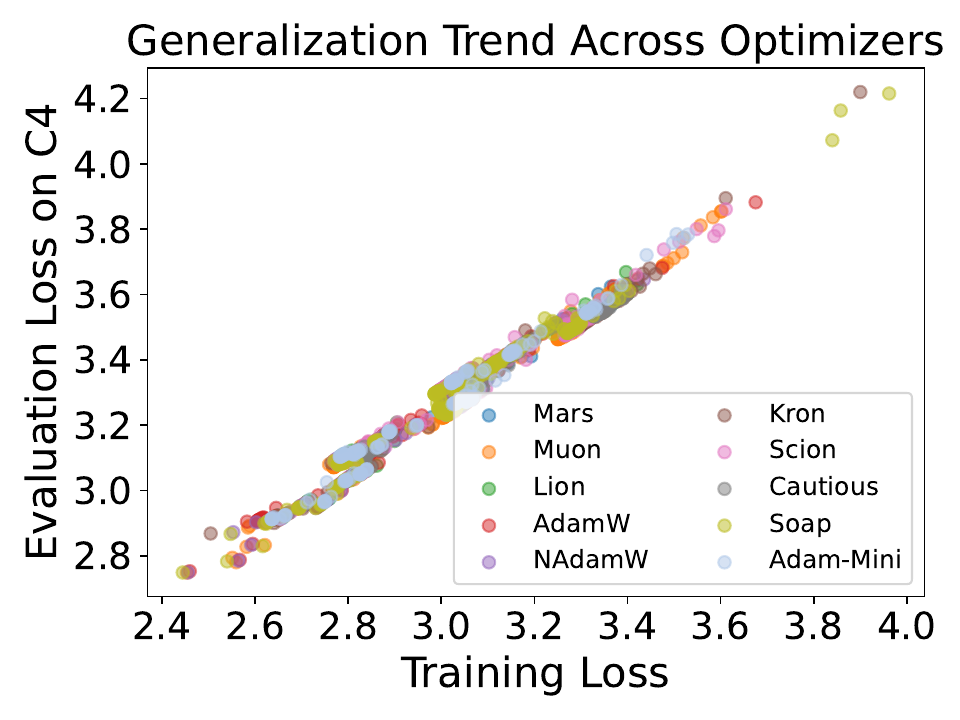}
    \end{minipage}
    \caption{\textbf{Common  Phenomena Across Optimizers.} 
    \textbf{Left}: Learning rate used for different optimizers.
    \textbf{Middle Left}: Parameter norm of all optimizers shows a similar trend of increment and decrease, closely aligning the increasing and decaying of learning rate schedule. \textbf{Middle Right}: Gradient norm increases during learning rate decay. However, this increase does not lead to a loss increase. \textbf{Right}: The training loss and evaluation loss follows the same trend for all optimizers.}
    \label{fig:common}
\end{figure}

\subsection{Common Phenomena Across Optimizers}

Through logging the evolution of weight and gradient norms, we discovered some shared optimization phenomena across optimizers.

\textbf{Parameter norms typically track learning rate decay when there is weight decay} We observe that the parameter norms of all optimizers show a similar pattern of increase and decrease, closely aligning with the increase and decrease of the learning rate if there is a decay phase. However, the absolute values of parameter norms are very different across optimizers (Figure~\ref{fig:common}, Middle Left).

\textbf{Gradient norm increases during learning rate decay.} Across all the optimizers, the gradient norms increase during the training run. However, this increase does not lead to a loss increase. Similar to the parameter norms, the absolute values of gradient norms are not consistent across optimizers.
These two phenomena are also reported in the previous work~\cite{defazio2025gradientsrapidlyincreasenear}, where the author also provides a theoretical explanation for both phenomena. We provide additional evidence that these two phenomena are not artifacts of the chosen optimizer AdamW but rather common phenomena across optimizers (Figure~\ref{fig:common}, Middle Right).

\textbf{Different optimizers have similar generalization behavior.} For architecture design, it has been observed that different architectures can have vastly different generalization behaviors~\cite{lu2025doessequencemodelingarchitecture}. However, this is not the case for optimizers, and the evaluation losses and training losses follow roughly the same trend for all optimizers at the regimes we experimented on (Figure~\ref{fig:common}, Right).

\section{Conclusion and Limitations}

We benchmarked 11 deep learning optimizers in pretraining and found that their true gains over AdamW are much smaller than previously reported. Our results highlight three key lessons: (i) many claimed speedups stem from insufficient hyperparameter tuning, as fair sweeps eliminate most of the apparent advantages; (ii) comparisons based on early or inconsistent evaluation can be misleading, since optimizer rankings often change over the full training trajectory; and (iii) even the best-performing alternatives provide only modest speedups, which further diminish with model scale, dropping to ~1.1× at 1.2B parameters. This benchmarking study has the limitation that it does we haven't scaled to models larger than 1.2B parameters. However, we believe that evaluating optimizers on models of comparable size to prior studies is still valuable, as it reveals that insufficient tuning is a major cause of subpar speedups. Promising future work includes extending our benchmarking to larger models beyond 1.2B parameters to test whether the diminishing speedup trend persists at frontier scales. Another direction is to design optimizers whose efficiency remains stable under scaling laws, ensuring consistent benefits as model size and data budget grow.

\section*{Acknowledgement}

This work was supported by the Google TPU Research Cloud (TRC), the Stanford HAI–Google Cloud Credits Program, the DARPA AIQ program, and NSF RI 2045685, and is a part of the Marin Project. The authors thank Evan Walters, Omead Pooladzandi, Jiacheng You, and Zhiyuan Li for discussions and help during our research.

\newpage
% Bibliography
\bibliographystyle{abbrvnat}
\bibliography{references}  

\begin{thebibliography}{114}
\providecommand{\natexlab}[1]{#1}
\providecommand{\url}[1]{\texttt{#1}}
\expandafter\ifx\csname urlstyle\endcsname\relax
  \providecommand{\doi}[1]{doi: #1}\else
  \providecommand{\doi}{doi: \begingroup \urlstyle{rm}\Url}\fi

\bibitem[Agarwal et~al.(2020)Agarwal, Anil, Hazan, Koren, and Zhang]{agarwal2020disentanglingadaptivegradientmethods}
N.~Agarwal, R.~Anil, E.~Hazan, T.~Koren, and C.~Zhang.
\newblock Disentangling adaptive gradient methods from learning rates, 2020.
\newblock URL \url{https://arxiv.org/abs/2002.11803}.

\bibitem[Ahn et~al.(2025)Ahn, Xu, Abreu, and Langford]{ahn2025diondistributedorthonormalizedupdates}
K.~Ahn, B.~Xu, N.~Abreu, and J.~Langford.
\newblock Dion: Distributed orthonormalized updates, 2025.
\newblock URL \url{https://arxiv.org/abs/2504.05295}.

\bibitem[AI et~al.(2025)AI, :, Shah, Polloreno, Stratos, Monk, Chaluvaraju, Hojel, Ma, Thomas, Tanwer, Shah, Nguyen, Smith, Callahan, Pust, Parmar, Rushton, Mazarakis, Kapila, Srivastava, Singla, Romanski, Vanjani, and Vaswani]{ai2025practicalefficiencymuonpretraining}
E.~AI, :, I.~Shah, A.~M. Polloreno, K.~Stratos, P.~Monk, A.~Chaluvaraju, A.~Hojel, A.~Ma, A.~Thomas, A.~Tanwer, D.~J. Shah, K.~Nguyen, K.~Smith, M.~Callahan, M.~Pust, M.~Parmar, P.~Rushton, P.~Mazarakis, R.~Kapila, S.~Srivastava, S.~Singla, T.~Romanski, Y.~Vanjani, and A.~Vaswani.
\newblock Practical efficiency of muon for pretraining, 2025.
\newblock URL \url{https://arxiv.org/abs/2505.02222}.

\bibitem[Anil et~al.(2021)Anil, Gupta, Koren, Regan, and Singer]{anil2021scalablesecondorderoptimization}
R.~Anil, V.~Gupta, T.~Koren, K.~Regan, and Y.~Singer.
\newblock Scalable second order optimization for deep learning, 2021.
\newblock URL \url{https://arxiv.org/abs/2002.09018}.

\bibitem[Azerbayev et~al.(2024)Azerbayev, Schoelkopf, Paster, Santos, McAleer, Jiang, Deng, Biderman, and Welleck]{azerbayev2024llemmaopenlanguagemodel}
Z.~Azerbayev, H.~Schoelkopf, K.~Paster, M.~D. Santos, S.~McAleer, A.~Q. Jiang, J.~Deng, S.~Biderman, and S.~Welleck.
\newblock Llemma: An open language model for mathematics, 2024.
\newblock URL \url{https://arxiv.org/abs/2310.10631}.

\bibitem[Becker and Cun(1989)]{becker:improving}
S.~Becker and Y.~L. Cun.
\newblock Improving the convergence of back-propagation learning with second order methods.
\newblock In D.~S. Touretzky, G.~E. Hinton, and T.~J. Sejnowski, editors, \emph{{P}roceedings of the 1988 Connectionist Models Summer School}, pages 29--37. San Francisco, CA: Morgan Kaufmann, 1989.

\bibitem[Bernstein and Newhouse(2024{\natexlab{a}})]{bernstein2024modulardualitydeeplearning}
J.~Bernstein and L.~Newhouse.
\newblock Modular duality in deep learning, 2024{\natexlab{a}}.
\newblock URL \url{https://arxiv.org/abs/2410.21265}.

\bibitem[Bernstein and Newhouse(2024{\natexlab{b}})]{bernstein2024oldoptimizernewnorm}
J.~Bernstein and L.~Newhouse.
\newblock Old optimizer, new norm: An anthology, 2024{\natexlab{b}}.
\newblock URL \url{https://arxiv.org/abs/2409.20325}.

\bibitem[Bernstein et~al.(2018)Bernstein, Wang, Azizzadenesheli, and Anandkumar]{pmlr-v80-bernstein18a}
J.~Bernstein, Y.-X. Wang, K.~Azizzadenesheli, and A.~Anandkumar.
\newblock sign{SGD}: Compressed optimisation for non-convex problems.
\newblock In J.~Dy and A.~Krause, editors, \emph{Proceedings of the 35th International Conference on Machine Learning}, volume~80 of \emph{Proceedings of Machine Learning Research}, pages 560--569. PMLR, 10--15 Jul 2018.
\newblock URL \url{https://proceedings.mlr.press/v80/bernstein18a.html}.

\bibitem[Bhagia et~al.(2024)Bhagia, Liu, Wettig, Heineman, Tafjord, Jha, Soldaini, Smith, Groeneveld, Koh, Dodge, and Hajishirzi]{bhagia2024establishingtaskscalinglaws}
A.~Bhagia, J.~Liu, A.~Wettig, D.~Heineman, O.~Tafjord, A.~H. Jha, L.~Soldaini, N.~A. Smith, D.~Groeneveld, P.~W. Koh, J.~Dodge, and H.~Hajishirzi.
\newblock Establishing task scaling laws via compute-efficient model ladders, 2024.
\newblock URL \url{https://arxiv.org/abs/2412.04403}.

\bibitem[Bisk et~al.(2020)Bisk, Zellers, Le~Bras, Gao, and Choi]{Bisk2020PIQA}
Y.~Bisk, R.~Zellers, R.~Le~Bras, J.~Gao, and Y.~Choi.
\newblock Piqa: Reasoning about physical commonsense in natural language.
\newblock In \emph{Proceedings of the 34th AAAI Conference on Artificial Intelligence (AAAI)}, pages 7432--7439. AAAI Press, 2020.
\newblock \doi{10.1609/aaai.v34i05.6239}.
\newblock URL \url{https://doi.org/10.1609/aaai.v34i05.6239}.

\bibitem[Brown et~al.(2020)Brown, Mann, Ryder, Subbiah, Kaplan, Dhariwal, Neelakantan, Shyam, Sastry, Askell, Agarwal, Herbert-Voss, Krueger, Henighan, Child, Ramesh, Ziegler, Wu, Winter, Hesse, Chen, Sigler, Litwin, Gray, Chess, Clark, Berner, McCandlish, Radford, Sutskever, and Amodei]{brown2020languagemodelsfewshotlearners}
T.~B. Brown, B.~Mann, N.~Ryder, M.~Subbiah, J.~Kaplan, P.~Dhariwal, A.~Neelakantan, P.~Shyam, G.~Sastry, A.~Askell, S.~Agarwal, A.~Herbert-Voss, G.~Krueger, T.~Henighan, R.~Child, A.~Ramesh, D.~M. Ziegler, J.~Wu, C.~Winter, C.~Hesse, M.~Chen, E.~Sigler, M.~Litwin, S.~Gray, B.~Chess, J.~Clark, C.~Berner, S.~McCandlish, A.~Radford, I.~Sutskever, and D.~Amodei.
\newblock Language models are few-shot learners, 2020.
\newblock URL \url{https://arxiv.org/abs/2005.14165}.

\bibitem[Chen et~al.(2023)Chen, Liang, Huang, Real, Wang, Liu, Pham, Dong, Luong, Hsieh, Lu, and Le]{chen2023symbolic}
X.~Chen, C.~Liang, D.~Huang, E.~Real, K.~Wang, Y.~Liu, H.~Pham, X.~Dong, T.~Luong, C.-J. Hsieh, Y.~Lu, and Q.~V. Le.
\newblock Symbolic discovery of optimization algorithms, 2023.
\newblock URL \url{https://arxiv.org/abs/2302.06675}.

\bibitem[Clark et~al.(2019)Clark, Lee, Chang, Kwiatkowski, Collins, and Toutanova]{Clark2019BoolQ}
C.~Clark, K.~Lee, M.-W. Chang, T.~Kwiatkowski, M.~Collins, and K.~Toutanova.
\newblock Boolq: Exploring the surprising difficulty of natural yes/no questions.
\newblock In J.~Burstein, C.~Doran, and T.~Solorio, editors, \emph{Proceedings of NAACL-HLT 2019}, pages 2924--2936, Minneapolis, Minnesota, June 2019. Association for Computational Linguistics.
\newblock \doi{10.18653/v1/N19-1300}.
\newblock URL \url{https://aclanthology.org/N19-1300/}.

\bibitem[Clark et~al.(2018)Clark, Cowhey, Etzioni, Khot, Sabharwal, Schoenick, and Tafjord]{Clark2018ARC}
P.~Clark, I.~Cowhey, O.~Etzioni, T.~Khot, A.~Sabharwal, C.~Schoenick, and O.~Tafjord.
\newblock Think you have solved question answering? try arc, the ai2 reasoning challenge.
\newblock \emph{CoRR}, abs/1803.05457, 2018.
\newblock URL \url{http://arxiv.org/abs/1803.05457}.

\bibitem[DeepSeek-AI et~al.(2024)DeepSeek-AI, :, Bi, Chen, Chen, Chen, Dai, Deng, Ding, Dong, Du, Fu, Gao, Gao, Gao, Ge, Guan, Guo, Guo, Hao, Hao, He, Hu, Huang, Li, Li, Li, Li, Li, Liang, Lin, Liu, Liu, Liu, Liu, Liu, Liu, Lu, Lu, Luo, Ma, Nie, Pei, Piao, Qiu, Qu, Ren, Ren, Ruan, Sha, Shao, Song, Su, Sun, Sun, Tang, Wang, Wang, Wang, Wang, Wang, Wu, Wu, Xie, Xie, Xie, Xiong, Xu, Xu, Xu, Yang, You, Yu, Yu, Zhang, Zhang, Zhang, Zhang, Zhang, Zhang, Zhang, Zhang, Zhao, Zhao, Zhou, Zhou, Zhu, and Zou]{deepseekai2024deepseekllmscalingopensource}
DeepSeek-AI, :, X.~Bi, D.~Chen, G.~Chen, S.~Chen, D.~Dai, C.~Deng, H.~Ding, K.~Dong, Q.~Du, Z.~Fu, H.~Gao, K.~Gao, W.~Gao, R.~Ge, K.~Guan, D.~Guo, J.~Guo, G.~Hao, Z.~Hao, Y.~He, W.~Hu, P.~Huang, E.~Li, G.~Li, J.~Li, Y.~Li, Y.~K. Li, W.~Liang, F.~Lin, A.~X. Liu, B.~Liu, W.~Liu, X.~Liu, X.~Liu, Y.~Liu, H.~Lu, S.~Lu, F.~Luo, S.~Ma, X.~Nie, T.~Pei, Y.~Piao, J.~Qiu, H.~Qu, T.~Ren, Z.~Ren, C.~Ruan, Z.~Sha, Z.~Shao, J.~Song, X.~Su, J.~Sun, Y.~Sun, M.~Tang, B.~Wang, P.~Wang, S.~Wang, Y.~Wang, Y.~Wang, T.~Wu, Y.~Wu, X.~Xie, Z.~Xie, Z.~Xie, Y.~Xiong, H.~Xu, R.~X. Xu, Y.~Xu, D.~Yang, Y.~You, S.~Yu, X.~Yu, B.~Zhang, H.~Zhang, L.~Zhang, L.~Zhang, M.~Zhang, M.~Zhang, W.~Zhang, Y.~Zhang, C.~Zhao, Y.~Zhao, S.~Zhou, S.~Zhou, Q.~Zhu, and Y.~Zou.
\newblock Deepseek llm: Scaling open-source language models with longtermism, 2024.
\newblock URL \url{https://arxiv.org/abs/2401.02954}.

\bibitem[DeepSeek-AI et~al.(2025{\natexlab{a}})DeepSeek-AI, Guo, Yang, Zhang, Song, Zhang, Xu, Zhu, Ma, Wang, Bi, Zhang, Yu, Wu, Wu, Gou, Shao, Li, Gao, Liu, Xue, Wang, Wu, Feng, Lu, Zhao, Deng, Zhang, Ruan, Dai, Chen, Ji, Li, Lin, Dai, Luo, Hao, Chen, Li, Zhang, Bao, Xu, Wang, Ding, Xin, Gao, Qu, Li, Guo, Li, Wang, Chen, Yuan, Qiu, Li, Cai, Ni, Liang, Chen, Dong, Hu, Gao, Guan, Huang, Yu, Wang, Zhang, Zhao, Wang, Zhang, Xu, Xia, Zhang, Zhang, Tang, Li, Wang, Li, Tian, Huang, Zhang, Wang, Chen, Du, Ge, Zhang, Pan, Wang, Chen, Jin, Chen, Lu, Zhou, Chen, Ye, Wang, Yu, Zhou, Pan, Li, Zhou, Wu, Ye, Yun, Pei, Sun, Wang, Zeng, Zhao, Liu, Liang, Gao, Yu, Zhang, Xiao, An, Liu, Wang, Chen, Nie, Cheng, Liu, Xie, Liu, Yang, Li, Su, Lin, Li, Jin, Shen, Chen, Sun, Wang, Song, Zhou, Wang, Shan, Li, Wang, Wei, Zhang, Xu, Li, Zhao, Sun, Wang, Yu, Zhang, Shi, Xiong, He, Piao, Wang, Tan, Ma, Liu, Guo, Ou, Wang, Gong, Zou, He, Xiong, Luo, You, Liu, Zhou, Zhu, Xu, Huang, Li, Zheng, Zhu, Ma, Tang, Zha, Yan, Ren, Ren, Sha, Fu, Xu,
  Xie, Zhang, Hao, Ma, Yan, Wu, Gu, Zhu, Liu, Li, Xie, Song, Pan, Huang, Xu, Zhang, and Zhang]{deepseekai2025deepseekr1incentivizingreasoningcapability}
DeepSeek-AI, D.~Guo, D.~Yang, H.~Zhang, J.~Song, R.~Zhang, R.~Xu, Q.~Zhu, S.~Ma, P.~Wang, X.~Bi, X.~Zhang, X.~Yu, Y.~Wu, Z.~F. Wu, Z.~Gou, Z.~Shao, Z.~Li, Z.~Gao, A.~Liu, B.~Xue, B.~Wang, B.~Wu, B.~Feng, C.~Lu, C.~Zhao, C.~Deng, C.~Zhang, C.~Ruan, D.~Dai, D.~Chen, D.~Ji, E.~Li, F.~Lin, F.~Dai, F.~Luo, G.~Hao, G.~Chen, G.~Li, H.~Zhang, H.~Bao, H.~Xu, H.~Wang, H.~Ding, H.~Xin, H.~Gao, H.~Qu, H.~Li, J.~Guo, J.~Li, J.~Wang, J.~Chen, J.~Yuan, J.~Qiu, J.~Li, J.~L. Cai, J.~Ni, J.~Liang, J.~Chen, K.~Dong, K.~Hu, K.~Gao, K.~Guan, K.~Huang, K.~Yu, L.~Wang, L.~Zhang, L.~Zhao, L.~Wang, L.~Zhang, L.~Xu, L.~Xia, M.~Zhang, M.~Zhang, M.~Tang, M.~Li, M.~Wang, M.~Li, N.~Tian, P.~Huang, P.~Zhang, Q.~Wang, Q.~Chen, Q.~Du, R.~Ge, R.~Zhang, R.~Pan, R.~Wang, R.~J. Chen, R.~L. Jin, R.~Chen, S.~Lu, S.~Zhou, S.~Chen, S.~Ye, S.~Wang, S.~Yu, S.~Zhou, S.~Pan, S.~S. Li, S.~Zhou, S.~Wu, S.~Ye, T.~Yun, T.~Pei, T.~Sun, T.~Wang, W.~Zeng, W.~Zhao, W.~Liu, W.~Liang, W.~Gao, W.~Yu, W.~Zhang, W.~L. Xiao, W.~An, X.~Liu, X.~Wang, X.~Chen, X.~Nie,
  X.~Cheng, X.~Liu, X.~Xie, X.~Liu, X.~Yang, X.~Li, X.~Su, X.~Lin, X.~Q. Li, X.~Jin, X.~Shen, X.~Chen, X.~Sun, X.~Wang, X.~Song, X.~Zhou, X.~Wang, X.~Shan, Y.~K. Li, Y.~Q. Wang, Y.~X. Wei, Y.~Zhang, Y.~Xu, Y.~Li, Y.~Zhao, Y.~Sun, Y.~Wang, Y.~Yu, Y.~Zhang, Y.~Shi, Y.~Xiong, Y.~He, Y.~Piao, Y.~Wang, Y.~Tan, Y.~Ma, Y.~Liu, Y.~Guo, Y.~Ou, Y.~Wang, Y.~Gong, Y.~Zou, Y.~He, Y.~Xiong, Y.~Luo, Y.~You, Y.~Liu, Y.~Zhou, Y.~X. Zhu, Y.~Xu, Y.~Huang, Y.~Li, Y.~Zheng, Y.~Zhu, Y.~Ma, Y.~Tang, Y.~Zha, Y.~Yan, Z.~Z. Ren, Z.~Ren, Z.~Sha, Z.~Fu, Z.~Xu, Z.~Xie, Z.~Zhang, Z.~Hao, Z.~Ma, Z.~Yan, Z.~Wu, Z.~Gu, Z.~Zhu, Z.~Liu, Z.~Li, Z.~Xie, Z.~Song, Z.~Pan, Z.~Huang, Z.~Xu, Z.~Zhang, and Z.~Zhang.
\newblock Deepseek-r1: Incentivizing reasoning capability in llms via reinforcement learning, 2025{\natexlab{a}}.
\newblock URL \url{https://arxiv.org/abs/2501.12948}.

\bibitem[DeepSeek-AI et~al.(2025{\natexlab{b}})DeepSeek-AI, Liu, Feng, Xue, Wang, Wu, Lu, Zhao, Deng, Zhang, Ruan, Dai, Guo, Yang, Chen, Ji, Li, Lin, Dai, Luo, Hao, Chen, Li, Zhang, Bao, Xu, Wang, Zhang, Ding, Xin, Gao, Li, Qu, Cai, Liang, Guo, Ni, Li, Wang, Chen, Chen, Yuan, Qiu, Li, Song, Dong, Hu, Gao, Guan, Huang, Yu, Wang, Zhang, Xu, Xia, Zhao, Wang, Zhang, Li, Wang, Zhang, Zhang, Tang, Li, Tian, Huang, Wang, Zhang, Wang, Zhu, Chen, Du, Chen, Jin, Ge, Zhang, Pan, Wang, Xu, Zhang, Chen, Li, Lu, Zhou, Chen, Wu, Ye, Ye, Ma, Wang, Zhou, Yu, Zhou, Pan, Wang, Yun, Pei, Sun, Xiao, Zeng, Zhao, An, Liu, Liang, Gao, Yu, Zhang, Li, Jin, Wang, Bi, Liu, Wang, Shen, Chen, Zhang, Chen, Nie, Sun, Wang, Cheng, Liu, Xie, Liu, Yu, Song, Shan, Zhou, Yang, Li, Su, Lin, Li, Wang, Wei, Zhu, Zhang, Xu, Xu, Huang, Li, Zhao, Sun, Li, Wang, Yu, Zheng, Zhang, Shi, Xiong, He, Tang, Piao, Wang, Tan, Ma, Liu, Guo, Wu, Ou, Zhu, Wang, Gong, Zou, He, Zha, Xiong, Ma, Yan, Luo, You, Liu, Zhou, Wu, Ren, Ren, Sha, Fu, Xu, Huang, Zhang, Xie,
  Zhang, Hao, Gou, Ma, Yan, Shao, Xu, Wu, Zhang, Li, Gu, Zhu, Liu, Li, Xie, Song, Gao, and Pan]{deepseekai2025deepseekv3technicalreport}
DeepSeek-AI, A.~Liu, B.~Feng, B.~Xue, B.~Wang, B.~Wu, C.~Lu, C.~Zhao, C.~Deng, C.~Zhang, C.~Ruan, D.~Dai, D.~Guo, D.~Yang, D.~Chen, D.~Ji, E.~Li, F.~Lin, F.~Dai, F.~Luo, G.~Hao, G.~Chen, G.~Li, H.~Zhang, H.~Bao, H.~Xu, H.~Wang, H.~Zhang, H.~Ding, H.~Xin, H.~Gao, H.~Li, H.~Qu, J.~L. Cai, J.~Liang, J.~Guo, J.~Ni, J.~Li, J.~Wang, J.~Chen, J.~Chen, J.~Yuan, J.~Qiu, J.~Li, J.~Song, K.~Dong, K.~Hu, K.~Gao, K.~Guan, K.~Huang, K.~Yu, L.~Wang, L.~Zhang, L.~Xu, L.~Xia, L.~Zhao, L.~Wang, L.~Zhang, M.~Li, M.~Wang, M.~Zhang, M.~Zhang, M.~Tang, M.~Li, N.~Tian, P.~Huang, P.~Wang, P.~Zhang, Q.~Wang, Q.~Zhu, Q.~Chen, Q.~Du, R.~J. Chen, R.~L. Jin, R.~Ge, R.~Zhang, R.~Pan, R.~Wang, R.~Xu, R.~Zhang, R.~Chen, S.~S. Li, S.~Lu, S.~Zhou, S.~Chen, S.~Wu, S.~Ye, S.~Ye, S.~Ma, S.~Wang, S.~Zhou, S.~Yu, S.~Zhou, S.~Pan, T.~Wang, T.~Yun, T.~Pei, T.~Sun, W.~L. Xiao, W.~Zeng, W.~Zhao, W.~An, W.~Liu, W.~Liang, W.~Gao, W.~Yu, W.~Zhang, X.~Q. Li, X.~Jin, X.~Wang, X.~Bi, X.~Liu, X.~Wang, X.~Shen, X.~Chen, X.~Zhang, X.~Chen, X.~Nie, X.~Sun,
  X.~Wang, X.~Cheng, X.~Liu, X.~Xie, X.~Liu, X.~Yu, X.~Song, X.~Shan, X.~Zhou, X.~Yang, X.~Li, X.~Su, X.~Lin, Y.~K. Li, Y.~Q. Wang, Y.~X. Wei, Y.~X. Zhu, Y.~Zhang, Y.~Xu, Y.~Xu, Y.~Huang, Y.~Li, Y.~Zhao, Y.~Sun, Y.~Li, Y.~Wang, Y.~Yu, Y.~Zheng, Y.~Zhang, Y.~Shi, Y.~Xiong, Y.~He, Y.~Tang, Y.~Piao, Y.~Wang, Y.~Tan, Y.~Ma, Y.~Liu, Y.~Guo, Y.~Wu, Y.~Ou, Y.~Zhu, Y.~Wang, Y.~Gong, Y.~Zou, Y.~He, Y.~Zha, Y.~Xiong, Y.~Ma, Y.~Yan, Y.~Luo, Y.~You, Y.~Liu, Y.~Zhou, Z.~F. Wu, Z.~Z. Ren, Z.~Ren, Z.~Sha, Z.~Fu, Z.~Xu, Z.~Huang, Z.~Zhang, Z.~Xie, Z.~Zhang, Z.~Hao, Z.~Gou, Z.~Ma, Z.~Yan, Z.~Shao, Z.~Xu, Z.~Wu, Z.~Zhang, Z.~Li, Z.~Gu, Z.~Zhu, Z.~Liu, Z.~Li, Z.~Xie, Z.~Song, Z.~Gao, and Z.~Pan.
\newblock Deepseek-v3 technical report, 2025{\natexlab{b}}.
\newblock URL \url{https://arxiv.org/abs/2412.19437}.

\bibitem[Defazio(2025)]{defazio2025gradientsrapidlyincreasenear}
A.~Defazio.
\newblock Why gradients rapidly increase near the end of training, 2025.
\newblock URL \url{https://arxiv.org/abs/2506.02285}.

\bibitem[Defazio and Mishchenko(2023)]{pmlr-v202-defazio23a}
A.~Defazio and K.~Mishchenko.
\newblock Learning-rate-free learning by d-adaptation.
\newblock In A.~Krause, E.~Brunskill, K.~Cho, B.~Engelhardt, S.~Sabato, and J.~Scarlett, editors, \emph{Proceedings of the 40th International Conference on Machine Learning}, volume 202 of \emph{Proceedings of Machine Learning Research}, pages 7449--7479. PMLR, 23--29 Jul 2023.
\newblock URL \url{https://proceedings.mlr.press/v202/defazio23a.html}.

\bibitem[Defazio et~al.(2024{\natexlab{a}})Defazio, Cutkosky, Mehta, and Mishchenko]{defazio2024optimallineardecaylearning}
A.~Defazio, A.~Cutkosky, H.~Mehta, and K.~Mishchenko.
\newblock Optimal linear decay learning rate schedules and further refinements, 2024{\natexlab{a}}.
\newblock URL \url{https://arxiv.org/abs/2310.07831}.

\bibitem[Defazio et~al.(2024{\natexlab{b}})Defazio, Yang, Mehta, Mishchenko, Khaled, and Cutkosky]{defazio2024roadscheduled}
A.~Defazio, X.~A. Yang, H.~Mehta, K.~Mishchenko, A.~Khaled, and A.~Cutkosky.
\newblock The road less scheduled, 2024{\natexlab{b}}.
\newblock URL \url{https://arxiv.org/abs/2405.15682}.

\bibitem[Dozat(2016)]{dozat2016incorporating}
T.~Dozat.
\newblock Incorporating nesterov momentum into adam.
\newblock 2016.

\bibitem[Duchi et~al.(2011)Duchi, Hazan, and Singer]{duchi2011adaptive}
J.~Duchi, E.~Hazan, and Y.~Singer.
\newblock Adaptive subgradient methods for online learning and stochastic optimization.
\newblock \emph{Journal of machine learning research}, 12\penalty0 (7), 2011.

\bibitem[Eschenhagen et~al.(2023)Eschenhagen, Immer, Turner, Schneider, and Hennig]{eschenhagen2023kronecker}
R.~Eschenhagen, A.~Immer, R.~Turner, F.~Schneider, and P.~Hennig.
\newblock Kronecker-factored approximate curvature for modern neural network architectures.
\newblock \emph{Advances in Neural Information Processing Systems}, 36:\penalty0 33624--33655, 2023.

\bibitem[Eschenhagen et~al.(2025)Eschenhagen, Defazio, Lee, Turner, and Shi]{eschenhagen2025purifyingshampooinvestigatingshampoos}
R.~Eschenhagen, A.~Defazio, T.-H. Lee, R.~E. Turner, and H.-J.~M. Shi.
\newblock Purifying shampoo: Investigating shampoo's heuristics by decomposing its preconditioner, 2025.
\newblock URL \url{https://arxiv.org/abs/2506.03595}.

\bibitem[Everett et~al.(2024)Everett, Xiao, Wortsman, Alemi, Novak, Liu, Gur, Sohl-Dickstein, Kaelbling, Lee, and Pennington]{everett2024scalingexponentsparameterizationsoptimizers}
K.~Everett, L.~Xiao, M.~Wortsman, A.~A. Alemi, R.~Novak, P.~J. Liu, I.~Gur, J.~Sohl-Dickstein, L.~P. Kaelbling, J.~Lee, and J.~Pennington.
\newblock Scaling exponents across parameterizations and optimizers, 2024.
\newblock URL \url{https://arxiv.org/abs/2407.05872}.

\bibitem[Foret et~al.(2021)Foret, Kleiner, Mobahi, and Neyshabur]{foret2021sharpnessawareminimizationefficientlyimproving}
P.~Foret, A.~Kleiner, H.~Mobahi, and B.~Neyshabur.
\newblock Sharpness-aware minimization for efficiently improving generalization, 2021.
\newblock URL \url{https://arxiv.org/abs/2010.01412}.

\bibitem[Frans et~al.(2025)Frans, Levine, and Abbeel]{frans2025stablewhiteningoptimizerefficient}
K.~Frans, S.~Levine, and P.~Abbeel.
\newblock A stable whitening optimizer for efficient neural network training, 2025.
\newblock URL \url{https://arxiv.org/abs/2506.07254}.

\bibitem[Gordon et~al.(2012)Gordon, Kozareva, and Roemmele]{Gordon2012COPA}
A.~Gordon, Z.~Kozareva, and M.~Roemmele.
\newblock Semeval-2012 task 7: Choice of plausible alternatives: An evaluation of commonsense causal reasoning.
\newblock In E.~Agirre, J.~Bos, M.~Diab, S.~Manandhar, Y.~Marton, and D.~Yuret, editors, \emph{*SEM 2012: The First Joint Conference on Lexical and Computational Semantics (SemEval 2012)}, pages 394--398, Montréal, Canada, June 7--8 2012. Association for Computational Linguistics.
\newblock URL \url{https://aclanthology.org/S12-1052/}.

\bibitem[Grattafiori et~al.(2024)Grattafiori, Dubey, Jauhri, Pandey, Kadian, Al-Dahle, Letman, Mathur, Schelten, Vaughan, Yang, Fan, Goyal, Hartshorn, Yang, Mitra, Sravankumar, Korenev, Hinsvark, Rao, Zhang, Rodriguez, Gregerson, Spataru, Roziere, Biron, Tang, Chern, Caucheteux, Nayak, Bi, Marra, McConnell, Keller, Touret, Wu, Wong, Ferrer, Nikolaidis, Allonsius, Song, Pintz, Livshits, Wyatt, Esiobu, Choudhary, Mahajan, Garcia-Olano, Perino, Hupkes, Lakomkin, AlBadawy, Lobanova, Dinan, Smith, Radenovic, Guzmán, Zhang, Synnaeve, Lee, Anderson, Thattai, Nail, Mialon, Pang, Cucurell, Nguyen, Korevaar, Xu, Touvron, Zarov, Ibarra, Kloumann, Misra, Evtimov, Zhang, Copet, Lee, Geffert, Vranes, Park, Mahadeokar, Shah, van~der Linde, Billock, Hong, Lee, Fu, Chi, Huang, Liu, Wang, Yu, Bitton, Spisak, Park, Rocca, Johnstun, Saxe, Jia, Alwala, Prasad, Upasani, Plawiak, Li, Heafield, Stone, El-Arini, Iyer, Malik, Chiu, Bhalla, Lakhotia, Rantala-Yeary, van~der Maaten, Chen, Tan, Jenkins, Martin, Madaan, Malo, Blecher,
  Landzaat, de~Oliveira, Muzzi, Pasupuleti, Singh, Paluri, Kardas, Tsimpoukelli, Oldham, Rita, Pavlova, Kambadur, Lewis, Si, Singh, Hassan, Goyal, Torabi, Bashlykov, Bogoychev, Chatterji, Zhang, Duchenne, Çelebi, Alrassy, Zhang, Li, Vasic, Weng, Bhargava, Dubal, Krishnan, Koura, Xu, He, Dong, Srinivasan, Ganapathy, Calderer, Cabral, Stojnic, Raileanu, Maheswari, Girdhar, Patel, Sauvestre, Polidoro, Sumbaly, Taylor, Silva, Hou, Wang, Hosseini, Chennabasappa, Singh, Bell, Kim, Edunov, Nie, Narang, Raparthy, Shen, Wan, Bhosale, Zhang, Vandenhende, Batra, Whitman, Sootla, Collot, Gururangan, Borodinsky, Herman, Fowler, Sheasha, Georgiou, Scialom, Speckbacher, Mihaylov, Xiao, Karn, Goswami, Gupta, Ramanathan, Kerkez, Gonguet, Do, Vogeti, Albiero, Petrovic, Chu, Xiong, Fu, Meers, Martinet, Wang, Wang, Tan, Xia, Xie, Jia, Wang, Goldschlag, Gaur, Babaei, Wen, Song, Zhang, Li, Mao, Coudert, Yan, Chen, Papakipos, Singh, Srivastava, Jain, Kelsey, Shajnfeld, Gangidi, Victoria, Goldstand, Menon, Sharma, Boesenberg,
  Baevski, Feinstein, Kallet, Sangani, Teo, Yunus, Lupu, Alvarado, Caples, Gu, Ho, Poulton, Ryan, Ramchandani, Dong, Franco, Goyal, Saraf, Chowdhury, Gabriel, Bharambe, Eisenman, Yazdan, James, Maurer, Leonhardi, Huang, Loyd, Paola, Paranjape, Liu, Wu, Ni, Hancock, Wasti, Spence, Stojkovic, Gamido, Montalvo, Parker, Burton, Mejia, Liu, Wang, Kim, Zhou, Hu, Chu, Cai, Tindal, Feichtenhofer, Gao, Civin, Beaty, Kreymer, Li, Adkins, Xu, Testuggine, David, Parikh, Liskovich, Foss, Wang, Le, Holland, Dowling, Jamil, Montgomery, Presani, Hahn, Wood, Le, Brinkman, Arcaute, Dunbar, Smothers, Sun, Kreuk, Tian, Kokkinos, Ozgenel, Caggioni, Kanayet, Seide, Florez, Schwarz, Badeer, Swee, Halpern, Herman, Sizov, Guangyi, Zhang, Lakshminarayanan, Inan, Shojanazeri, Zou, Wang, Zha, Habeeb, Rudolph, Suk, Aspegren, Goldman, Zhan, Damlaj, Molybog, Tufanov, Leontiadis, Veliche, Gat, Weissman, Geboski, Kohli, Lam, Asher, Gaya, Marcus, Tang, Chan, Zhen, Reizenstein, Teboul, Zhong, Jin, Yang, Cummings, Carvill, Shepard, McPhie,
  Torres, Ginsburg, Wang, Wu, U, Saxena, Khandelwal, Zand, Matosich, Veeraraghavan, Michelena, Li, Jagadeesh, Huang, Chawla, Huang, Chen, Garg, A, Silva, Bell, Zhang, Guo, Yu, Moshkovich, Wehrstedt, Khabsa, Avalani, Bhatt, Mankus, Hasson, Lennie, Reso, Groshev, Naumov, Lathi, Keneally, Liu, Seltzer, Valko, Restrepo, Patel, Vyatskov, Samvelyan, Clark, Macey, Wang, Hermoso, Metanat, Rastegari, Bansal, Santhanam, Parks, White, Bawa, Singhal, Egebo, Usunier, Mehta, Laptev, Dong, Cheng, Chernoguz, Hart, Salpekar, Kalinli, Kent, Parekh, Saab, Balaji, Rittner, Bontrager, Roux, Dollar, Zvyagina, Ratanchandani, Yuvraj, Liang, Alao, Rodriguez, Ayub, Murthy, Nayani, Mitra, Parthasarathy, Li, Hogan, Battey, Wang, Howes, Rinott, Mehta, Siby, Bondu, Datta, Chugh, Hunt, Dhillon, Sidorov, Pan, Mahajan, Verma, Yamamoto, Ramaswamy, Lindsay, Lindsay, Feng, Lin, Zha, Patil, Shankar, Zhang, Zhang, Wang, Agarwal, Sajuyigbe, Chintala, Max, Chen, Kehoe, Satterfield, Govindaprasad, Gupta, Deng, Cho, Virk, Subramanian, Choudhury,
  Goldman, Remez, Glaser, Best, Koehler, Robinson, Li, Zhang, Matthews, Chou, Shaked, Vontimitta, Ajayi, Montanez, Mohan, Kumar, Mangla, Ionescu, Poenaru, Mihailescu, Ivanov, Li, Wang, Jiang, Bouaziz, Constable, Tang, Wu, Wang, Wu, Gao, Kleinman, Chen, Hu, Jia, Qi, Li, Zhang, Zhang, Adi, Nam, Yu, Wang, Zhao, Hao, Qian, Li, He, Rait, DeVito, Rosnbrick, Wen, Yang, Zhao, and Ma]{grattafiori2024llama3herdmodels}
A.~Grattafiori, A.~Dubey, A.~Jauhri, A.~Pandey, A.~Kadian, A.~Al-Dahle, A.~Letman, A.~Mathur, A.~Schelten, A.~Vaughan, A.~Yang, A.~Fan, A.~Goyal, A.~Hartshorn, A.~Yang, A.~Mitra, A.~Sravankumar, A.~Korenev, A.~Hinsvark, A.~Rao, A.~Zhang, A.~Rodriguez, A.~Gregerson, A.~Spataru, B.~Roziere, B.~Biron, B.~Tang, B.~Chern, C.~Caucheteux, C.~Nayak, C.~Bi, C.~Marra, C.~McConnell, C.~Keller, C.~Touret, C.~Wu, C.~Wong, C.~C. Ferrer, C.~Nikolaidis, D.~Allonsius, D.~Song, D.~Pintz, D.~Livshits, D.~Wyatt, D.~Esiobu, D.~Choudhary, D.~Mahajan, D.~Garcia-Olano, D.~Perino, D.~Hupkes, E.~Lakomkin, E.~AlBadawy, E.~Lobanova, E.~Dinan, E.~M. Smith, F.~Radenovic, F.~Guzmán, F.~Zhang, G.~Synnaeve, G.~Lee, G.~L. Anderson, G.~Thattai, G.~Nail, G.~Mialon, G.~Pang, G.~Cucurell, H.~Nguyen, H.~Korevaar, H.~Xu, H.~Touvron, I.~Zarov, I.~A. Ibarra, I.~Kloumann, I.~Misra, I.~Evtimov, J.~Zhang, J.~Copet, J.~Lee, J.~Geffert, J.~Vranes, J.~Park, J.~Mahadeokar, J.~Shah, J.~van~der Linde, J.~Billock, J.~Hong, J.~Lee, J.~Fu, J.~Chi, J.~Huang,
  J.~Liu, J.~Wang, J.~Yu, J.~Bitton, J.~Spisak, J.~Park, J.~Rocca, J.~Johnstun, J.~Saxe, J.~Jia, K.~V. Alwala, K.~Prasad, K.~Upasani, K.~Plawiak, K.~Li, K.~Heafield, K.~Stone, K.~El-Arini, K.~Iyer, K.~Malik, K.~Chiu, K.~Bhalla, K.~Lakhotia, L.~Rantala-Yeary, L.~van~der Maaten, L.~Chen, L.~Tan, L.~Jenkins, L.~Martin, L.~Madaan, L.~Malo, L.~Blecher, L.~Landzaat, L.~de~Oliveira, M.~Muzzi, M.~Pasupuleti, M.~Singh, M.~Paluri, M.~Kardas, M.~Tsimpoukelli, M.~Oldham, M.~Rita, M.~Pavlova, M.~Kambadur, M.~Lewis, M.~Si, M.~K. Singh, M.~Hassan, N.~Goyal, N.~Torabi, N.~Bashlykov, N.~Bogoychev, N.~Chatterji, N.~Zhang, O.~Duchenne, O.~Çelebi, P.~Alrassy, P.~Zhang, P.~Li, P.~Vasic, P.~Weng, P.~Bhargava, P.~Dubal, P.~Krishnan, P.~S. Koura, P.~Xu, Q.~He, Q.~Dong, R.~Srinivasan, R.~Ganapathy, R.~Calderer, R.~S. Cabral, R.~Stojnic, R.~Raileanu, R.~Maheswari, R.~Girdhar, R.~Patel, R.~Sauvestre, R.~Polidoro, R.~Sumbaly, R.~Taylor, R.~Silva, R.~Hou, R.~Wang, S.~Hosseini, S.~Chennabasappa, S.~Singh, S.~Bell, S.~S. Kim, S.~Edunov,
  S.~Nie, S.~Narang, S.~Raparthy, S.~Shen, S.~Wan, S.~Bhosale, S.~Zhang, S.~Vandenhende, S.~Batra, S.~Whitman, S.~Sootla, S.~Collot, S.~Gururangan, S.~Borodinsky, T.~Herman, T.~Fowler, T.~Sheasha, T.~Georgiou, T.~Scialom, T.~Speckbacher, T.~Mihaylov, T.~Xiao, U.~Karn, V.~Goswami, V.~Gupta, V.~Ramanathan, V.~Kerkez, V.~Gonguet, V.~Do, V.~Vogeti, V.~Albiero, V.~Petrovic, W.~Chu, W.~Xiong, W.~Fu, W.~Meers, X.~Martinet, X.~Wang, X.~Wang, X.~E. Tan, X.~Xia, X.~Xie, X.~Jia, X.~Wang, Y.~Goldschlag, Y.~Gaur, Y.~Babaei, Y.~Wen, Y.~Song, Y.~Zhang, Y.~Li, Y.~Mao, Z.~D. Coudert, Z.~Yan, Z.~Chen, Z.~Papakipos, A.~Singh, A.~Srivastava, A.~Jain, A.~Kelsey, A.~Shajnfeld, A.~Gangidi, A.~Victoria, A.~Goldstand, A.~Menon, A.~Sharma, A.~Boesenberg, A.~Baevski, A.~Feinstein, A.~Kallet, A.~Sangani, A.~Teo, A.~Yunus, A.~Lupu, A.~Alvarado, A.~Caples, A.~Gu, A.~Ho, A.~Poulton, A.~Ryan, A.~Ramchandani, A.~Dong, A.~Franco, A.~Goyal, A.~Saraf, A.~Chowdhury, A.~Gabriel, A.~Bharambe, A.~Eisenman, A.~Yazdan, B.~James, B.~Maurer,
  B.~Leonhardi, B.~Huang, B.~Loyd, B.~D. Paola, B.~Paranjape, B.~Liu, B.~Wu, B.~Ni, B.~Hancock, B.~Wasti, B.~Spence, B.~Stojkovic, B.~Gamido, B.~Montalvo, C.~Parker, C.~Burton, C.~Mejia, C.~Liu, C.~Wang, C.~Kim, C.~Zhou, C.~Hu, C.-H. Chu, C.~Cai, C.~Tindal, C.~Feichtenhofer, C.~Gao, D.~Civin, D.~Beaty, D.~Kreymer, D.~Li, D.~Adkins, D.~Xu, D.~Testuggine, D.~David, D.~Parikh, D.~Liskovich, D.~Foss, D.~Wang, D.~Le, D.~Holland, E.~Dowling, E.~Jamil, E.~Montgomery, E.~Presani, E.~Hahn, E.~Wood, E.-T. Le, E.~Brinkman, E.~Arcaute, E.~Dunbar, E.~Smothers, F.~Sun, F.~Kreuk, F.~Tian, F.~Kokkinos, F.~Ozgenel, F.~Caggioni, F.~Kanayet, F.~Seide, G.~M. Florez, G.~Schwarz, G.~Badeer, G.~Swee, G.~Halpern, G.~Herman, G.~Sizov, Guangyi, Zhang, G.~Lakshminarayanan, H.~Inan, H.~Shojanazeri, H.~Zou, H.~Wang, H.~Zha, H.~Habeeb, H.~Rudolph, H.~Suk, H.~Aspegren, H.~Goldman, H.~Zhan, I.~Damlaj, I.~Molybog, I.~Tufanov, I.~Leontiadis, I.-E. Veliche, I.~Gat, J.~Weissman, J.~Geboski, J.~Kohli, J.~Lam, J.~Asher, J.-B. Gaya, J.~Marcus,
  J.~Tang, J.~Chan, J.~Zhen, J.~Reizenstein, J.~Teboul, J.~Zhong, J.~Jin, J.~Yang, J.~Cummings, J.~Carvill, J.~Shepard, J.~McPhie, J.~Torres, J.~Ginsburg, J.~Wang, K.~Wu, K.~H. U, K.~Saxena, K.~Khandelwal, K.~Zand, K.~Matosich, K.~Veeraraghavan, K.~Michelena, K.~Li, K.~Jagadeesh, K.~Huang, K.~Chawla, K.~Huang, L.~Chen, L.~Garg, L.~A, L.~Silva, L.~Bell, L.~Zhang, L.~Guo, L.~Yu, L.~Moshkovich, L.~Wehrstedt, M.~Khabsa, M.~Avalani, M.~Bhatt, M.~Mankus, M.~Hasson, M.~Lennie, M.~Reso, M.~Groshev, M.~Naumov, M.~Lathi, M.~Keneally, M.~Liu, M.~L. Seltzer, M.~Valko, M.~Restrepo, M.~Patel, M.~Vyatskov, M.~Samvelyan, M.~Clark, M.~Macey, M.~Wang, M.~J. Hermoso, M.~Metanat, M.~Rastegari, M.~Bansal, N.~Santhanam, N.~Parks, N.~White, N.~Bawa, N.~Singhal, N.~Egebo, N.~Usunier, N.~Mehta, N.~P. Laptev, N.~Dong, N.~Cheng, O.~Chernoguz, O.~Hart, O.~Salpekar, O.~Kalinli, P.~Kent, P.~Parekh, P.~Saab, P.~Balaji, P.~Rittner, P.~Bontrager, P.~Roux, P.~Dollar, P.~Zvyagina, P.~Ratanchandani, P.~Yuvraj, Q.~Liang, R.~Alao, R.~Rodriguez,
  R.~Ayub, R.~Murthy, R.~Nayani, R.~Mitra, R.~Parthasarathy, R.~Li, R.~Hogan, R.~Battey, R.~Wang, R.~Howes, R.~Rinott, S.~Mehta, S.~Siby, S.~J. Bondu, S.~Datta, S.~Chugh, S.~Hunt, S.~Dhillon, S.~Sidorov, S.~Pan, S.~Mahajan, S.~Verma, S.~Yamamoto, S.~Ramaswamy, S.~Lindsay, S.~Lindsay, S.~Feng, S.~Lin, S.~C. Zha, S.~Patil, S.~Shankar, S.~Zhang, S.~Zhang, S.~Wang, S.~Agarwal, S.~Sajuyigbe, S.~Chintala, S.~Max, S.~Chen, S.~Kehoe, S.~Satterfield, S.~Govindaprasad, S.~Gupta, S.~Deng, S.~Cho, S.~Virk, S.~Subramanian, S.~Choudhury, S.~Goldman, T.~Remez, T.~Glaser, T.~Best, T.~Koehler, T.~Robinson, T.~Li, T.~Zhang, T.~Matthews, T.~Chou, T.~Shaked, V.~Vontimitta, V.~Ajayi, V.~Montanez, V.~Mohan, V.~S. Kumar, V.~Mangla, V.~Ionescu, V.~Poenaru, V.~T. Mihailescu, V.~Ivanov, W.~Li, W.~Wang, W.~Jiang, W.~Bouaziz, W.~Constable, X.~Tang, X.~Wu, X.~Wang, X.~Wu, X.~Gao, Y.~Kleinman, Y.~Chen, Y.~Hu, Y.~Jia, Y.~Qi, Y.~Li, Y.~Zhang, Y.~Zhang, Y.~Adi, Y.~Nam, Yu, Wang, Y.~Zhao, Y.~Hao, Y.~Qian, Y.~Li, Y.~He, Z.~Rait, Z.~DeVito,
  Z.~Rosnbrick, Z.~Wen, Z.~Yang, Z.~Zhao, and Z.~Ma.
\newblock The llama 3 herd of models, 2024.
\newblock URL \url{https://arxiv.org/abs/2407.21783}.

\bibitem[Gupta et~al.(2018)Gupta, Koren, and Singer]{gupta2018shampoopreconditionedstochastictensor}
V.~Gupta, T.~Koren, and Y.~Singer.
\newblock Sahampoo: Preconditioned stochastic tensor optimization, 2018.
\newblock URL \url{https://arxiv.org/abs/1802.09568}.

\bibitem[Hoffmann et~al.(2022)Hoffmann, Borgeaud, Mensch, Buchatskaya, Cai, Rutherford, de~Las~Casas, Hendricks, Welbl, Clark, Hennigan, Noland, Millican, van~den Driessche, Damoc, Guy, Osindero, Simonyan, Elsen, Rae, Vinyals, and Sifre]{hoffmann2022trainingcomputeoptimallargelanguage}
J.~Hoffmann, S.~Borgeaud, A.~Mensch, E.~Buchatskaya, T.~Cai, E.~Rutherford, D.~de~Las~Casas, L.~A. Hendricks, J.~Welbl, A.~Clark, T.~Hennigan, E.~Noland, K.~Millican, G.~van~den Driessche, B.~Damoc, A.~Guy, S.~Osindero, K.~Simonyan, E.~Elsen, J.~W. Rae, O.~Vinyals, and L.~Sifre.
\newblock Training compute-optimal large language models, 2022.
\newblock URL \url{https://arxiv.org/abs/2203.15556}.

\bibitem[Hu et~al.(2024)Hu, Tu, Han, He, Cui, Long, Zheng, Fang, Huang, Zhao, Zhang, Thai, Zhang, Wang, Yao, Zhao, Zhou, Cai, Zhai, Ding, Jia, Zeng, Li, Liu, and Sun]{hu2024minicpmunveilingpotentialsmall}
S.~Hu, Y.~Tu, X.~Han, C.~He, G.~Cui, X.~Long, Z.~Zheng, Y.~Fang, Y.~Huang, W.~Zhao, X.~Zhang, Z.~L. Thai, K.~Zhang, C.~Wang, Y.~Yao, C.~Zhao, J.~Zhou, J.~Cai, Z.~Zhai, N.~Ding, C.~Jia, G.~Zeng, D.~Li, Z.~Liu, and M.~Sun.
\newblock Minicpm: Unveiling the potential of small language models with scalable training strategies, 2024.
\newblock URL \url{https://arxiv.org/abs/2404.06395}.

\bibitem[Ivgi et~al.(2023)Ivgi, Hinder, and Carmon]{ivgi2023dogsgdsbestfriend}
M.~Ivgi, O.~Hinder, and Y.~Carmon.
\newblock Dog is sgd's best friend: A parameter-free dynamic step size schedule, 2023.
\newblock URL \url{https://arxiv.org/abs/2302.12022}.

\bibitem[Jiang et~al.(2019)Jiang, Neyshabur, Mobahi, Krishnan, and Bengio]{jiang2019fantasticgeneralizationmeasures}
Y.~Jiang, B.~Neyshabur, H.~Mobahi, D.~Krishnan, and S.~Bengio.
\newblock Fantastic generalization measures and where to find them, 2019.
\newblock URL \url{https://arxiv.org/abs/1912.02178}.

\bibitem[Jordan et~al.(2024)Jordan, Jin, Boza, You, Cesista, Newhouse, and Bernstein]{jordan2024muon}
K.~Jordan, Y.~Jin, V.~Boza, J.~You, F.~Cesista, L.~Newhouse, and J.~Bernstein.
\newblock Muon: An optimizer for hidden layers in neural networks, 2024.
\newblock URL \url{https://kellerjordan.github.io/posts/muon/}.

\bibitem[Kaddour et~al.(2023)Kaddour, Key, Nawrot, Minervini, and Kusner]{kaddour2023traingainrevisitingefficient}
J.~Kaddour, O.~Key, P.~Nawrot, P.~Minervini, and M.~J. Kusner.
\newblock No train no gain: Revisiting efficient training algorithms for transformer-based language models, 2023.
\newblock URL \url{https://arxiv.org/abs/2307.06440}.

\bibitem[Kasimbeg et~al.(2025)Kasimbeg, Schneider, Eschenhagen, Bae, Sastry, Saroufim, Feng, Wright, Yang, Nado, Medapati, Hennig, Rabbat, and Dahl]{kasimbeg2025acceleratingneuralnetworktraining}
P.~Kasimbeg, F.~Schneider, R.~Eschenhagen, J.~Bae, C.~S. Sastry, M.~Saroufim, B.~Feng, L.~Wright, E.~Z. Yang, Z.~Nado, S.~Medapati, P.~Hennig, M.~Rabbat, and G.~E. Dahl.
\newblock Accelerating neural network training: An analysis of the algoperf competition, 2025.
\newblock URL \url{https://arxiv.org/abs/2502.15015}.

\bibitem[Kingma and Ba(2017)]{kingma2017adammethodstochasticoptimization}
D.~P. Kingma and J.~Ba.
\newblock Adam: A method for stochastic optimization, 2017.
\newblock URL \url{https://arxiv.org/abs/1412.6980}.

\bibitem[Large et~al.(2024)Large, Liu, Huh, Bahng, Isola, and Bernstein]{large2024scalableoptimizationmodularnorm}
T.~Large, Y.~Liu, M.~Huh, H.~Bahng, P.~Isola, and J.~Bernstein.
\newblock Scalable optimization in the modular norm, 2024.
\newblock URL \url{https://arxiv.org/abs/2405.14813}.

\bibitem[Levesque et~al.(2012)Levesque, Davis, and Morgenstern]{levesque2012winograd}
H.~J. Levesque, E.~Davis, and L.~Morgenstern.
\newblock The winograd schema challenge.
\newblock \emph{KR}, 2012:\penalty0 13th, 2012.

\bibitem[Li et~al.(2025{\natexlab{a}})Li, Zheng, Hu, Wang, Zhang, Wang, Xuyang, Fan, Zhou, Zhang, and Jiang]{li2025predictablescalei}
H.~Li, W.~Zheng, J.~Hu, Q.~Wang, H.~Zhang, Z.~Wang, S.~Xuyang, Y.~Fan, S.~Zhou, X.~Zhang, and D.~Jiang.
\newblock Predictable scale: Part i -- optimal hyperparameter scaling law in large language model pretraining, 2025{\natexlab{a}}.
\newblock URL \url{https://arxiv.org/abs/2503.04715}.

\bibitem[Li et~al.(2025{\natexlab{b}})Li, Fang, Smyrnis, Ivgi, Jordan, Gadre, Bansal, Guha, Keh, Arora, Garg, Xin, Muennighoff, Heckel, Mercat, Chen, Gururangan, Wortsman, Albalak, Bitton, Nezhurina, Abbas, Hsieh, Ghosh, Gardner, Kilian, Zhang, Shao, Pratt, Sanyal, Ilharco, Daras, Marathe, Gokaslan, Zhang, Chandu, Nguyen, Vasiljevic, Kakade, Song, Sanghavi, Faghri, Oh, Zettlemoyer, Lo, El-Nouby, Pouransari, Toshev, Wang, Groeneveld, Soldaini, Koh, Jitsev, Kollar, Dimakis, Carmon, Dave, Schmidt, and Shankar]{li2025datacomplmsearchgenerationtraining}
J.~Li, A.~Fang, G.~Smyrnis, M.~Ivgi, M.~Jordan, S.~Gadre, H.~Bansal, E.~Guha, S.~Keh, K.~Arora, S.~Garg, R.~Xin, N.~Muennighoff, R.~Heckel, J.~Mercat, M.~Chen, S.~Gururangan, M.~Wortsman, A.~Albalak, Y.~Bitton, M.~Nezhurina, A.~Abbas, C.-Y. Hsieh, D.~Ghosh, J.~Gardner, M.~Kilian, H.~Zhang, R.~Shao, S.~Pratt, S.~Sanyal, G.~Ilharco, G.~Daras, K.~Marathe, A.~Gokaslan, J.~Zhang, K.~Chandu, T.~Nguyen, I.~Vasiljevic, S.~Kakade, S.~Song, S.~Sanghavi, F.~Faghri, S.~Oh, L.~Zettlemoyer, K.~Lo, A.~El-Nouby, H.~Pouransari, A.~Toshev, S.~Wang, D.~Groeneveld, L.~Soldaini, P.~W. Koh, J.~Jitsev, T.~Kollar, A.~G. Dimakis, Y.~Carmon, A.~Dave, L.~Schmidt, and V.~Shankar.
\newblock Datacomp-lm: In search of the next generation of training sets for language models, 2025{\natexlab{b}}.
\newblock URL \url{https://arxiv.org/abs/2406.11794}.

\bibitem[Li(2022)]{li2022blackboxliegroup}
X.~Li.
\newblock Black box lie group preconditioners for sgd, 2022.
\newblock URL \url{https://arxiv.org/abs/2211.04422}.

\bibitem[Li(2018{\natexlab{a}})]{Li_2018}
X.-L. Li.
\newblock Preconditioned stochastic gradient descent.
\newblock \emph{IEEE Transactions on Neural Networks and Learning Systems}, 29\penalty0 (5):\penalty0 1454–1466, May 2018{\natexlab{a}}.
\newblock ISSN 2162-2388.
\newblock \doi{10.1109/tnnls.2017.2672978}.
\newblock URL \url{http://dx.doi.org/10.1109/TNNLS.2017.2672978}.

\bibitem[Li(2018{\natexlab{b}})]{li2018preconditionermatrixliegroup}
X.-L. Li.
\newblock Preconditioner on matrix lie group for sgd, 2018{\natexlab{b}}.
\newblock URL \url{https://arxiv.org/abs/1809.10232}.

\bibitem[Liang et~al.(2025)Liang, Chen, Liu, and Liu]{liang2025cautious}
K.~Liang, L.~Chen, B.~Liu, and Q.~Liu.
\newblock Cautious optimizers: Improving training with one line of code, 2025.
\newblock URL \url{https://arxiv.org/abs/2411.16085}.

\bibitem[Liu et~al.(2024{\natexlab{a}})Liu, Li, Hall, Liang, and Ma]{liu2024sophiascalablestochasticsecondorder}
H.~Liu, Z.~Li, D.~Hall, P.~Liang, and T.~Ma.
\newblock Sophia: A scalable stochastic second-order optimizer for language model pre-training, 2024{\natexlab{a}}.
\newblock URL \url{https://arxiv.org/abs/2305.14342}.

\bibitem[Liu et~al.(2025{\natexlab{a}})Liu, Su, Yao, Jiang, Lai, Du, Qin, Xu, Lu, Yan, Chen, Zheng, Liu, Liu, Yin, He, Zhu, Wang, Wang, Dong, Zhang, Kang, Zhang, Xu, Zhang, Wu, Zhou, and Yang]{liu2025muonscalablellmtraining}
J.~Liu, J.~Su, X.~Yao, Z.~Jiang, G.~Lai, Y.~Du, Y.~Qin, W.~Xu, E.~Lu, J.~Yan, Y.~Chen, H.~Zheng, Y.~Liu, S.~Liu, B.~Yin, W.~He, H.~Zhu, Y.~Wang, J.~Wang, M.~Dong, Z.~Zhang, Y.~Kang, H.~Zhang, X.~Xu, Y.~Zhang, Y.~Wu, X.~Zhou, and Z.~Yang.
\newblock Muon is scalable for llm training, 2025{\natexlab{a}}.
\newblock URL \url{https://arxiv.org/abs/2502.16982}.

\bibitem[Liu et~al.(2021)Liu, Jiang, He, Chen, Liu, Gao, and Han]{liu2021varianceadaptivelearningrate}
L.~Liu, H.~Jiang, P.~He, W.~Chen, X.~Liu, J.~Gao, and J.~Han.
\newblock On the variance of the adaptive learning rate and beyond, 2021.
\newblock URL \url{https://arxiv.org/abs/1908.03265}.

\bibitem[Liu et~al.(2025{\natexlab{b}})Liu, Zheng, Muennighoff, Zeng, Dou, Pang, Jiang, and Lin]{liu2025regmixdatamixtureregression}
Q.~Liu, X.~Zheng, N.~Muennighoff, G.~Zeng, L.~Dou, T.~Pang, J.~Jiang, and M.~Lin.
\newblock Regmix: Data mixture as regression for language model pre-training, 2025{\natexlab{b}}.
\newblock URL \url{https://arxiv.org/abs/2407.01492}.

\bibitem[Liu et~al.(2025{\natexlab{c}})Liu, Liu, and Gore]{liu2025focusorderconcentratedupdating}
Y.~Liu, Z.~Liu, and J.~Gore.
\newblock Focus: First order concentrated updating scheme, 2025{\natexlab{c}}.
\newblock URL \url{https://arxiv.org/abs/2501.12243}.

\bibitem[Liu et~al.(2024{\natexlab{b}})Liu, Zhao, Iandola, Lai, Tian, Fedorov, Xiong, Chang, Shi, Krishnamoorthi, Lai, and Chandra]{liu2024mobilellmoptimizingsubbillionparameter}
Z.~Liu, C.~Zhao, F.~Iandola, C.~Lai, Y.~Tian, I.~Fedorov, Y.~Xiong, E.~Chang, Y.~Shi, R.~Krishnamoorthi, L.~Lai, and V.~Chandra.
\newblock Mobilellm: Optimizing sub-billion parameter language models for on-device use cases, 2024{\natexlab{b}}.
\newblock URL \url{https://arxiv.org/abs/2402.14905}.

\bibitem[Liu et~al.(2025{\natexlab{d}})Liu, Liu, Michaud, Gore, and Tegmark]{liu2025physicsskilllearning}
Z.~Liu, Y.~Liu, E.~J. Michaud, J.~Gore, and M.~Tegmark.
\newblock Physics of skill learning, 2025{\natexlab{d}}.
\newblock URL \url{https://arxiv.org/abs/2501.12391}.

\bibitem[Loshchilov and Hutter(2019)]{loshchilov2019decoupled}
I.~Loshchilov and F.~Hutter.
\newblock Decoupled weight decay regularization, 2019.
\newblock URL \url{https://arxiv.org/abs/1711.05101}.

\bibitem[Lozhkov et~al.(2024)Lozhkov, Li, Allal, Cassano, Lamy-Poirier, Tazi, Tang, Pykhtar, Liu, Wei, Liu, Tian, Kocetkov, Zucker, Belkada, Wang, Liu, Abulkhanov, Paul, Li, Li, Risdal, Li, Zhu, Zhuo, Zheltonozhskii, Dade, Yu, Krauß, Jain, Su, He, Dey, Abati, Chai, Muennighoff, Tang, Oblokulov, Akiki, Marone, Mou, Mishra, Gu, Hui, Dao, Zebaze, Dehaene, Patry, Xu, McAuley, Hu, Scholak, Paquet, Robinson, Anderson, Chapados, Patwary, Tajbakhsh, Jernite, Ferrandis, Zhang, Hughes, Wolf, Guha, von Werra, and de~Vries]{lozhkov2024starcoder2stackv2}
A.~Lozhkov, R.~Li, L.~B. Allal, F.~Cassano, J.~Lamy-Poirier, N.~Tazi, A.~Tang, D.~Pykhtar, J.~Liu, Y.~Wei, T.~Liu, M.~Tian, D.~Kocetkov, A.~Zucker, Y.~Belkada, Z.~Wang, Q.~Liu, D.~Abulkhanov, I.~Paul, Z.~Li, W.-D. Li, M.~Risdal, J.~Li, J.~Zhu, T.~Y. Zhuo, E.~Zheltonozhskii, N.~O.~O. Dade, W.~Yu, L.~Krauß, N.~Jain, Y.~Su, X.~He, M.~Dey, E.~Abati, Y.~Chai, N.~Muennighoff, X.~Tang, M.~Oblokulov, C.~Akiki, M.~Marone, C.~Mou, M.~Mishra, A.~Gu, B.~Hui, T.~Dao, A.~Zebaze, O.~Dehaene, N.~Patry, C.~Xu, J.~McAuley, H.~Hu, T.~Scholak, S.~Paquet, J.~Robinson, C.~J. Anderson, N.~Chapados, M.~Patwary, N.~Tajbakhsh, Y.~Jernite, C.~M. Ferrandis, L.~Zhang, S.~Hughes, T.~Wolf, A.~Guha, L.~von Werra, and H.~de~Vries.
\newblock Starcoder 2 and the stack v2: The next generation, 2024.
\newblock URL \url{https://arxiv.org/abs/2402.19173}.

\bibitem[Lu et~al.(2025)Lu, Zhao, Wei, Wang, Qin, and Liu]{lu2025doessequencemodelingarchitecture}
X.~Lu, Y.~Zhao, S.~Wei, S.~Wang, B.~Qin, and T.~Liu.
\newblock How does sequence modeling architecture influence base capabilities of pre-trained language models? exploring key architecture design principles to avoid base capabilities degradation, 2025.
\newblock URL \url{https://arxiv.org/abs/2505.18522}.

\bibitem[Luo et~al.(2019)Luo, Xiong, Liu, and Sun]{luo2019adaptivegradientmethodsdynamic}
L.~Luo, Y.~Xiong, Y.~Liu, and X.~Sun.
\newblock Adaptive gradient methods with dynamic bound of learning rate, 2019.
\newblock URL \url{https://arxiv.org/abs/1902.09843}.

\bibitem[Luo et~al.(2023)Luo, Ren, Zheng, Jiang, Jiang, and You]{luo2023cameconfidenceguidedadaptivememory}
Y.~Luo, X.~Ren, Z.~Zheng, Z.~Jiang, X.~Jiang, and Y.~You.
\newblock Came: Confidence-guided adaptive memory efficient optimization, 2023.
\newblock URL \url{https://arxiv.org/abs/2307.02047}.

\bibitem[Ma et~al.(2025)Ma, Gong, Scetbon, and Meeds]{ma2025swansgdnormalizationwhitening}
C.~Ma, W.~Gong, M.~Scetbon, and E.~Meeds.
\newblock Swan: Sgd with normalization and whitening enables stateless llm training, 2025.
\newblock URL \url{https://arxiv.org/abs/2412.13148}.

\bibitem[Martens and Grosse(2020)]{martens2020optimizingneuralnetworkskroneckerfactored}
J.~Martens and R.~Grosse.
\newblock Optimizing neural networks with kronecker-factored approximate curvature, 2020.
\newblock URL \url{https://arxiv.org/abs/1503.05671}.

\bibitem[Mihaylov et~al.(2018)Mihaylov, Clark, Khot, and Sabharwal]{Mihaylov2018OpenBookQA}
T.~Mihaylov, P.~Clark, T.~Khot, and A.~Sabharwal.
\newblock Can a suit of armor conduct electricity? a new dataset for open book question answering.
\newblock In E.~Riloff, D.~Chiang, J.~Hockenmaier, and J.~Tsujii, editors, \emph{Proceedings of the 2018 Conference on Empirical Methods in Natural Language Processing}, pages 2381--2391, Brussels, Belgium, Oct--Nov 2018. Association for Computational Linguistics.
\newblock \doi{10.18653/v1/D18-1260}.
\newblock URL \url{https://aclanthology.org/D18-1260/}.

\bibitem[Mishchenko and Defazio(2024)]{mishchenko2024prodigyexpeditiouslyadaptiveparameterfree}
K.~Mishchenko and A.~Defazio.
\newblock Prodigy: An expeditiously adaptive parameter-free learner, 2024.
\newblock URL \url{https://arxiv.org/abs/2306.06101}.

\bibitem[Modoranu et~al.(2024)Modoranu, Safaryan, Malinovsky, Kurti{\'c}, Robert, Richt{\'a}rik, and Alistarh]{modoranu2024microadam}
I.-V. Modoranu, M.~Safaryan, G.~Malinovsky, E.~Kurti{\'c}, T.~Robert, P.~Richt{\'a}rik, and D.~Alistarh.
\newblock Microadam: Accurate adaptive optimization with low space overhead and provable convergence.
\newblock \emph{Advances in Neural Information Processing Systems}, 37:\penalty0 1--43, 2024.

\bibitem[Morwani et~al.(2024)Morwani, Shapira, Vyas, Malach, Kakade, and Janson]{morwani2024newperspectiveshampoospreconditioner}
D.~Morwani, I.~Shapira, N.~Vyas, E.~Malach, S.~Kakade, and L.~Janson.
\newblock A new perspective on shampoo's preconditioner, 2024.
\newblock URL \url{https://arxiv.org/abs/2406.17748}.

\bibitem[Nesterov(1983)]{nesterov1983method}
Y.~Nesterov.
\newblock A method for solving the convex programming problem with convergence rate o (1/k2).
\newblock In \emph{Dokl akad nauk Sssr}, volume 269, page 543, 1983.

\bibitem[OLMo et~al.(2025)OLMo, Walsh, Soldaini, Groeneveld, Lo, Arora, Bhagia, Gu, Huang, Jordan, Lambert, Schwenk, Tafjord, Anderson, Atkinson, Brahman, Clark, Dasigi, Dziri, Guerquin, Ivison, Koh, Liu, Malik, Merrill, Miranda, Morrison, Murray, Nam, Pyatkin, Rangapur, Schmitz, Skjonsberg, Wadden, Wilhelm, Wilson, Zettlemoyer, Farhadi, Smith, and Hajishirzi]{olmo20252olmo2furious}
T.~OLMo, P.~Walsh, L.~Soldaini, D.~Groeneveld, K.~Lo, S.~Arora, A.~Bhagia, Y.~Gu, S.~Huang, M.~Jordan, N.~Lambert, D.~Schwenk, O.~Tafjord, T.~Anderson, D.~Atkinson, F.~Brahman, C.~Clark, P.~Dasigi, N.~Dziri, M.~Guerquin, H.~Ivison, P.~W. Koh, J.~Liu, S.~Malik, W.~Merrill, L.~J.~V. Miranda, J.~Morrison, T.~Murray, C.~Nam, V.~Pyatkin, A.~Rangapur, M.~Schmitz, S.~Skjonsberg, D.~Wadden, C.~Wilhelm, M.~Wilson, L.~Zettlemoyer, A.~Farhadi, N.~A. Smith, and H.~Hajishirzi.
\newblock 2 olmo 2 furious, 2025.
\newblock URL \url{https://arxiv.org/abs/2501.00656}.

\bibitem[Pagliardini et~al.(2024)Pagliardini, Ablin, and Grangier]{pagliardini2024ademamixoptimizerbetterfaster}
M.~Pagliardini, P.~Ablin, and D.~Grangier.
\newblock The ademamix optimizer: Better, faster, older, 2024.
\newblock URL \url{https://arxiv.org/abs/2409.03137}.

\bibitem[Paperno et~al.(2016)Paperno, Kruszewski, Lazaridou, Pham, Bernardi, Pezzelle, Baroni, Boleda, and Fern{\'a}ndez]{Paperno2016LAMBADA}
D.~Paperno, G.~Kruszewski, A.~Lazaridou, N.~Q. Pham, R.~Bernardi, S.~Pezzelle, M.~Baroni, G.~Boleda, and R.~Fern{\'a}ndez.
\newblock The {LAMBADA} dataset: Word prediction requiring a broad discourse context.
\newblock In K.~Erk and N.~A. Smith, editors, \emph{Proceedings of the 54th Annual Meeting of the Association for Computational Linguistics (Volume 1: Long Papers)}, pages 1525--1534, Berlin, Germany, August 2016. Association for Computational Linguistics.
\newblock \doi{10.18653/v1/P16-1144}.
\newblock URL \url{https://aclanthology.org/P16-1144/}.

\bibitem[Pethick et~al.(2025)Pethick, Xie, Antonakopoulos, Zhu, Silveti-Falls, and Cevher]{pethick2025trainingdeeplearningmodels}
T.~Pethick, W.~Xie, K.~Antonakopoulos, Z.~Zhu, A.~Silveti-Falls, and V.~Cevher.
\newblock Training deep learning models with norm-constrained lmos, 2025.
\newblock URL \url{https://arxiv.org/abs/2502.07529}.

\bibitem[Raffel et~al.(2023)Raffel, Shazeer, Roberts, Lee, Narang, Matena, Zhou, Li, and Liu]{raffel2023exploringlimitstransferlearning}
C.~Raffel, N.~Shazeer, A.~Roberts, K.~Lee, S.~Narang, M.~Matena, Y.~Zhou, W.~Li, and P.~J. Liu.
\newblock Exploring the limits of transfer learning with a unified text-to-text transformer, 2023.
\newblock URL \url{https://arxiv.org/abs/1910.10683}.

\bibitem[Reddi et~al.(2018)Reddi, Kale, and Kumar]{j.2018on}
S.~J. Reddi, S.~Kale, and S.~Kumar.
\newblock On the convergence of adam and beyond.
\newblock In \emph{International Conference on Learning Representations}, 2018.
\newblock URL \url{https://openreview.net/forum?id=ryQu7f-RZ}.

\bibitem[Robbins and Monro(1951)]{robbins1951stochastic}
H.~Robbins and S.~Monro.
\newblock A stochastic approximation method.
\newblock \emph{The annals of mathematical statistics}, pages 400--407, 1951.

\bibitem[Sakaguchi et~al.(2020)Sakaguchi, Le~Bras, Bhagavatula, and Choi]{Sakaguchi2020WinoGrande}
K.~Sakaguchi, R.~Le~Bras, C.~Bhagavatula, and Y.~Choi.
\newblock Winogrande: An adversarial winograd schema challenge at scale.
\newblock In \emph{Proceedings of the 34th AAAI Conference on Artificial Intelligence (AAAI)}, pages 8732--8740. AAAI Press, 2020.
\newblock \doi{10.1609/aaai.v34i05.6399}.
\newblock URL \url{https://doi.org/10.1609/aaai.v34i05.6399}.

\bibitem[Schaul et~al.(2013)Schaul, Zhang, and LeCun]{schaul2013peskylearningrates}
T.~Schaul, S.~Zhang, and Y.~LeCun.
\newblock No more pesky learning rates, 2013.
\newblock URL \url{https://arxiv.org/abs/1206.1106}.

\bibitem[Schmidt et~al.(2021)Schmidt, Schneider, and Hennig]{schmidt2021descendingcrowdedvalley}
R.~M. Schmidt, F.~Schneider, and P.~Hennig.
\newblock Descending through a crowded valley - benchmarking deep learning optimizers, 2021.
\newblock URL \url{https://arxiv.org/abs/2007.01547}.

\bibitem[Semenov et~al.(2025)Semenov, Pagliardini, and Jaggi]{semenov2025benchmarkingoptimizerslargelanguage}
A.~Semenov, M.~Pagliardini, and M.~Jaggi.
\newblock Benchmarking optimizers for large language model pretraining, 2025.
\newblock URL \url{https://arxiv.org/abs/2509.01440}.

\bibitem[Shazeer and Stern(2018)]{shazeer2018adafactoradaptivelearningrates}
N.~Shazeer and M.~Stern.
\newblock Adafactor: Adaptive learning rates with sublinear memory cost, 2018.
\newblock URL \url{https://arxiv.org/abs/1804.04235}.

\bibitem[Sutskever et~al.(2013)Sutskever, Martens, Dahl, and Hinton]{pmlr-v28-sutskever13}
I.~Sutskever, J.~Martens, G.~Dahl, and G.~Hinton.
\newblock On the importance of initialization and momentum in deep learning.
\newblock In S.~Dasgupta and D.~McAllester, editors, \emph{Proceedings of the 30th International Conference on Machine Learning}, volume~28 of \emph{Proceedings of Machine Learning Research}, pages 1139--1147, Atlanta, Georgia, USA, 17--19 Jun 2013. PMLR.
\newblock URL \url{https://proceedings.mlr.press/v28/sutskever13.html}.

\bibitem[Talmor et~al.(2018)Talmor, Herzig, Lourie, and Berant]{Talmor2018CommonsenseQA}
A.~Talmor, J.~Herzig, N.~Lourie, and J.~Berant.
\newblock Commonsenseqa: A question answering challenge targeting commonsense knowledge.
\newblock \emph{CoRR}, abs/1811.00937, 2018.
\newblock URL \url{http://arxiv.org/abs/1811.00937}.

\bibitem[Taniguchi et~al.(2024)Taniguchi, Harada, Minegishi, Oshima, Jeong, Nagahara, Iiyama, Suzuki, Iwasawa, and Matsuo]{taniguchi2024adoptmodifiedadamconverge}
S.~Taniguchi, K.~Harada, G.~Minegishi, Y.~Oshima, S.~C. Jeong, G.~Nagahara, T.~Iiyama, M.~Suzuki, Y.~Iwasawa, and Y.~Matsuo.
\newblock Adopt: Modified adam can converge with any $\beta_2$ with the optimal rate, 2024.
\newblock URL \url{https://arxiv.org/abs/2411.02853}.

\bibitem[Team et~al.(2025)Team, Bai, Bao, Chen, Chen, Chen, Chen, Chen, Chen, Chen, Chen, Cui, Ding, Dong, Du, Du, Du, Du, Fan, Feng, Fu, Gao, Gao, Gao, Gao, Gu, Guan, Guo, Guo, Hu, Hao, He, He, He, Hong, Hu, Hu, Huang, Huang, Huang, Jiang, Jiang, Jin, Kang, Lai, Li, Li, Li, Li, Li, Li, Li, Li, Li, Lin, Lin, Lin, Liu, Liu, Liu, Liu, Liu, Liu, Liu, Liu, Liu, Liu, Liu, Liu, Liu, Liu, Liu, Lu, Lu, Ma, Ma, Ma, Mao, Mei, Men, Miao, Pan, Peng, Qin, Qu, Shang, Shi, Shi, Song, Su, Su, Sun, Sung, Tang, Tao, Teng, Wang, Wang, Wang, Wang, Wang, Wang, Wang, Wang, Wang, Wang, Wang, Wang, Wang, Wang, Wang, Wang, Wang, Wei, Wei, Wu, Wu, Wu, Xiao, Xie, Xiong, Xu, Xu, Xu, Xu, Xu, Xu, Xu, Xu, Xu, Xu, Yan, Yan, Yang, Yang, Yang, Yang, Yang, Yao, Yao, Ye, Ye, Yin, Yu, Yuan, Yuan, Yuan, Zhan, Zhang, Zhang, Zhang, Zhang, Zhang, Zhang, Zhang, Zhang, Zhang, Zhang, Zhang, Zhao, Zhao, Zheng, Zheng, Zhou, Zhou, Zhou, Zhu, Zhuang, and Zu]{kimiteam2025kimik2openagentic}
K.~Team, Y.~Bai, Y.~Bao, G.~Chen, J.~Chen, N.~Chen, R.~Chen, Y.~Chen, Y.~Chen, Y.~Chen, Z.~Chen, J.~Cui, H.~Ding, M.~Dong, A.~Du, C.~Du, D.~Du, Y.~Du, Y.~Fan, Y.~Feng, K.~Fu, B.~Gao, H.~Gao, P.~Gao, T.~Gao, X.~Gu, L.~Guan, H.~Guo, J.~Guo, H.~Hu, X.~Hao, T.~He, W.~He, W.~He, C.~Hong, Y.~Hu, Z.~Hu, W.~Huang, Z.~Huang, Z.~Huang, T.~Jiang, Z.~Jiang, X.~Jin, Y.~Kang, G.~Lai, C.~Li, F.~Li, H.~Li, M.~Li, W.~Li, Y.~Li, Y.~Li, Z.~Li, Z.~Li, H.~Lin, X.~Lin, Z.~Lin, C.~Liu, C.~Liu, H.~Liu, J.~Liu, J.~Liu, L.~Liu, S.~Liu, T.~Y. Liu, T.~Liu, W.~Liu, Y.~Liu, Y.~Liu, Y.~Liu, Y.~Liu, Z.~Liu, E.~Lu, L.~Lu, S.~Ma, X.~Ma, Y.~Ma, S.~Mao, J.~Mei, X.~Men, Y.~Miao, S.~Pan, Y.~Peng, R.~Qin, B.~Qu, Z.~Shang, L.~Shi, S.~Shi, F.~Song, J.~Su, Z.~Su, X.~Sun, F.~Sung, H.~Tang, J.~Tao, Q.~Teng, C.~Wang, D.~Wang, F.~Wang, H.~Wang, J.~Wang, J.~Wang, J.~Wang, S.~Wang, S.~Wang, Y.~Wang, Y.~Wang, Y.~Wang, Y.~Wang, Y.~Wang, Z.~Wang, Z.~Wang, Z.~Wang, C.~Wei, Q.~Wei, W.~Wu, X.~Wu, Y.~Wu, C.~Xiao, X.~Xie, W.~Xiong, B.~Xu, J.~Xu, J.~Xu, L.~H. Xu,
  L.~Xu, S.~Xu, W.~Xu, X.~Xu, Y.~Xu, Z.~Xu, J.~Yan, Y.~Yan, X.~Yang, Y.~Yang, Z.~Yang, Z.~Yang, Z.~Yang, H.~Yao, X.~Yao, W.~Ye, Z.~Ye, B.~Yin, L.~Yu, E.~Yuan, H.~Yuan, M.~Yuan, H.~Zhan, D.~Zhang, H.~Zhang, W.~Zhang, X.~Zhang, Y.~Zhang, Y.~Zhang, Y.~Zhang, Y.~Zhang, Y.~Zhang, Y.~Zhang, Z.~Zhang, H.~Zhao, Y.~Zhao, H.~Zheng, S.~Zheng, J.~Zhou, X.~Zhou, Z.~Zhou, Z.~Zhu, W.~Zhuang, and X.~Zu.
\newblock Kimi k2: Open agentic intelligence, 2025.
\newblock URL \url{https://arxiv.org/abs/2507.20534}.

\bibitem[Tieleman(2012)]{tieleman2012lecture}
T.~Tieleman.
\newblock Lecture 6.5-rmsprop: Divide the gradient by a running average of its recent magnitude.
\newblock \emph{COURSERA: Neural networks for machine learning}, 4\penalty0 (2):\penalty0 26, 2012.

\bibitem[Touvron et~al.(2023{\natexlab{a}})Touvron, Lavril, Izacard, Martinet, Lachaux, Lacroix, Rozière, Goyal, Hambro, Azhar, Rodriguez, Joulin, Grave, and Lample]{touvron2023llamaopenefficientfoundation}
H.~Touvron, T.~Lavril, G.~Izacard, X.~Martinet, M.-A. Lachaux, T.~Lacroix, B.~Rozière, N.~Goyal, E.~Hambro, F.~Azhar, A.~Rodriguez, A.~Joulin, E.~Grave, and G.~Lample.
\newblock Llama: Open and efficient foundation language models, 2023{\natexlab{a}}.
\newblock URL \url{https://arxiv.org/abs/2302.13971}.

\bibitem[Touvron et~al.(2023{\natexlab{b}})Touvron, Martin, Stone, Albert, Almahairi, Babaei, Bashlykov, Batra, Bhargava, Bhosale, Bikel, Blecher, Ferrer, Chen, Cucurull, Esiobu, Fernandes, Fu, Fu, Fuller, Gao, Goswami, Goyal, Hartshorn, Hosseini, Hou, Inan, Kardas, Kerkez, Khabsa, Kloumann, Korenev, Koura, Lachaux, Lavril, Lee, Liskovich, Lu, Mao, Martinet, Mihaylov, Mishra, Molybog, Nie, Poulton, Reizenstein, Rungta, Saladi, Schelten, Silva, Smith, Subramanian, Tan, Tang, Taylor, Williams, Kuan, Xu, Yan, Zarov, Zhang, Fan, Kambadur, Narang, Rodriguez, Stojnic, Edunov, and Scialom]{touvron2023llama2openfoundation}
H.~Touvron, L.~Martin, K.~Stone, P.~Albert, A.~Almahairi, Y.~Babaei, N.~Bashlykov, S.~Batra, P.~Bhargava, S.~Bhosale, D.~Bikel, L.~Blecher, C.~C. Ferrer, M.~Chen, G.~Cucurull, D.~Esiobu, J.~Fernandes, J.~Fu, W.~Fu, B.~Fuller, C.~Gao, V.~Goswami, N.~Goyal, A.~Hartshorn, S.~Hosseini, R.~Hou, H.~Inan, M.~Kardas, V.~Kerkez, M.~Khabsa, I.~Kloumann, A.~Korenev, P.~S. Koura, M.-A. Lachaux, T.~Lavril, J.~Lee, D.~Liskovich, Y.~Lu, Y.~Mao, X.~Martinet, T.~Mihaylov, P.~Mishra, I.~Molybog, Y.~Nie, A.~Poulton, J.~Reizenstein, R.~Rungta, K.~Saladi, A.~Schelten, R.~Silva, E.~M. Smith, R.~Subramanian, X.~E. Tan, B.~Tang, R.~Taylor, A.~Williams, J.~X. Kuan, P.~Xu, Z.~Yan, I.~Zarov, Y.~Zhang, A.~Fan, M.~Kambadur, S.~Narang, A.~Rodriguez, R.~Stojnic, S.~Edunov, and T.~Scialom.
\newblock Llama 2: Open foundation and fine-tuned chat models, 2023{\natexlab{b}}.
\newblock URL \url{https://arxiv.org/abs/2307.09288}.

\bibitem[Vyas et~al.(2025)Vyas, Morwani, Zhao, Kwun, Shapira, Brandfonbrener, Janson, and Kakade]{vyas2025soapimprovingstabilizingshampoo}
N.~Vyas, D.~Morwani, R.~Zhao, M.~Kwun, I.~Shapira, D.~Brandfonbrener, L.~Janson, and S.~Kakade.
\newblock Soap: Improving and stabilizing shampoo using adam, 2025.
\newblock URL \url{https://arxiv.org/abs/2409.11321}.

\bibitem[Wang et~al.(2025)Wang, Wang, Zhou, Yan, E, and Wu]{wang2025sharpnessdisparityprincipletransformers}
J.~Wang, M.~Wang, Z.~Zhou, J.~Yan, W.~E, and L.~Wu.
\newblock The sharpness disparity principle in transformers for accelerating language model pre-training, 2025.
\newblock URL \url{https://arxiv.org/abs/2502.19002}.

\bibitem[Wang et~al.(2024)Wang, Liu, Xiao, Liu, Yang, Xu, Pu, Zheng, Zhang, and Li]{wang2024cadamconfidencebasedoptimizationonline}
S.~Wang, A.~Liu, J.~Xiao, H.~Liu, Y.~Yang, C.~Xu, Q.~Pu, S.~Zheng, W.~Zhang, and J.~Li.
\newblock Cadam: Confidence-based optimization for online learning, 2024.
\newblock URL \url{https://arxiv.org/abs/2411.19647}.

\bibitem[Wen et~al.(2024)Wen, Li, Wang, Hall, Liang, and Ma]{wen2024understandingwarmupstabledecaylearningrates}
K.~Wen, Z.~Li, J.~Wang, D.~Hall, P.~Liang, and T.~Ma.
\newblock Understanding warmup-stable-decay learning rates: A river valley loss landscape perspective, 2024.
\newblock URL \url{https://arxiv.org/abs/2410.05192}.

\bibitem[Xie et~al.(2024)Xie, Zhou, Li, Lin, and Yan]{xie2024adanadaptivenesterovmomentum}
X.~Xie, P.~Zhou, H.~Li, Z.~Lin, and S.~Yan.
\newblock Adan: Adaptive nesterov momentum algorithm for faster optimizing deep models, 2024.
\newblock URL \url{https://arxiv.org/abs/2208.06677}.

\bibitem[Xie et~al.(2022)Xie, Wang, Zhang, Sato, and Sugiyama]{xie2022adaptiveinertiadisentanglingeffects}
Z.~Xie, X.~Wang, H.~Zhang, I.~Sato, and M.~Sugiyama.
\newblock Adaptive inertia: Disentangling the effects of adaptive learning rate and momentum, 2022.
\newblock URL \url{https://arxiv.org/abs/2006.15815}.

\bibitem[Yang et~al.(2025)Yang, Li, Yang, Zhang, Hui, Zheng, Yu, Gao, Huang, Lv, Zheng, Liu, Zhou, Huang, Hu, Ge, Wei, Lin, Tang, Yang, Tu, Zhang, Yang, Yang, Zhou, Zhou, Lin, Dang, Bao, Yang, Yu, Deng, Li, Xue, Li, Zhang, Wang, Zhu, Men, Gao, Liu, Luo, Li, Tang, Yin, Ren, Wang, Zhang, Ren, Fan, Su, Zhang, Zhang, Wan, Liu, Wang, Cui, Zhang, Zhou, and Qiu]{yang2025qwen3technicalreport}
A.~Yang, A.~Li, B.~Yang, B.~Zhang, B.~Hui, B.~Zheng, B.~Yu, C.~Gao, C.~Huang, C.~Lv, C.~Zheng, D.~Liu, F.~Zhou, F.~Huang, F.~Hu, H.~Ge, H.~Wei, H.~Lin, J.~Tang, J.~Yang, J.~Tu, J.~Zhang, J.~Yang, J.~Yang, J.~Zhou, J.~Zhou, J.~Lin, K.~Dang, K.~Bao, K.~Yang, L.~Yu, L.~Deng, M.~Li, M.~Xue, M.~Li, P.~Zhang, P.~Wang, Q.~Zhu, R.~Men, R.~Gao, S.~Liu, S.~Luo, T.~Li, T.~Tang, W.~Yin, X.~Ren, X.~Wang, X.~Zhang, X.~Ren, Y.~Fan, Y.~Su, Y.~Zhang, Y.~Zhang, Y.~Wan, Y.~Liu, Z.~Wang, Z.~Cui, Z.~Zhang, Z.~Zhou, and Z.~Qiu.
\newblock Qwen3 technical report, 2025.
\newblock URL \url{https://arxiv.org/abs/2505.09388}.

\bibitem[Yang(2020{\natexlab{a}})]{yang2020scalinglimitswideneural}
G.~Yang.
\newblock Scaling limits of wide neural networks with weight sharing: Gaussian process behavior, gradient independence, and neural tangent kernel derivation, 2020{\natexlab{a}}.
\newblock URL \url{https://arxiv.org/abs/1902.04760}.

\bibitem[Yang(2020{\natexlab{b}})]{yang2020tensor}
G.~Yang.
\newblock Tensor programs iii: Neural matrix laws.
\newblock \emph{arXiv preprint arXiv:2009.10685}, 2020{\natexlab{b}}.

\bibitem[Yang(2020{\natexlab{c}})]{yang2020tensor2}
G.~Yang.
\newblock Tensor programs ii: Neural tangent kernel for any architecture.
\newblock \emph{arXiv preprint arXiv:2006.14548}, 2020{\natexlab{c}}.

\bibitem[Yang(2021)]{yang2021tensorprogramsiwide}
G.~Yang.
\newblock Tensor programs i: Wide feedforward or recurrent neural networks of any architecture are gaussian processes, 2021.
\newblock URL \url{https://arxiv.org/abs/1910.12478}.

\bibitem[Yang and Littwin(2023)]{yang2023tensorprogramsivbadaptive}
G.~Yang and E.~Littwin.
\newblock Tensor programs ivb: Adaptive optimization in the infinite-width limit, 2023.
\newblock URL \url{https://arxiv.org/abs/2308.01814}.

\bibitem[Yang et~al.(2022)Yang, Hu, Babuschkin, Sidor, Liu, Farhi, Ryder, Pachocki, Chen, and Gao]{yang2022tensorprogramsvtuning}
G.~Yang, E.~J. Hu, I.~Babuschkin, S.~Sidor, X.~Liu, D.~Farhi, N.~Ryder, J.~Pachocki, W.~Chen, and J.~Gao.
\newblock Tensor programs v: Tuning large neural networks via zero-shot hyperparameter transfer, 2022.
\newblock URL \url{https://arxiv.org/abs/2203.03466}.

\bibitem[Yang et~al.(2024)Yang, Simon, and Bernstein]{yang2024spectralconditionfeaturelearning}
G.~Yang, J.~B. Simon, and J.~Bernstein.
\newblock A spectral condition for feature learning, 2024.
\newblock URL \url{https://arxiv.org/abs/2310.17813}.

\bibitem[Yao et~al.(2021)Yao, Gholami, Shen, Mustafa, Keutzer, and Mahoney]{Yao_Gholami_Shen_Mustafa_Keutzer_Mahoney_2021}
Z.~Yao, A.~Gholami, S.~Shen, M.~Mustafa, K.~Keutzer, and M.~Mahoney.
\newblock Adahessian: An adaptive second order optimizer for machine learning.
\newblock \emph{Proceedings of the AAAI Conference on Artificial Intelligence}, 35\penalty0 (12):\penalty0 10665--10673, May 2021.
\newblock \doi{10.1609/aaai.v35i12.17275}.
\newblock URL \url{https://ojs.aaai.org/index.php/AAAI/article/view/17275}.

\bibitem[You et~al.(2017)You, Gitman, and Ginsburg]{you2017largebatchtrainingconvolutional}
Y.~You, I.~Gitman, and B.~Ginsburg.
\newblock Large batch training of convolutional networks, 2017.
\newblock URL \url{https://arxiv.org/abs/1708.03888}.

\bibitem[You et~al.(2020)You, Li, Reddi, Hseu, Kumar, Bhojanapalli, Song, Demmel, Keutzer, and Hsieh]{you2020largebatchoptimizationdeep}
Y.~You, J.~Li, S.~Reddi, J.~Hseu, S.~Kumar, S.~Bhojanapalli, X.~Song, J.~Demmel, K.~Keutzer, and C.-J. Hsieh.
\newblock Large batch optimization for deep learning: Training bert in 76 minutes, 2020.
\newblock URL \url{https://arxiv.org/abs/1904.00962}.

\bibitem[Yuan et~al.(2025)Yuan, Liu, Wu, Zhou, and Gu]{yuan2025mars}
H.~Yuan, Y.~Liu, S.~Wu, X.~Zhou, and Q.~Gu.
\newblock Mars: Unleashing the power of variance reduction for training large models, 2025.
\newblock URL \url{https://arxiv.org/abs/2411.10438}.

\bibitem[Zaheer et~al.(2018)Zaheer, Reddi, Sachan, Kale, and Kumar]{zaheer2018adaptive}
M.~Zaheer, S.~Reddi, D.~Sachan, S.~Kale, and S.~Kumar.
\newblock Adaptive methods for nonconvex optimization.
\newblock \emph{Advances in neural information processing systems}, 31, 2018.

\bibitem[Zeiler(2012)]{zeiler2012adadeltaadaptivelearningrate}
M.~D. Zeiler.
\newblock Adadelta: An adaptive learning rate method, 2012.
\newblock URL \url{https://arxiv.org/abs/1212.5701}.

\bibitem[Zellers et~al.(2019)Zellers, Holtzman, Bisk, Farhadi, and Choi]{Zellers2019HellaSwag}
R.~Zellers, A.~Holtzman, Y.~Bisk, A.~Farhadi, and Y.~Choi.
\newblock Hellaswag: Can a machine really finish your sentence?
\newblock In \emph{Proceedings of the 57th Annual Meeting of the Association for Computational Linguistics (ACL)}, pages 4791--4800, 2019.
\newblock \doi{10.18653/V1/P19-1472}.
\newblock URL \url{https://doi.org/10.18653/V1/P19-1472}.

\bibitem[Zhang et~al.(2025{\natexlab{a}})Zhang, Morwani, Vyas, Wu, Zou, Ghai, Foster, and Kakade]{zhang2025doescriticalbatchsize}
H.~Zhang, D.~Morwani, N.~Vyas, J.~Wu, D.~Zou, U.~Ghai, D.~Foster, and S.~Kakade.
\newblock How does critical batch size scale in pre-training?, 2025{\natexlab{a}}.
\newblock URL \url{https://arxiv.org/abs/2410.21676}.

\bibitem[Zhang et~al.(2019)Zhang, Lucas, Hinton, and Ba]{zhang2019lookaheadoptimizerksteps}
M.~R. Zhang, J.~Lucas, G.~Hinton, and J.~Ba.
\newblock Lookahead optimizer: k steps forward, 1 step back, 2019.
\newblock URL \url{https://arxiv.org/abs/1907.08610}.

\bibitem[Zhang et~al.(2025{\natexlab{b}})Zhang, Chen, Li, Ding, Wu, Kingma, Ye, Luo, and Sun]{zhang2025adamminiusefewerlearning}
Y.~Zhang, C.~Chen, Z.~Li, T.~Ding, C.~Wu, D.~P. Kingma, Y.~Ye, Z.-Q. Luo, and R.~Sun.
\newblock Adam-mini: Use fewer learning rates to gain more, 2025{\natexlab{b}}.
\newblock URL \url{https://arxiv.org/abs/2406.16793}.

\bibitem[Zhao et~al.(2024)Zhao, Zhang, Chen, Wang, Anandkumar, and Tian]{zhao2024galorememoryefficientllmtraining}
J.~Zhao, Z.~Zhang, B.~Chen, Z.~Wang, A.~Anandkumar, and Y.~Tian.
\newblock Galore: Memory-efficient llm training by gradient low-rank projection, 2024.
\newblock URL \url{https://arxiv.org/abs/2403.03507}.

\bibitem[Zhao et~al.(2025)Zhao, Morwani, Brandfonbrener, Vyas, and Kakade]{zhao2025deconstructingmakesgoodoptimizer}
R.~Zhao, D.~Morwani, D.~Brandfonbrener, N.~Vyas, and S.~Kakade.
\newblock Deconstructing what makes a good optimizer for language models, 2025.
\newblock URL \url{https://arxiv.org/abs/2407.07972}.

\bibitem[Zhu et~al.(2025)Zhu, Zhang, Cong, Liu, Park, Chandra, Long, Pan, Wang, and Lee]{zhu2025apollosgdlikememoryadamwlevel}
H.~Zhu, Z.~Zhang, W.~Cong, X.~Liu, S.~Park, V.~Chandra, B.~Long, D.~Z. Pan, Z.~Wang, and J.~Lee.
\newblock Apollo: Sgd-like memory, adamw-level performance, 2025.
\newblock URL \url{https://arxiv.org/abs/2412.05270}.

\bibitem[Zhuang et~al.(2020)Zhuang, Tang, Ding, Tatikonda, Dvornek, Papademetris, and Duncan]{zhuang2020adabeliefoptimizeradaptingstepsizes}
J.~Zhuang, T.~Tang, Y.~Ding, S.~Tatikonda, N.~Dvornek, X.~Papademetris, and J.~S. Duncan.
\newblock Adabelief optimizer: Adapting stepsizes by the belief in observed gradients, 2020.
\newblock URL \url{https://arxiv.org/abs/2010.07468}.

\end{thebibliography}

% Appendices
\newpage
\appendix
\section{Optimizer Definitions}
\label{def:optimizerdesign}
In this section, we present the algorithm for each optimizer we evaluated.

Throughout this section, we use the following notation: $w_t$ for model parameters, $g_t$ for gradients at step $t$, $\eta$ for learning rate, $\lambda$ for weight decay, $\beta_1,\beta_2$ for moment decay rates, $\epsilon$ for numerical stability, $\gradnorm$ for gradient norm, and $m,v$ for first and second moments. All operations are element-wise unless specified.

\begin{algorithm}[H]
  \caption{AdamW}
  \label{alg:adamw}
  \begin{tcolorbox}
  \textbf{Hyperparameters:} $\beta_1,\;\beta_2,\;\epsilon,\;\eta,\;\lambda, \; \gradnorm$

  \textbf{State:} $m,\;v$

  \textbf{Update Rule:}
  \begin{align*} 
  \hat g_t &= g_t \max\{1, \frac{\gradnorm}{\|g_t\|_2}\} \\
  m_t &= \beta_1\,m_{t-1} + (1-\beta_1)\, \hat g_t,\\
  v_t &= \beta_2\,v_{t-1} + (1-\beta_2)\, \hat g_t^2,\\
  \hat m_t &= \frac{m_t}{1-\beta_1^t},\quad
  \hat v_t = \frac{v_t}{1-\beta_2^t},\\
  w_{t+1} &= w_t - \eta\,\frac{\hat m_t}{\sqrt{\hat v_t} + \epsilon}
                \;-\;\eta\,\lambda\,w_t.
  \end{align*}
  \end{tcolorbox}
\end{algorithm}
\begin{algorithm}[H]
  \caption{Nesterov AdamW}
  \label{alg:nadamw}
  \begin{tcolorbox}
  \textbf{Hyperparameters:} $\beta_1,\;\beta_2,\;\epsilon,\;\eta,\;\lambda,\;\gradnorm$

  \textbf{State:} $m,\;v$

  \textbf{Update Rule:}
  \begin{align*}
  \hat g_t &= g_t \max\{1, \frac{\gradnorm}{\|g_t\|_2}\} \\
  m_t &= \beta_1\,m_{t-1} + (1-\beta_1)\,\hat g_t,\\
  v_t &= \beta_2\,v_{t-1} + (1-\beta_2)\,\hat g_t^2,\\
  \tilde m_t &= \beta_1\,m_t + (1-\beta_1)\,\hat g_t,\\
  \hat m_t &= \frac{\tilde m_t}{1-\beta_1^t},\quad
  \hat v_t = \frac{v_t}{1-\beta_2^t},\\
  w_{t+1} &= w_t 
    - \eta\,\frac{\hat m_t}{\sqrt{\hat v_t} + \epsilon}
    \;-\;\eta\,\lambda\,w_t.
  \end{align*}
  \end{tcolorbox}
\end{algorithm}
\begin{algorithm}[H]
  \caption{Lion}
  \label{alg:lion}
  \begin{tcolorbox}
  \textbf{Hyperparameters:} $\beta_1, \beta_2,\;\eta,\;\epsilon,\;\lambda,\;\gradnorm$

  \textbf{State:} $m$

  \textbf{Update Rule:}
  \begin{align*}
  \hat g_t &= g_t \max\{1, \frac{\gradnorm}{\|g_t\|_2}\} \\
  \hat m_t &= \beta_1 \,m_{t-1} + (1-\beta_1)\,\hat g_t,\\
  m_{t + 1} &= \beta_2 \,m_{t - 1} + (1-\beta_2)\,\hat g_t,\\
  w_{t+1} &= w_t - \eta\,\mathrm{sign}\bigl(\hat m_t\bigr) - \eta\,\lambda\,w_t.
  \end{align*}
  \end{tcolorbox}
\end{algorithm}

\begin{algorithm}[H]
  \caption{Sophia-H}
  \label{alg:sophia}
  \begin{tcolorbox}
    \textbf{Hyperparameters:}
      $\{\eta_t\}_{t=1}^T,\; \lambda,\; k,\; \beta_1,\;\beta_2,\;\gamma,\;\epsilon$ \\[0.5ex]
    \textbf{State:} $m_0 = 0,\; h_{1-k} = 0,\; \theta_1$ \\[0.5ex]
    \textbf{For $t=1,\dots,T$:}
    \begin{align*}
      & g_t = \nabla_{\theta}L_t(\theta_t) \\[0.3em]
      & m_t = \beta_1\,m_{t-1} + (1-\beta_1)\,g_t \\[0.6em]
      &\text{If }t \bmod k = 1\text{:}\\
      &\quad r\sim\{\pm1\}^d,\quad 
            v = g_t \cdot r,\quad
            u = \nabla_{\theta}v,\\
      &\quad \hat h \;\;=\;\; r \odot u \\[0.3em]
      &\quad
            h_t = \beta_2\,h_{t-k} + (1-\beta_2)\,\hat h \\[0.6em]
      &\text{Else:}\quad h_t = h_{t-1} \\[0.6em]
      &\theta_t \;=\; \theta_t \;-\; \eta_t\,\lambda\,\theta_t \\[0.6em]
      &\theta_{t+1} = \theta_t
           \;-\; \eta_t\,\mathrm{clip}\!\Bigl(\tfrac{m_t}{\max(\gamma\,h_t,\epsilon)},\,1\Bigr)
    \end{align*}
  \end{tcolorbox}
\end{algorithm}

\begin{algorithm}[H]
  \caption{MARS}
  \label{alg:mars}
  \begin{tcolorbox}
  \textbf{Hyperparameters:} $\beta_1,\;\beta_2,\;\gamma,\;\epsilon,\;\eta,\; \lambda,\;\gradnorm$

  \textbf{State:} $m,\;v,\;g_{t-1}$

  \textbf{Update Rule:}
  \begin{align*}
  c_t &= g_t + \gamma\,\frac{\beta_1}{1-\beta_1}\,(g_t - g_{t-1}),\\
  \hat c_t &= c_t \max\{1, \frac{\gradnorm}{\|c_t\|_2}\}, \\
  m_t &= \beta_1\,m_{t-1} + (1-\beta_1)\, \hat c_t,\\
  v_t &= \beta_2\,v_{t-1} + (1-\beta_2)\, \hat c_t^2,\\
  \hat m_t &= \frac{m_t}{1-\beta_1^t},\quad
  \hat v_t = \frac{v_t}{1-\beta_2^t},\\
  w_{t+1} &= w_t - \eta\,\frac{\hat m_t}{\sqrt{\hat v_t} + \epsilon} - \eta\,\lambda\,w_t.
  \end{align*}
  \end{tcolorbox}
\end{algorithm}
\begin{algorithm}[H]
  \caption{Adam-mini}
  \label{alg:adammini}
  \begin{tcolorbox}
  \textbf{Hyperparameters:} $\beta_1,\;\beta_2,\;\epsilon,\;\eta,\;\lambda,\;\gradnorm$

  \textbf{State:} $m$ (with the same shape as $w$),  $v$ (one scalar for each block)

  \textbf{Setup:}
  \begin{enumerate}[left=1em]
    \item Partition all parameters into \texttt{param\_blocks}: please refer to~\cite{zhang2025adamminiusefewerlearning} for the exact partition scheme.
    \item We will use $w_{t,b}$ and $m_{t,b}$ to denote the parameters in block $b$ at step $t$
  \end{enumerate}

  \vspace{0.5ex}

  \textbf{Update}
    \begin{align*}
      \hat g_{t} &= g_{t} \max\{1, \frac{\gradnorm}{\|g_{t}\|_2}\} \\
      m_{t} &= \beta_1\,m_{t-1} + (1-\beta_1)\,\hat g_{t},\\
      \hat m_{t} &= \frac{m_{t}}{1-\beta_1^t},\quad
    \end{align*}
    
    \textbf{for each block} $b\in\texttt{param\_blocks}$ 
    \begin{align*}
            v_{t,b} &= \beta_2\,v_{t-1,b} + (1-\beta_2) \mathrm{mean}(\,g_{t,b}^2)\\
             \hat v_{t,b} &= \frac{v_{t,b}}{1-\beta_2^t}.
        \end{align*}

    \[
      w_{t,b}
        = w_{t-1,b}
          - \eta\,\hat m_{t,b} / (\sqrt{v_{t,b}} + \epsilon)
          - \eta\,\lambda\,w_{t-1,b}.
    \]
  \end{tcolorbox}
\end{algorithm}

\begin{algorithm}[H]
  \caption{Cautious}
  \label{alg:cautious}
  \begin{tcolorbox}
  \textbf{Hyperparameters:} $\beta_1,\;\beta_2,\;\epsilon,\;\eta,\;\lambda,\;\gradnorm$

  \textbf{State:} $m,\;v$

  \textbf{Update Rule:}
  \begin{align*}
  \hat g_t &= g_t \max\{1, \frac{\gradnorm}{\|g_t\|_2}\} \\
  m_t &= \beta_1\,m_{t-1} + (1-\beta_1)\,\hat g_t,\\
  v_t &= \beta_2\,v_{t-1} + (1-\beta_2)\,\hat g_t^2,\\
  \hat m_t &= \frac{m_t}{1-\beta_1^t},\quad
  \hat v_t = \frac{v_t}{1-\beta_2^t},\\
  u_t &= \frac{\hat m_t}{\sqrt{\hat v_t} + \epsilon},\quad
  s_t = \mathbb{I}\bigl(u_t\cdot \hat g_t > 0\bigr),\\
  \hat u_t &= \frac{u_t \cdot s_t}{\mathrm{mean}(s_t)},\quad
  w_{t+1} = w_t - \eta\,\hat u_t - \eta\,\lambda\,w_t.
  \end{align*}
  \end{tcolorbox}
\end{algorithm}

\begin{algorithm}[H]
  \caption{Muon}
  \label{alg:muon}
  \begin{tcolorbox}
  \textbf{Hyperparameters:} $\beta,\;\eta,\;\epsilon, \beta_1, \beta_2, \epsilon_{Adam}, \eta_{Adam}, \lambda, \gradnorm$ 

  \textbf{State:} $m$

  \textbf{Update Rule For Weights in LM Head, Embedding, and LayerNorm:} Same as AdamW

  \textbf{Update Rule For Matrices in Transformer Layer :}
  \begin{align*}
  \hat g_t &= g_t \max\{1, \frac{\gradnorm}{\|g_t\|_2}\} \\
  m_t &= \beta\,m_{t-1} + \hat g_t,\\
  u   &= \beta\,m_t + \hat g_t,\\
  u   &= \mathrm{NewtonSchulz}(u,\,\text{steps}=5),\\
  s   &= \sqrt{\max\!\bigl(1,\tfrac{\mathrm{rows}(w)}{\mathrm{cols}(w)}\bigr)},\\
  u   &= s\,u,\\
  w_{t+1} &= w_t - \eta\,u - \eta\,\lambda\,w_t.
  \end{align*}

  \textbf{Newton–Schulz Orthogonalization (Operating on Matrices):}
  \begin{align*}
  &X \leftarrow \frac{X}{\lVert X\rVert + \epsilon},\quad
    \text{transpose}\leftarrow (\mathrm{rows}(X)>\mathrm{cols}(X)),\\
  &\text{if transpose: }X\leftarrow X^\top,\\
  &\text{for }i=1\ldots5:\;
    A=X\,X^\top,\;
    B=3.4445\,A - 4.7750\,A^2 + 2.0315\,A^3,\\
  &\quad X\leftarrow 3.4445\,X + B\,X,\\
  &\text{if transpose: }X\leftarrow X^\top,\quad
    \text{return }X.
  \end{align*}
  \end{tcolorbox}
\end{algorithm}

\begin{algorithm}[H]
  \caption{Scion}
  \label{alg:scion}
  \begin{tcolorbox}
  \textbf{Hyperparameters:} $\beta,\;\eta,\;\epsilon, \beta_1, \beta_2, \eta_{\mathrm{SignGD}}, \gradnorm$ 

  \textbf{State:} $m$

  \textbf{Update Rule For Matrices in LM Head and Embedding:} 
  \begin{align*}
  \hat g_t &= g_t \max\{1, \frac{\gradnorm}{\|g_t\|_2}\} \\
  m_t &= \beta_1\,m_{t-1} + (1-\beta_1)\,\hat g_t,\\
  w_{t+1} &= w_t - \eta_{\mathrm{SignGD}}\,\mathrm{sign}\bigl(\hat m_t\bigr).
  \end{align*}
  
  \textbf{Update Rule For Matrices in Transformer Layer :} Same as Muon

  \end{tcolorbox}
\end{algorithm}

\begin{algorithm}[H]
  \caption{PSGD Kron}
  \label{alg:kron}
  \begin{tcolorbox}
  \textbf{Hyperparameters:}
  $\beta_{1}$, 
  $\eta$,
  $\lambda$,
  $\epsilon$,
  $\gradnorm$,
  $\mathrm{normalize\_grads}$,
  $\text{partition\_grads\_into\_blocks}$,
  $\text{merge\_small\_dims}$,
  $\text{block\_size}$,
  $\text{target\_merged\_dim\_size}$,
  $p_{\mathrm{upd}}(t)$,

  \textbf{Setup:}

  If $\text{merge\_small\_dims}$ is True, then try merging small dimensions into a single dimension with size $\text{target\_merged\_dim\_size}$ greedily.

  If $\text{partition\_grads\_into\_blocks}$ is True, then partition all parameters into $\text{block\_size} \times \text{block\_size}$ blocks.

  Define $\text{unfold}_i$ as the function that unfolds all dimensions except the $i$-th dimension into a single dimension and $\text{fold}_i$ as the inverse function.

  \textbf{State (per block $\ell$):}
  Denote $w^{(\ell)}$ as the parameters in block $\ell$ and assume it has shape $d_1 \times d_2 \times \cdots \times d_n$.

  \begin{align*}
    &\mu^{(\ell)}, \; \text{with the same shape as } w^{(\ell)}, \\
    &Q_i^{(\ell)}\,(i=1,\dots,n), \; \text{ a lower-triangular matrix with shape } d_i \times d_i\\ 
        &\text{if }\gradnorm > 0: 
           \hat g_t = g_t \max\{1, \frac{\gradnorm}{\|g_t\|_2}\}, \text{else } \hat g_t = g_t\\
        &\text{if }\mathrm{normalize\_grads}: 
           \hat g_t = \frac{\hat g_t}{\|\hat g_t\|_2 + \epsilon}, \text{else } \hat g_t = \hat g_t
        \mu_t \leftarrow \beta_{1}\,\mu_{t-1} + (1-\beta_{1})\,\hat g_t,\quad
        \hat\mu_t \leftarrow \frac{\mu_t}{1-\beta_{1}^{\,t}}.
      \end{align*}
     \textbf{Balance check:}  $\mathrm{bal\_ctr}\leftarrow \mathrm{bal\_ctr}+1$.If $\mathrm{bal\_ctr}\ge100$, then 
          $Q_i^{(\ell)}\leftarrow \mathrm{balance}(Q_i^{(\ell)})$, 
          $\mathrm{bal\_ctr}\leftarrow0$.
          \textbf{Balance function:} $\mathrm{balance}(Q_i^{(\ell)})$ performs the following steps:
          \begin{enumerate}[left=1em]
            \item Compute the maximum absolute value for each row/column of $Q_i^{(\ell)}$: $\text{norms}_i = \max_{j,k} |Q_i^{(\ell)}[j,k]|$
            \item Calculate the geometric mean: $\text{gmean}_i = (\Pi_{j=1}^{n} \text{norms}_i[j])^{\frac{1}{d_i}}$
            \item Scale each element: $Q_i^{(\ell)} \leftarrow Q_i^{(\ell)} \cdot \frac{\text{gmean}_i}{\text{norms}_i}$
            \item Return the balanced $Q_i^{(\ell)}$
          \end{enumerate}
      \textbf{Preconditioner update:} with probability $p_{\mathrm{upd}}(t)$,
      \begin{enumerate}[left=1em]
        \item \textbf{Random probe:} 
          $V^{(\ell)}\leftarrow \mathrm{tree\_random\_like}(g_t^{(\ell)})$.
        \item \textbf{Dampen:}
          $\hat\mu_t^{(\ell)} \leftarrow \hat\mu_t^{(\ell)} + \epsilon \cdot \mathrm{mean}(|\hat\mu_t^{(\ell)}|)\cdot V^{(\ell)}$
        \item \textbf{Conjugate sketch ($B$):}
          \[
            X^{(0)}=V^{(\ell)},\quad
            X^{(i)}=\text{fold}_i\left(\left(Q_i^{(\ell)}\right)^{-T}\,\text{unfold}_i(X^{(i-1)})\right) ,\quad
            B=X^{(n)}.
          \]
        \item \textbf{Pre‐sketch ($A$):}
          \[
            Y^{(0)}=\hat\mu^{(\ell)},\quad
            Y^{(i)}=\text{fold}_i\left(Q_i^{(\ell)}\,\text{unfold}_i(Y^{(i-1)})\right) ,\quad
            A=Y^{(n)}.
          \]
        \item For each $i$ in $1,\dots,n$,
          \[
            M_i = \mathrm{unfold}_i(A),\quad
            C_i = \mathrm{unfold}_i(B),
          \]
          \[
            T_1 = M_i M_i^{T},\quad
            T_2 = C_i C_i^{T},\quad
            s = \max_{u,v}\lvert (T_1+T_2)_{u,v}\rvert,
          \]
          \[
            Q_i^{(\ell)} \;\leftarrow\;
            Q_i^{(\ell)} \;-\;\alpha
            \,\frac{T_1 - T_2}{s}\;Q_i^{(\ell)}.
          \]
      \end{enumerate}
      \[
        G^{(0)} = \hat\mu^{(\ell)},\quad
        G^{(i)} = \mathrm{fold}_i\bigl(Q_i^{(\ell)}\,\mathrm{unfold}_i(G^{(i-1)})\bigr),\quad
        \tilde g_t^{(\ell)} = G^{(d)}.
      \]
    \textbf{(Weight‐decay \& update)}  
      \[
        \tilde g_t^{(\ell)} \leftarrow \tilde g_t^{(\ell)} + \lambda\,w_{t-1}^{(\ell)},\quad
        w_t^{(\ell)} \leftarrow w_{t-1}^{(\ell)} - \eta\,\tilde g_t^{(\ell)}.
      \]
  \end{tcolorbox}
\end{algorithm}

\begin{algorithm}[H]
  \caption{SOAP}
  \label{alg:soap}
  \begin{tcolorbox}
  \textbf{Hyperparameters:} $\beta_1,\;\beta_2,\;\mu,\;k,\;\epsilon,\text{block\_size}, \gradnorm$
  
  \textbf{State:} $m,\;v,\;G_A,\;G_B,\;Q_A,\;Q_B$
  
  Partition all parameters into $\text{block\_size} \times \text{block\_size}$ block
  
  \textbf{Update Rule For Each Block:}
  \begin{align*}
  &\hat g_t = g_t \max\{1, \frac{\gradnorm}{\|g_t\|_2}\} \\
  &\hat g_t = Q_A\,\hat g_t\,Q_B,\\
  &m_t = \beta_1\,m_{t-1} + (1-\beta_1)\,\hat g_t,\quad
   v_t = \beta_2\,v_{t-1} + (1-\beta_2)\,\hat g_t^2,\\
  &\hat m_t = \frac{m_t}{1-\beta_1^{t}},\quad
   \hat v_t = \frac{v_t}{1-\beta_2^{t}},\\
  &w_{t+1} = w_t - \eta_t\;Q_A^\top\!
     \Bigl(\frac{\hat m_t}{\sqrt{\hat v_t} + \epsilon}\Bigr)\!Q_B^\top,\\
  &G_A = \mu\,G_A + (1-\mu)\,\hat g_t\,\hat g_t^\top + \epsilon I,\quad
  G_B = \mu\,G_B + (1-\mu)\,\hat g_t^\top\,\hat g_t + \epsilon I,\\
  &\text{if }t \bmod k = 0: \quad Q_A = \mathrm{QR}(G_A\,Q_A),\quad
     Q_B = \mathrm{QR}(G_B\,Q_B).
  \end{align*}
  \end{tcolorbox}
\end{algorithm}

\section{Omitted Experiments}
\label{app:experiments}

\subsection{Scaling Law}
\label{app:scaling_law}

Based on~\Cref{app:detailed_experiments,app:detailed_experiments_2}, we fitted our scaling law for the 1.2B run of Muon, NAdamW and AdamW, we round the fitted value to our grid of hyperparameter and used hyperparameters are shown in~\Cref{app:detailed_experiments_3}.
After training the models, we fitted a scaling law for Muon and AdamW of the following form:
\begin{equation}
    L(N,D) = \alpha N^{-A} + \beta D^{{-B}} + \gamma
\end{equation}
The fitted values are 
\begin{itemize}
    \item For AdamW, $  \alpha = 21.4289,
  A = 0.1555,
  \beta = 276.4235,
  B = 0.2804,
  \gamma = 1.7324$, with a RMS error of $3 \times 10^{-3}$.
    \item For Muon   $\alpha = 32.7458, A = 0.1864, \beta = 59.0221, B = 0.2074,\gamma = 1.8063$ , with a RMS error of $5 \times 10^{-3}$.
\end{itemize}

This scaling law predicts that when the parameter scale reaches 7B, Muon will actually result in a higher loss compared to AdamW in 1$\times$ Chinchilla regime.

\subsection{Sophia Experiments}
\label{app:sophia_experiments}

We performed a Phase I experiment with Sophia, with detailed hyperparameter settings shown in~\Cref{app:detailed_experiments}. We found that Sophia tends to underperform AdamW in smaller compute regimes and eventually slightly outperforms AdamW when either the model size or the data size increases.

\begin{figure}[H]
\centering
    \begin{minipage}{0.45\textwidth}
\includegraphics[width=\linewidth]{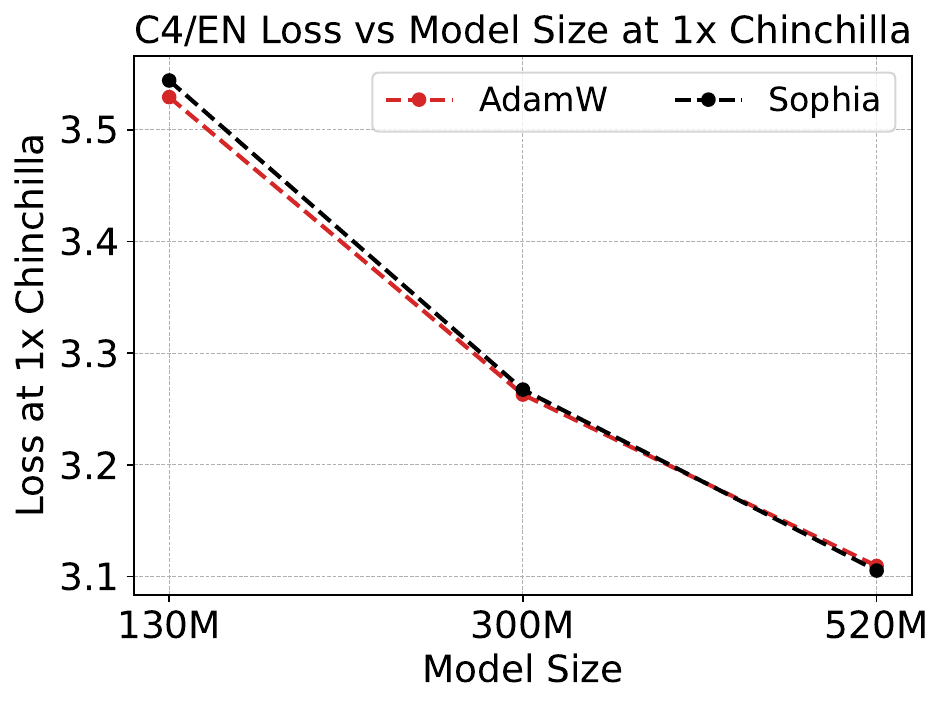}
    \end{minipage}
    \begin{minipage}{0.45\textwidth}
        \includegraphics[width=\linewidth]{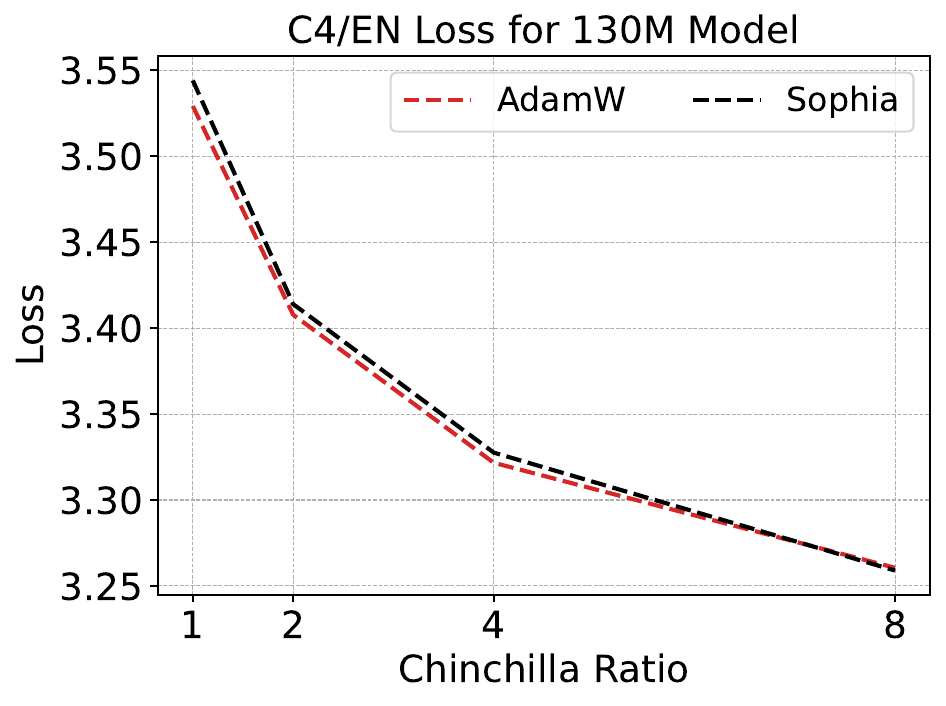}
            \end{minipage}
    \caption{\textbf{Sophia Experiments.} Left: Loss curve of Sophia and AdamW in $1\times$ Chinchilla setting. Right: Loss curve of Sophia and AdamW for 130M model size.}
    \vspace{-0.2in}
    \label{fig:sophia_experiments}
\end{figure}

\subsection{High Data-to-model Ratio}
\label{app:more_overtrained}

\begin{figure}[H]
\centering
    \begin{minipage}{0.45\textwidth}
\includegraphics[width=\linewidth]{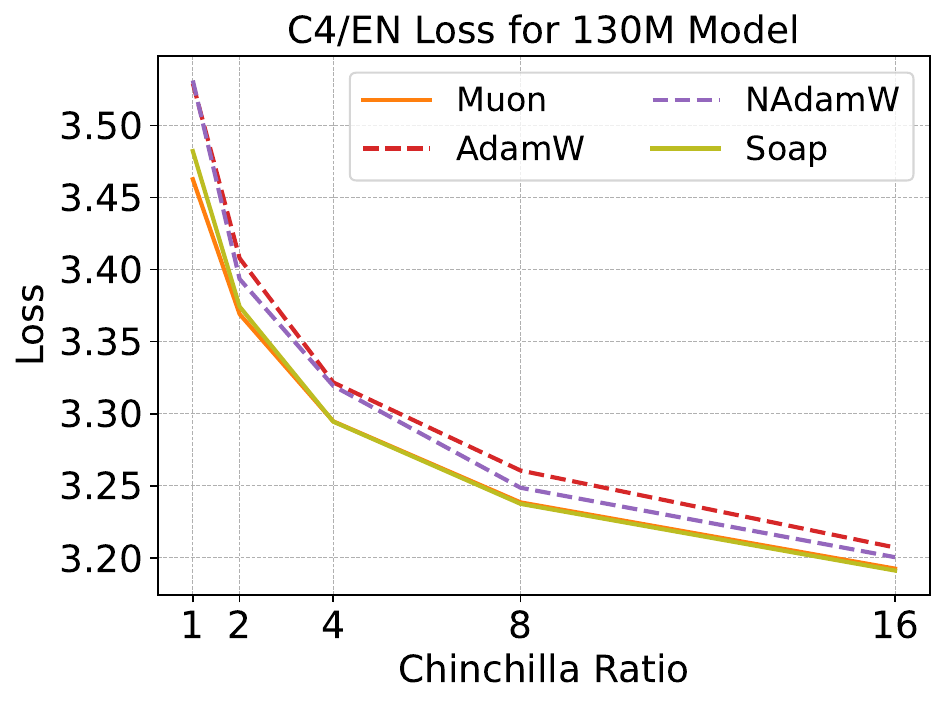}
    \end{minipage}
    \caption{\textbf{More Case Studies.} Experiment with 130M $16 \times$ Chinchilla setting, SOAP outperforms Muon in the overtraining setting.}
    \vspace{-0.2in}
    \label{fig:more_case}
\end{figure}

\subsection{Evaluation Performance}

\begin{table}
[H] \centering \caption{Evaluation Performance for Mars, Model Size = 130m}
\label{tab:eval-performance-mars-130M}
\begin{tabular}{lrrrr}
\toprule
DATA SIZE & 2B & 5B & 10B & 21B \\
PERFORMANCE METRIC &  &  &  &  \\
\midrule
LAMBADA OPENAI & 0.327 & 0.377 & 0.402 & 0.451 \\
OPENBOOKQA & 0.274 & 0.304 & 0.298 & 0.302 \\
WINOGRANDE & 0.521 & 0.512 & 0.526 & 0.522 \\
PIQA & 0.625 & 0.652 & 0.662 & 0.678 \\
BOOLQ & 0.616 & 0.504 & 0.569 & 0.557 \\
WSC273 & 0.557 & 0.560 & 0.579 & 0.590 \\
HELLASWAG 0SHOT & 0.308 & 0.343 & 0.365 & 0.399 \\
ARC CHALLENGE & 0.225 & 0.258 & 0.272 & 0.247 \\
ARC EASY & 0.462 & 0.517 & 0.532 & 0.544 \\
COPA & 0.650 & 0.690 & 0.660 & 0.710 \\
FINAL C4 LOSS & 3.537 & 3.396 & 3.323 & 3.247 \\
\bottomrule
\end{tabular}
\end{table}

\begin{table}
[H] \centering \caption{Evaluation Performance for Mars, Model Size = 300m}
\label{tab:eval-performance-mars-300M}
\begin{tabular}{lrrrr}
\toprule
DATA SIZE & 6B & 12B & 24B & 48B \\
PERFORMANCE METRIC &  &  &  &  \\
\midrule
LAMBADA OPENAI & 0.450 & 0.493 & 0.514 & 0.538 \\
OPENBOOKQA & 0.300 & 0.312 & 0.328 & 0.328 \\
WINOGRANDE & 0.534 & 0.557 & 0.549 & 0.561 \\
PIQA & 0.680 & 0.686 & 0.695 & 0.720 \\
BOOLQ & 0.480 & 0.533 & 0.612 & 0.608 \\
WSC273 & 0.586 & 0.608 & 0.667 & 0.667 \\
HELLASWAG 0SHOT & 0.395 & 0.437 & 0.471 & 0.502 \\
ARC CHALLENGE & 0.253 & 0.269 & 0.298 & 0.307 \\
ARC EASY & 0.547 & 0.580 & 0.608 & 0.616 \\
COPA & 0.700 & 0.740 & 0.720 & 0.730 \\
FINAL C4 LOSS & 3.249 & 3.158 & 3.097 & 3.040 \\
\bottomrule
\end{tabular}
\end{table}

\begin{table}
[H] \centering \caption{Evaluation Performance for Mars, Model Size = 520m}
\label{tab:eval-performance-mars-520M}
\begin{tabular}{lrrrr}
\toprule
DATA SIZE & 10B & 21B & 42B & 85B \\
PERFORMANCE METRIC &  &  &  &  \\
\midrule
LAMBADA OPENAI & 0.513 & 0.548 & 0.576 & 0.601 \\
OPENBOOKQA & 0.330 & 0.336 & 0.366 & 0.378 \\
WINOGRANDE & 0.548 & 0.559 & 0.586 & 0.601 \\
PIQA & 0.704 & 0.717 & 0.725 & 0.735 \\
BOOLQ & 0.559 & 0.598 & 0.625 & 0.631 \\
WSC273 & 0.601 & 0.696 & 0.736 & 0.780 \\
HELLASWAG 0SHOT & 0.462 & 0.513 & 0.546 & 0.579 \\
ARC CHALLENGE & 0.294 & 0.317 & 0.333 & 0.351 \\
ARC EASY & 0.611 & 0.648 & 0.665 & 0.676 \\
COPA & 0.730 & 0.760 & 0.750 & 0.750 \\
FINAL C4 LOSS & 3.101 & 3.015 & 2.955 & 2.906 \\
\bottomrule
\end{tabular}
\end{table}

\begin{table}
[H] \centering \caption{Evaluation Performance for Muon, Model Size = 130m}
\label{tab:eval-performance-muon-130M}
\begin{tabular}{lrrrr}
\toprule
DATA SIZE & 2B & 5B & 10B & 21B \\
PERFORMANCE METRIC &  &  &  &  \\
\midrule
LAMBADA OPENAI & 0.358 & 0.396 & 0.431 & 0.456 \\
OPENBOOKQA & 0.286 & 0.288 & 0.318 & 0.300 \\
WINOGRANDE & 0.515 & 0.518 & 0.530 & 0.541 \\
PIQA & 0.640 & 0.660 & 0.670 & 0.678 \\
BOOLQ & 0.571 & 0.504 & 0.583 & 0.452 \\
WSC273 & 0.553 & 0.564 & 0.579 & 0.582 \\
HELLASWAG 0SHOT & 0.330 & 0.354 & 0.381 & 0.407 \\
ARC CHALLENGE & 0.231 & 0.244 & 0.269 & 0.276 \\
ARC EASY & 0.493 & 0.511 & 0.543 & 0.559 \\
COPA & 0.650 & 0.720 & 0.730 & 0.660 \\
FINAL C4 LOSS & 3.464 & 3.369 & 3.296 & 3.240 \\
\bottomrule
\end{tabular}
\end{table}

\begin{table}
[H] \centering \caption{Evaluation Performance for Muon, Model Size = 300m}
\label{tab:eval-performance-muon-300M}
\begin{tabular}{lrrrr}
\toprule
DATA SIZE & 6B & 12B & 24B & 48B \\
PERFORMANCE METRIC &  &  &  &  \\
\midrule
LAMBADA OPENAI & 0.459 & 0.492 & 0.529 & 0.550 \\
OPENBOOKQA & 0.302 & 0.310 & 0.330 & 0.346 \\
WINOGRANDE & 0.504 & 0.534 & 0.548 & 0.570 \\
PIQA & 0.679 & 0.689 & 0.695 & 0.714 \\
BOOLQ & 0.482 & 0.555 & 0.599 & 0.570 \\
WSC273 & 0.615 & 0.641 & 0.659 & 0.656 \\
HELLASWAG 0SHOT & 0.406 & 0.447 & 0.481 & 0.509 \\
ARC CHALLENGE & 0.270 & 0.289 & 0.290 & 0.316 \\
ARC EASY & 0.564 & 0.591 & 0.620 & 0.638 \\
COPA & 0.670 & 0.730 & 0.700 & 0.700 \\
FINAL C4 LOSS & 3.224 & 3.143 & 3.079 & 3.032 \\
\bottomrule
\end{tabular}
\end{table}

\begin{table}
[H] \centering \caption{Evaluation Performance for Muon, Model Size = 520m}
\label{tab:eval-performance-muon-520M}
\begin{tabular}{lrrrr}
\toprule
DATA SIZE & 10B & 21B & 42B & 85B \\
PERFORMANCE METRIC &  &  &  &  \\
\midrule
LAMBADA OPENAI & 0.520 & 0.559 & 0.597 & 0.611 \\
OPENBOOKQA & 0.328 & 0.340 & 0.342 & 0.368 \\
WINOGRANDE & 0.560 & 0.567 & 0.594 & 0.590 \\
PIQA & 0.713 & 0.718 & 0.730 & 0.736 \\
BOOLQ & 0.590 & 0.555 & 0.633 & 0.613 \\
WSC273 & 0.659 & 0.667 & 0.762 & 0.736 \\
HELLASWAG 0SHOT & 0.482 & 0.525 & 0.554 & 0.587 \\
ARC CHALLENGE & 0.300 & 0.318 & 0.351 & 0.348 \\
ARC EASY & 0.623 & 0.640 & 0.671 & 0.674 \\
COPA & 0.730 & 0.760 & 0.730 & 0.780 \\
FINAL C4 LOSS & 3.073 & 3.002 & 2.945 & 2.900 \\
\bottomrule
\end{tabular}
\end{table}

\begin{table}
[H] \centering \caption{Evaluation Performance for Lion, Model Size = 130m}
\label{tab:eval-performance-lion-130M}
\begin{tabular}{lrrrr}
\toprule
DATA SIZE & 2B & 5B & 10B & 21B \\
PERFORMANCE METRIC &  &  &  &  \\
\midrule
LAMBADA OPENAI & 0.316 & 0.380 & 0.417 & 0.450 \\
OPENBOOKQA & 0.280 & 0.296 & 0.312 & 0.292 \\
WINOGRANDE & 0.503 & 0.518 & 0.519 & 0.521 \\
PIQA & 0.629 & 0.649 & 0.663 & 0.680 \\
BOOLQ & 0.603 & 0.507 & 0.419 & 0.489 \\
WSC273 & 0.564 & 0.538 & 0.590 & 0.623 \\
HELLASWAG 0SHOT & 0.304 & 0.343 & 0.366 & 0.402 \\
ARC CHALLENGE & 0.229 & 0.272 & 0.262 & 0.272 \\
ARC EASY & 0.464 & 0.500 & 0.536 & 0.539 \\
COPA & 0.700 & 0.670 & 0.730 & 0.730 \\
FINAL C4 LOSS & 3.552 & 3.409 & 3.331 & 3.252 \\
\bottomrule
\end{tabular}
\end{table}

\begin{table}
[H] \centering \caption{Evaluation Performance for Lion, Model Size = 300m}
\label{tab:eval-performance-lion-300M}
\begin{tabular}{lrrrr}
\toprule
DATA SIZE & 6B & 12B & 24B & 48B \\
PERFORMANCE METRIC &  &  &  &  \\
\midrule
LAMBADA OPENAI & 0.428 & 0.471 & 0.509 & 0.535 \\
OPENBOOKQA & 0.306 & 0.320 & 0.326 & 0.346 \\
WINOGRANDE & 0.506 & 0.530 & 0.532 & 0.564 \\
PIQA & 0.676 & 0.686 & 0.699 & 0.713 \\
BOOLQ & 0.486 & 0.490 & 0.609 & 0.593 \\
WSC273 & 0.579 & 0.615 & 0.608 & 0.667 \\
HELLASWAG 0SHOT & 0.386 & 0.428 & 0.466 & 0.502 \\
ARC CHALLENGE & 0.267 & 0.296 & 0.294 & 0.322 \\
ARC EASY & 0.535 & 0.586 & 0.600 & 0.618 \\
COPA & 0.690 & 0.730 & 0.710 & 0.760 \\
FINAL C4 LOSS & 3.268 & 3.170 & 3.100 & 3.046 \\
\bottomrule
\end{tabular}
\end{table}

\begin{table}
[H] \centering \caption{Evaluation Performance for Lion, Model Size = 520m}
\label{tab:eval-performance-lion-520M}
\begin{tabular}{lrrrr}
\toprule
DATA SIZE & 10B & 21B & 42B & 85B \\
PERFORMANCE METRIC &  &  &  &  \\
\midrule
LAMBADA OPENAI & 0.507 & 0.543 & 0.569 & 0.597 \\
OPENBOOKQA & 0.336 & 0.334 & 0.356 & 0.358 \\
WINOGRANDE & 0.560 & 0.548 & 0.581 & 0.590 \\
PIQA & 0.699 & 0.713 & 0.726 & 0.740 \\
BOOLQ & 0.600 & 0.546 & 0.612 & 0.631 \\
WSC273 & 0.608 & 0.667 & 0.685 & 0.700 \\
HELLASWAG 0SHOT & 0.463 & 0.509 & 0.545 & 0.571 \\
ARC CHALLENGE & 0.303 & 0.320 & 0.340 & 0.371 \\
ARC EASY & 0.603 & 0.638 & 0.664 & 0.673 \\
COPA & 0.710 & 0.750 & 0.740 & 0.770 \\
FINAL C4 LOSS & 3.108 & 3.029 & 2.965 & 2.915 \\
\bottomrule
\end{tabular}
\end{table}

\begin{table}
[H] \centering \caption{Evaluation Performance for NAdamW, Model Size = 130m}
\label{tab:eval-performance-nadamw-130M}
\begin{tabular}{lrrrr}
\toprule
DATA SIZE & 2B & 5B & 10B & 21B \\
PERFORMANCE METRIC &  &  &  &  \\
\midrule
LAMBADA OPENAI & 0.325 & 0.383 & 0.413 & 0.446 \\
OPENBOOKQA & 0.284 & 0.286 & 0.308 & 0.322 \\
WINOGRANDE & 0.515 & 0.507 & 0.546 & 0.507 \\
PIQA & 0.633 & 0.652 & 0.667 & 0.682 \\
BOOLQ & 0.591 & 0.569 & 0.459 & 0.557 \\
WSC273 & 0.535 & 0.560 & 0.612 & 0.597 \\
HELLASWAG 0SHOT & 0.312 & 0.348 & 0.368 & 0.399 \\
ARC CHALLENGE & 0.239 & 0.245 & 0.242 & 0.270 \\
ARC EASY & 0.465 & 0.503 & 0.518 & 0.547 \\
COPA & 0.680 & 0.700 & 0.650 & 0.720 \\
FINAL C4 LOSS & 3.531 & 3.394 & 3.319 & 3.251 \\
\bottomrule
\end{tabular}
\end{table}

\begin{table}
[H] \centering \caption{Evaluation Performance for NAdamW, Model Size = 300m}
\label{tab:eval-performance-nadamw-300M}
\begin{tabular}{lrrrr}
\toprule
DATA SIZE & 6B & 12B & 24B & 48B \\
PERFORMANCE METRIC &  &  &  &  \\
\midrule
LAMBADA OPENAI & 0.437 & 0.477 & 0.514 & 0.543 \\
OPENBOOKQA & 0.310 & 0.332 & 0.316 & 0.348 \\
WINOGRANDE & 0.520 & 0.536 & 0.549 & 0.564 \\
PIQA & 0.681 & 0.693 & 0.708 & 0.712 \\
BOOLQ & 0.539 & 0.564 & 0.520 & 0.612 \\
WSC273 & 0.582 & 0.637 & 0.645 & 0.700 \\
HELLASWAG 0SHOT & 0.397 & 0.434 & 0.475 & 0.505 \\
ARC CHALLENGE & 0.264 & 0.275 & 0.292 & 0.307 \\
ARC EASY & 0.553 & 0.589 & 0.605 & 0.632 \\
COPA & 0.710 & 0.730 & 0.700 & 0.750 \\
FINAL C4 LOSS & 3.248 & 3.160 & 3.090 & 3.039 \\
\bottomrule
\end{tabular}
\end{table}

\begin{table}
[H] \centering \caption{Evaluation Performance for NAdamW, Model Size = 520m}
\label{tab:eval-performance-nadamw-520M}
\begin{tabular}{lrrrr}
\toprule
DATA SIZE & 10B & 21B & 42B & 85B \\
PERFORMANCE METRIC &  &  &  &  \\
\midrule
LAMBADA OPENAI & 0.500 & 0.551 & 0.577 & 0.599 \\
OPENBOOKQA & 0.332 & 0.350 & 0.358 & 0.342 \\
WINOGRANDE & 0.553 & 0.571 & 0.594 & 0.597 \\
PIQA & 0.695 & 0.719 & 0.731 & 0.738 \\
BOOLQ & 0.613 & 0.606 & 0.639 & 0.618 \\
WSC273 & 0.637 & 0.689 & 0.718 & 0.714 \\
HELLASWAG 0SHOT & 0.469 & 0.513 & 0.550 & 0.580 \\
ARC CHALLENGE & 0.294 & 0.322 & 0.342 & 0.354 \\
ARC EASY & 0.605 & 0.646 & 0.657 & 0.674 \\
COPA & 0.700 & 0.720 & 0.760 & 0.780 \\
FINAL C4 LOSS & 3.100 & 3.013 & 2.955 & 2.907 \\
\bottomrule
\end{tabular}
\end{table}

\begin{table}
[H] \centering \caption{Evaluation Performance for Kron, Model Size = 130m}
\label{tab:eval-performance-kron-130M}
\begin{tabular}{lrrrr}
\toprule
DATA SIZE & 2B & 5B & 10B & 21B \\
PERFORMANCE METRIC &  &  &  &  \\
\midrule
LAMBADA OPENAI & 0.341 & 0.387 & 0.426 & 0.451 \\
OPENBOOKQA & 0.268 & 0.278 & 0.296 & 0.314 \\
WINOGRANDE & 0.520 & 0.522 & 0.515 & 0.543 \\
PIQA & 0.636 & 0.662 & 0.662 & 0.683 \\
BOOLQ & 0.533 & 0.530 & 0.575 & 0.557 \\
WSC273 & 0.564 & 0.564 & 0.553 & 0.604 \\
HELLASWAG 0SHOT & 0.323 & 0.347 & 0.374 & 0.397 \\
ARC CHALLENGE & 0.235 & 0.259 & 0.256 & 0.281 \\
ARC EASY & 0.487 & 0.519 & 0.543 & 0.572 \\
COPA & 0.700 & 0.690 & 0.710 & 0.710 \\
FINAL C4 LOSS & 3.492 & 3.389 & 3.307 & 3.239 \\
\bottomrule
\end{tabular}
\end{table}

\begin{table}
[H] \centering \caption{Evaluation Performance for Kron, Model Size = 300m}
\label{tab:eval-performance-kron-300M}
\begin{tabular}{lrrrr}
\toprule
DATA SIZE & 6B & 12B & 24B & 48B \\
PERFORMANCE METRIC &  &  &  &  \\
\midrule
LAMBADA OPENAI & 0.429 & 0.485 & 0.516 & 0.548 \\
OPENBOOKQA & 0.310 & 0.314 & 0.314 & 0.326 \\
WINOGRANDE & 0.527 & 0.534 & 0.542 & 0.571 \\
PIQA & 0.675 & 0.690 & 0.706 & 0.709 \\
BOOLQ & 0.487 & 0.475 & 0.610 & 0.611 \\
WSC273 & 0.546 & 0.619 & 0.659 & 0.718 \\
HELLASWAG 0SHOT & 0.396 & 0.438 & 0.479 & 0.501 \\
ARC CHALLENGE & 0.262 & 0.285 & 0.298 & 0.323 \\
ARC EASY & 0.551 & 0.586 & 0.618 & 0.632 \\
COPA & 0.670 & 0.720 & 0.720 & 0.750 \\
FINAL C4 LOSS & 3.244 & 3.151 & 3.083 & 3.031 \\
\bottomrule
\end{tabular}
\end{table}

\begin{table}
[H] \centering \caption{Evaluation Performance for Kron, Model Size = 520m}
\label{tab:eval-performance-kron-520M}
\begin{tabular}{lrrrr}
\toprule
DATA SIZE & 10B & 21B & 42B & 85B \\
PERFORMANCE METRIC &  &  &  &  \\
\midrule
LAMBADA OPENAI & 0.512 & 0.549 & 0.569 & 0.603 \\
OPENBOOKQA & 0.338 & 0.332 & 0.350 & 0.352 \\
WINOGRANDE & 0.536 & 0.567 & 0.575 & 0.601 \\
PIQA & 0.721 & 0.713 & 0.724 & 0.739 \\
BOOLQ & 0.529 & 0.560 & 0.537 & 0.559 \\
WSC273 & 0.648 & 0.663 & 0.696 & 0.762 \\
HELLASWAG 0SHOT & 0.474 & 0.516 & 0.556 & 0.582 \\
ARC CHALLENGE & 0.287 & 0.312 & 0.344 & 0.353 \\
ARC EASY & 0.616 & 0.648 & 0.662 & 0.681 \\
COPA & 0.710 & 0.720 & 0.790 & 0.750 \\
FINAL C4 LOSS & 3.084 & 3.009 & 2.946 & 2.900 \\
\bottomrule
\end{tabular}
\end{table}

\begin{table}
[H] \centering \caption{Evaluation Performance for Scion, Model Size = 130m}
\label{tab:eval-performance-scion-130M}
\begin{tabular}{lrrrr}
\toprule
DATA SIZE & 2B & 5B & 10B & 21B \\
PERFORMANCE METRIC &  &  &  &  \\
\midrule
LAMBADA OPENAI & 0.355 & 0.382 & 0.427 & 0.446 \\
OPENBOOKQA & 0.282 & 0.300 & 0.288 & 0.314 \\
WINOGRANDE & 0.515 & 0.502 & 0.502 & 0.535 \\
PIQA & 0.641 & 0.655 & 0.675 & 0.684 \\
BOOLQ & 0.574 & 0.518 & 0.532 & 0.474 \\
WSC273 & 0.505 & 0.531 & 0.553 & 0.608 \\
HELLASWAG 0SHOT & 0.323 & 0.354 & 0.374 & 0.401 \\
ARC CHALLENGE & 0.247 & 0.241 & 0.276 & 0.264 \\
ARC EASY & 0.501 & 0.513 & 0.541 & 0.557 \\
COPA & 0.650 & 0.660 & 0.660 & 0.680 \\
FINAL C4 LOSS & 3.477 & 3.379 & 3.302 & 3.246 \\
\bottomrule
\end{tabular}
\end{table}

\begin{table}
[H] \centering \caption{Evaluation Performance for Scion, Model Size = 300m}
\label{tab:eval-performance-scion-300M}
\begin{tabular}{lrrrr}
\toprule
DATA SIZE & 6B & 12B & 24B & 48B \\
PERFORMANCE METRIC &  &  &  &  \\
\midrule
LAMBADA OPENAI & 0.445 & 0.489 & 0.511 & 0.540 \\
OPENBOOKQA & 0.306 & 0.308 & 0.326 & 0.358 \\
WINOGRANDE & 0.531 & 0.550 & 0.555 & 0.566 \\
PIQA & 0.684 & 0.689 & 0.701 & 0.712 \\
BOOLQ & 0.576 & 0.610 & 0.574 & 0.585 \\
WSC273 & 0.597 & 0.612 & 0.685 & 0.696 \\
HELLASWAG 0SHOT & 0.406 & 0.442 & 0.478 & 0.506 \\
ARC CHALLENGE & 0.272 & 0.285 & 0.298 & 0.316 \\
ARC EASY & 0.564 & 0.592 & 0.625 & 0.633 \\
COPA & 0.680 & 0.730 & 0.700 & 0.740 \\
FINAL C4 LOSS & 3.232 & 3.152 & 3.086 & 3.039 \\
\bottomrule
\end{tabular}
\end{table}

\begin{table}
[H] \centering \caption{Evaluation Performance for Scion, Model Size = 520m}
\label{tab:eval-performance-scion-520M}
\begin{tabular}{lrrrr}
\toprule
DATA SIZE & 10B & 21B & 42B & 85B \\
PERFORMANCE METRIC &  &  &  &  \\
\midrule
LAMBADA OPENAI & 0.521 & 0.552 & 0.575 & 0.601 \\
OPENBOOKQA & 0.338 & 0.362 & 0.366 & 0.374 \\
WINOGRANDE & 0.569 & 0.551 & 0.605 & 0.586 \\
PIQA & 0.704 & 0.717 & 0.736 & 0.743 \\
BOOLQ & 0.575 & 0.615 & 0.620 & 0.641 \\
WSC273 & 0.674 & 0.692 & 0.736 & 0.736 \\
HELLASWAG 0SHOT & 0.481 & 0.520 & 0.549 & 0.581 \\
ARC CHALLENGE & 0.289 & 0.314 & 0.346 & 0.354 \\
ARC EASY & 0.625 & 0.647 & 0.667 & 0.686 \\
COPA & 0.730 & 0.730 & 0.750 & 0.770 \\
FINAL C4 LOSS & 3.080 & 3.007 & 2.952 & 2.904 \\
\bottomrule
\end{tabular}
\end{table}

\begin{table}
[H] \centering \caption{Evaluation Performance for Cautious, Model Size = 130m}
\label{tab:eval-performance-cautious-130M}
\begin{tabular}{lrrrr}
\toprule
DATA SIZE & 2B & 5B & 10B & 21B \\
PERFORMANCE METRIC &  &  &  &  \\
\midrule
LAMBADA OPENAI & 0.314 & 0.385 & 0.420 & 0.458 \\
OPENBOOKQA & 0.298 & 0.280 & 0.308 & 0.318 \\
WINOGRANDE & 0.515 & 0.520 & 0.519 & 0.508 \\
PIQA & 0.638 & 0.652 & 0.664 & 0.669 \\
BOOLQ & 0.580 & 0.512 & 0.512 & 0.550 \\
WSC273 & 0.560 & 0.527 & 0.593 & 0.604 \\
HELLASWAG 0SHOT & 0.314 & 0.347 & 0.371 & 0.401 \\
ARC CHALLENGE & 0.239 & 0.244 & 0.261 & 0.270 \\
ARC EASY & 0.476 & 0.508 & 0.532 & 0.544 \\
COPA & 0.690 & 0.720 & 0.710 & 0.690 \\
FINAL C4 LOSS & 3.535 & 3.403 & 3.334 & 3.253 \\
\bottomrule
\end{tabular}
\end{table}

\begin{table}
[H] \centering \caption{Evaluation Performance for Cautious, Model Size = 300m}
\label{tab:eval-performance-cautious-300M}
\begin{tabular}{lrrrr}
\toprule
DATA SIZE & 6B & 12B & 24B & 48B \\
PERFORMANCE METRIC &  &  &  &  \\
\midrule
LAMBADA OPENAI & 0.437 & 0.479 & 0.510 & 0.544 \\
OPENBOOKQA & 0.316 & 0.314 & 0.326 & 0.328 \\
WINOGRANDE & 0.515 & 0.533 & 0.530 & 0.576 \\
PIQA & 0.677 & 0.699 & 0.707 & 0.711 \\
BOOLQ & 0.547 & 0.492 & 0.584 & 0.574 \\
WSC273 & 0.568 & 0.608 & 0.634 & 0.667 \\
HELLASWAG 0SHOT & 0.397 & 0.432 & 0.477 & 0.505 \\
ARC CHALLENGE & 0.267 & 0.280 & 0.292 & 0.308 \\
ARC EASY & 0.541 & 0.569 & 0.606 & 0.618 \\
COPA & 0.670 & 0.730 & 0.760 & 0.720 \\
FINAL C4 LOSS & 3.260 & 3.165 & 3.094 & 3.043 \\
\bottomrule
\end{tabular}
\end{table}

\begin{table}
[H] \centering \caption{Evaluation Performance for Cautious, Model Size = 520m}
\label{tab:eval-performance-cautious-520M}
\begin{tabular}{lrrrr}
\toprule
DATA SIZE & 10B & 21B & 42B & 85B \\
PERFORMANCE METRIC &  &  &  &  \\
\midrule
LAMBADA OPENAI & 0.509 & 0.549 & 0.578 & 0.604 \\
OPENBOOKQA & 0.318 & 0.338 & 0.342 & 0.364 \\
WINOGRANDE & 0.548 & 0.575 & 0.571 & 0.605 \\
PIQA & 0.705 & 0.715 & 0.733 & 0.739 \\
BOOLQ & 0.601 & 0.611 & 0.582 & 0.606 \\
WSC273 & 0.619 & 0.718 & 0.659 & 0.736 \\
HELLASWAG 0SHOT & 0.467 & 0.514 & 0.550 & 0.583 \\
ARC CHALLENGE & 0.305 & 0.315 & 0.344 & 0.358 \\
ARC EASY & 0.609 & 0.649 & 0.676 & 0.689 \\
COPA & 0.730 & 0.760 & 0.770 & 0.770 \\
FINAL C4 LOSS & 3.100 & 3.017 & 2.956 & 2.910 \\
\bottomrule
\end{tabular}
\end{table}

\begin{table}
[H] \centering \caption{Evaluation Performance for SOAP, Model Size = 130m}
\label{tab:eval-performance-soap-130M}
\begin{tabular}{lrrrr}
\toprule
DATA SIZE & 2B & 5B & 10B & 21B \\
PERFORMANCE METRIC &  &  &  &  \\
\midrule
LAMBADA OPENAI & 0.344 & 0.404 & 0.434 & 0.458 \\
OPENBOOKQA & 0.280 & 0.270 & 0.300 & 0.318 \\
WINOGRANDE & 0.511 & 0.510 & 0.523 & 0.517 \\
PIQA & 0.643 & 0.652 & 0.666 & 0.676 \\
BOOLQ & 0.569 & 0.491 & 0.519 & 0.559 \\
WSC273 & 0.513 & 0.568 & 0.597 & 0.612 \\
HELLASWAG 0SHOT & 0.321 & 0.351 & 0.376 & 0.401 \\
ARC CHALLENGE & 0.247 & 0.255 & 0.270 & 0.270 \\
ARC EASY & 0.475 & 0.519 & 0.537 & 0.571 \\
COPA & 0.690 & 0.710 & 0.720 & 0.710 \\
FINAL C4 LOSS & 3.487 & 3.376 & 3.295 & 3.240 \\
\bottomrule
\end{tabular}
\end{table}

\begin{table}[H]
\centering
\caption{Evaluation Performance for SOAP, Model Size = 300m}
\label{tab:eval-performance-soap-300M}
\begin{tabular}{lrrrr}
\toprule
DATA SIZE & 6B & 12B & 24B & 48B \\
PERFORMANCE METRIC &  &  &  &  \\
\midrule
LAMBADA OPENAI & 0.001 & 0.492 & 0.518 & 0.541 \\
OPENBOOKQA & 0.256 & 0.324 & 0.324 & 0.334 \\
WINOGRANDE & 0.506 & 0.534 & 0.546 & 0.555 \\
PIQA & 0.546 & 0.689 & 0.706 & 0.711 \\
BOOLQ & 0.402 & 0.593 & 0.565 & 0.593 \\
WSC273 & 0.498 & 0.619 & 0.703 & 0.670 \\
HELLASWAG 0SHOT & 0.257 & 0.443 & 0.478 & 0.506 \\
ARC CHALLENGE & 0.219 & 0.281 & 0.305 & 0.324 \\
ARC EASY & 0.309 & 0.590 & 0.612 & 0.632 \\
COPA & 0.590 & 0.740 & 0.720 & 0.700 \\
FINAL C4 LOSS & 5.437 & 3.147 & 3.082 & 3.030 \\
\bottomrule
\end{tabular}
\end{table}

\begin{table}
[H] \centering \caption{Evaluation Performance for SOAP, Model Size = 520m}
\label{tab:eval-performance-soap-520M}
\begin{tabular}{lrrrr}
\toprule
DATA SIZE & 10B & 21B & 42B & 85B \\
PERFORMANCE METRIC &  &  &  &  \\
\midrule
LAMBADA OPENAI & 0.513 & 0.555 & 0.582 & 0.606 \\
OPENBOOKQA & 0.308 & 0.312 & 0.368 & 0.350 \\
WINOGRANDE & 0.548 & 0.562 & 0.594 & 0.583 \\
PIQA & 0.695 & 0.714 & 0.729 & 0.739 \\
BOOLQ & 0.563 & 0.623 & 0.596 & 0.643 \\
WSC273 & 0.648 & 0.703 & 0.722 & 0.755 \\
HELLASWAG 0SHOT & 0.478 & 0.523 & 0.553 & 0.586 \\
ARC CHALLENGE & 0.292 & 0.345 & 0.334 & 0.356 \\
ARC EASY & 0.610 & 0.656 & 0.655 & 0.678 \\
COPA & 0.740 & 0.760 & 0.740 & 0.770 \\
FINAL C4 LOSS & 3.079 & 3.004 & 2.957 & 2.899 \\
\bottomrule
\end{tabular}
\end{table}

\begin{table}
[H] \centering \caption{Evaluation Performance for AdamW, Model Size = 130m}
\label{tab:eval-performance-adamw-130M}
\begin{tabular}{lrrrr}
\toprule
DATA SIZE & 2B & 5B & 10B & 21B \\
PERFORMANCE METRIC &  &  &  &  \\
\midrule
LAMBADA OPENAI & 0.322 & 0.380 & 0.410 & 0.442 \\
OPENBOOKQA & 0.306 & 0.290 & 0.298 & 0.292 \\
WINOGRANDE & 0.506 & 0.519 & 0.515 & 0.530 \\
PIQA & 0.632 & 0.663 & 0.665 & 0.674 \\
BOOLQ & 0.568 & 0.415 & 0.553 & 0.576 \\
WSC273 & 0.582 & 0.553 & 0.564 & 0.623 \\
HELLASWAG 0SHOT & 0.312 & 0.340 & 0.365 & 0.391 \\
ARC CHALLENGE & 0.229 & 0.255 & 0.262 & 0.270 \\
ARC EASY & 0.463 & 0.497 & 0.523 & 0.544 \\
COPA & 0.660 & 0.650 & 0.720 & 0.730 \\
FINAL C4 LOSS & 3.529 & 3.409 & 3.322 & 3.262 \\
\bottomrule
\end{tabular}
\end{table}

\begin{table}
[H] \centering \caption{Evaluation Performance for AdamW, Model Size = 300m}
\label{tab:eval-performance-adamw-300M}
\begin{tabular}{lrrrr}
\toprule
DATA SIZE & 6B & 12B & 24B & 48B \\
PERFORMANCE METRIC &  &  &  &  \\
\midrule
LAMBADA OPENAI & 0.435 & 0.484 & 0.500 & 0.537 \\
OPENBOOKQA & 0.300 & 0.312 & 0.328 & 0.318 \\
WINOGRANDE & 0.519 & 0.530 & 0.542 & 0.552 \\
PIQA & 0.673 & 0.691 & 0.697 & 0.713 \\
BOOLQ & 0.454 & 0.496 & 0.508 & 0.602 \\
WSC273 & 0.586 & 0.626 & 0.667 & 0.692 \\
HELLASWAG 0SHOT & 0.386 & 0.432 & 0.471 & 0.501 \\
ARC CHALLENGE & 0.269 & 0.295 & 0.304 & 0.310 \\
ARC EASY & 0.535 & 0.581 & 0.612 & 0.616 \\
COPA & 0.700 & 0.700 & 0.690 & 0.710 \\
FINAL C4 LOSS & 3.264 & 3.166 & 3.094 & 3.043 \\
\bottomrule
\end{tabular}
\end{table}

\begin{table}
[H] \centering \caption{Evaluation Performance for AdamW, Model Size = 520m}
\label{tab:eval-performance-adamw-520M}
\begin{tabular}{lrrrr}
\toprule
DATA SIZE & 10B & 21B & 42B & 85B \\
PERFORMANCE METRIC &  &  &  &  \\
\midrule
LAMBADA OPENAI & 0.500 & 0.540 & 0.570 & 0.591 \\
OPENBOOKQA & 0.320 & 0.336 & 0.342 & 0.356 \\
WINOGRANDE & 0.539 & 0.586 & 0.583 & 0.601 \\
PIQA & 0.705 & 0.719 & 0.732 & 0.739 \\
BOOLQ & 0.555 & 0.615 & 0.596 & 0.609 \\
WSC273 & 0.604 & 0.656 & 0.703 & 0.729 \\
HELLASWAG 0SHOT & 0.456 & 0.507 & 0.543 & 0.578 \\
ARC CHALLENGE & 0.299 & 0.311 & 0.323 & 0.349 \\
ARC EASY & 0.613 & 0.639 & 0.669 & 0.688 \\
COPA & 0.740 & 0.680 & 0.710 & 0.780 \\
FINAL C4 LOSS & 3.110 & 3.023 & 2.958 & 2.913 \\
\bottomrule
\end{tabular}
\end{table}

\begin{table}
[H] \centering \caption{Evaluation Performance for Adam-Mini, Model Size = 130m}
\label{tab:eval-performance-adam-mini-130M}
\begin{tabular}{lrrrr}
\toprule
DATA SIZE & 2B & 5B & 10B & 21B \\
PERFORMANCE METRIC &  &  &  &  \\
\midrule
LAMBADA OPENAI & 0.317 & 0.369 & 0.412 & 0.444 \\
OPENBOOKQA & 0.286 & 0.282 & 0.282 & 0.302 \\
WINOGRANDE & 0.525 & 0.500 & 0.515 & 0.528 \\
PIQA & 0.640 & 0.655 & 0.662 & 0.670 \\
BOOLQ & 0.506 & 0.496 & 0.445 & 0.583 \\
WSC273 & 0.571 & 0.557 & 0.546 & 0.634 \\
HELLASWAG 0SHOT & 0.310 & 0.337 & 0.364 & 0.390 \\
ARC CHALLENGE & 0.241 & 0.247 & 0.255 & 0.264 \\
ARC EASY & 0.465 & 0.495 & 0.525 & 0.555 \\
COPA & 0.690 & 0.710 & 0.730 & 0.700 \\
FINAL C4 LOSS & 3.542 & 3.416 & 3.328 & 3.266 \\
\bottomrule
\end{tabular}
\end{table}

\begin{table}
[H] \centering \caption{Evaluation Performance for Adam-Mini, Model Size = 300m}
\label{tab:eval-performance-adam-mini-300M}
\begin{tabular}{lrrrr}
\toprule
DATA SIZE & 6B & 12B & 24B & 48B \\
PERFORMANCE METRIC &  &  &  &  \\
\midrule
LAMBADA OPENAI & 0.429 & 0.468 & 0.509 & 0.535 \\
OPENBOOKQA & 0.300 & 0.318 & 0.318 & 0.350 \\
WINOGRANDE & 0.502 & 0.528 & 0.545 & 0.586 \\
PIQA & 0.688 & 0.693 & 0.683 & 0.701 \\
BOOLQ & 0.480 & 0.546 & 0.609 & 0.573 \\
WSC273 & 0.586 & 0.546 & 0.593 & 0.689 \\
HELLASWAG 0SHOT & 0.381 & 0.424 & 0.464 & 0.492 \\
ARC CHALLENGE & 0.270 & 0.292 & 0.294 & 0.300 \\
ARC EASY & 0.545 & 0.579 & 0.607 & 0.625 \\
COPA & 0.700 & 0.720 & 0.740 & 0.730 \\
FINAL C4 LOSS & 3.272 & 3.178 & 3.103 & 3.049 \\
\bottomrule
\end{tabular}
\end{table}

\begin{table}
[H] \centering \caption{Evaluation Performance for Adam-Mini, Model Size = 520m}
\label{tab:eval-performance-adam-mini-520M}
\begin{tabular}{lrrrr}
\toprule
DATA SIZE & 10B & 21B & 42B & 85B \\
PERFORMANCE METRIC &  &  &  &  \\
\midrule
LAMBADA OPENAI & 0.500 & 0.538 & 0.576 & 0.604 \\
OPENBOOKQA & 0.306 & 0.326 & 0.330 & 0.368 \\
WINOGRANDE & 0.534 & 0.561 & 0.571 & 0.594 \\
PIQA & 0.707 & 0.721 & 0.723 & 0.733 \\
BOOLQ & 0.543 & 0.626 & 0.619 & 0.536 \\
WSC273 & 0.619 & 0.700 & 0.733 & 0.744 \\
HELLASWAG 0SHOT & 0.459 & 0.505 & 0.541 & 0.576 \\
ARC CHALLENGE & 0.288 & 0.327 & 0.332 & 0.352 \\
ARC EASY & 0.613 & 0.638 & 0.654 & 0.671 \\
COPA & 0.710 & 0.720 & 0.760 & 0.760 \\
FINAL C4 LOSS & 3.112 & 3.027 & 2.966 & 2.912 \\
\bottomrule
\end{tabular}
\end{table}

\section{Hyperparameter Ablation in Phase I}
\label{app:detailed_experiments}

We reported the results for the optimizers we swept in Phase I. The result is formulated as follows: the first row shows the approximately best configuration found and the following rows show the results for the 1-dimensional ablations centered around the found configuration. The loss presented here is the final loss on the C4/EN validation set.

\subsection{Sweeping Results for AdamW}%
\begin{table}[H]
\centering
\caption{Hyperparameter ablation for AdamW on 130m on 1x Chinchilla Data}
\label{tab:ablation_adamw_130m_on_1x_chinchilla_data}
\begin{tabular}{ccccccccccc}
\toprule
$\beta_1$ & $\beta_2$ & $\epsilon$ & $\eta$ & $\gradnorm$ & $\eta_{min}$ & $\mathrm{BSZ}$ & $\mathrm{warmup}$ & $\lambda$ & Loss & Link \\
\midrule
0.9 & 0.98 & 1e-20 & 0.008 & 1 & 0 & 128 & 2000 & 0.1 & 3.529 & \href{https://wandb.ai/marin-community/optimizer-scaling/runs/sweep-130m-2B-adamw5098e9lr0.008-wd0.1-minlr0-warmup2000-b10.9-b-bc5e36}{0} \\
\midrule
0.95 & -- & -- & -- & -- & -- & -- & -- & -- & 3.539 & \href{https://wandb.ai/marin-community/optimizer-scaling/runs/sweep-130m-2B-adamw6b5e10lr0.008-wd0.1-minlr0-warmup2000-b10.95--b1de93}{1} \\
0.98 & -- & -- & -- & -- & -- & -- & -- & -- & 3.882 & \href{https://wandb.ai/marin-community/optimizer-scaling/runs/sweep-130m-2B-adamwae7d4dlr0.008-wd0.1-minlr0-warmup2000-b10.98--df53b1}{2} \\
-- & 0.9 & -- & -- & -- & -- & -- & -- & -- & 3.545 & \href{https://wandb.ai/marin-community/optimizer-scaling/runs/sweep-130m-2B-adamw5a9f1blr0.008-wd0.1-minlr0-warmup2000-b10.9-b-691bb8}{3} \\
-- & 0.95 & -- & -- & -- & -- & -- & -- & -- & 3.535 & \href{https://wandb.ai/marin-community/optimizer-scaling/runs/sweep-130m-2B-adamwb5ba64lr0.008-wd0.1-minlr0-warmup2000-b10.9-b-77d771}{4} \\
-- & -- & 1e-25 & -- & -- & -- & -- & -- & -- & 3.529 & \href{https://wandb.ai/marin-community/optimizer-scaling/runs/sweep-130m-2B-adamw0848aelr0.008-wd0.1-minlr0-warmup2000-b10.9-b-413f11}{5} \\
-- & -- & 1e-15 & -- & -- & -- & -- & -- & -- & 3.531 & \href{https://wandb.ai/marin-community/optimizer-scaling/runs/sweep-130m-2B-adamwca195dlr0.008-wd0.1-minlr0-warmup2000-b10.9-b-1f0e34}{6} \\
\bottomrule
\end{tabular}
\end{table}

\begin{table}[H]
\centering
\caption{Hyperparameter ablation for AdamW on 130m on 2x Chinchilla Data}
\label{tab:ablation_adamw_130m_on_2x_chinchilla_data}
\begin{tabular}{ccccccccccc}
\toprule
$\beta_1$ & $\beta_2$ & $\epsilon$ & $\eta$ & $\gradnorm$ & $\eta_{min}$ & $\mathrm{BSZ}$ & $\mathrm{warmup}$ & $\lambda$ & Loss & Link \\
\midrule
0.9 & 0.98 & 1e-20 & 0.008 & 1 & 0 & 128 & 2000 & 0.1 & 3.409 & \href{https://wandb.ai/marin-community/optimizer-scaling/runs/sweep-130m-5B-adamw90f5c1lr0.008-wd0.1-minlr0-warmup2000-b10.9-b-67adb6}{0} \\
\midrule
0.95 & -- & -- & -- & -- & -- & -- & -- & -- & 3.417 & \href{https://wandb.ai/marin-community/optimizer-scaling/runs/sweep-130m-5B-adamw995582lr0.008-wd0.1-minlr0-warmup2000-b10.95--0fb321}{1} \\
0.98 & -- & -- & -- & -- & -- & -- & -- & -- & 7.557 & \href{https://wandb.ai/marin-community/optimizer-scaling/runs/sweep-130m-5B-adamw220e11lr0.008-wd0.1-minlr0-warmup2000-b10.98--f8f851}{2} \\
-- & 0.9 & -- & -- & -- & -- & -- & -- & -- & 3.423 & \href{https://wandb.ai/marin-community/optimizer-scaling/runs/sweep-130m-5B-adamw757bfblr0.008-wd0.1-minlr0-warmup2000-b10.9-b-da391a}{3} \\
-- & 0.95 & -- & -- & -- & -- & -- & -- & -- & 3.413 & \href{https://wandb.ai/marin-community/optimizer-scaling/runs/sweep-130m-5B-adamwc562dclr0.008-wd0.1-minlr0-warmup2000-b10.9-b-9752b5}{4} \\
-- & -- & 1e-25 & -- & -- & -- & -- & -- & -- & 3.409 & \href{https://wandb.ai/marin-community/optimizer-scaling/runs/sweep-130m-5B-adamw0da527lr0.008-wd0.1-minlr0-warmup2000-b10.9-b-9c6463}{5} \\
-- & -- & 1e-15 & -- & -- & -- & -- & -- & -- & 3.409 & \href{https://wandb.ai/marin-community/optimizer-scaling/runs/sweep-130m-5B-adamw4f03a8lr0.008-wd0.1-minlr0-warmup2000-b10.9-b-c453b4}{6} \\
\bottomrule
\end{tabular}
\end{table}

\begin{table}[H]
\centering
\caption{Hyperparameter ablation for AdamW on 130m on 4x Chinchilla Data}
\label{tab:ablation_adamw_130m_on_4x_chinchilla_data}
\begin{tabular}{ccccccccccc}
\toprule
$\beta_1$ & $\beta_2$ & $\epsilon$ & $\eta$ & $\gradnorm$ & $\eta_{min}$ & $\mathrm{BSZ}$ & $\mathrm{warmup}$ & $\lambda$ & Loss & Link \\
\midrule
0.9 & 0.98 & 1e-20 & 0.008 & 1 & 0 & 128 & 2000 & 0.1 & 3.322 & \href{https://wandb.ai/marin-community/optimizer-scaling/runs/sweep-130m-10B-adamw28a1ddlr0.008-wd0.1-minlr0-warmup2000-b10.9--a5a31c}{0} \\
\midrule
0.95 & -- & -- & -- & -- & -- & -- & -- & -- & 3.330 & \href{https://wandb.ai/marin-community/optimizer-scaling/runs/sweep-130m-10B-adamw3cefb4lr0.008-wd0.1-minlr0-warmup2000-b10.95-d2de07}{1} \\
0.98 & -- & -- & -- & -- & -- & -- & -- & -- & 3.416 & \href{https://wandb.ai/marin-community/optimizer-scaling/runs/sweep-130m-10B-adamwdec095lr0.008-wd0.1-minlr0-warmup2000-b10.98-5164f7}{2} \\
-- & 0.9 & -- & -- & -- & -- & -- & -- & -- & 3.338 & \href{https://wandb.ai/marin-community/optimizer-scaling/runs/sweep-130m-10B-adamw50c768lr0.008-wd0.1-minlr0-warmup2000-b10.9--374622}{3} \\
-- & 0.95 & -- & -- & -- & -- & -- & -- & -- & 3.329 & \href{https://wandb.ai/marin-community/optimizer-scaling/runs/sweep-130m-10B-adamw41da7flr0.008-wd0.1-minlr0-warmup2000-b10.9--e79638}{4} \\
-- & -- & 1e-25 & -- & -- & -- & -- & -- & -- & 3.322 & \href{https://wandb.ai/marin-community/optimizer-scaling/runs/sweep-130m-10B-adamw11bb27lr0.008-wd0.1-minlr0-warmup2000-b10.9--112190}{5} \\
-- & -- & 1e-15 & -- & -- & -- & -- & -- & -- & 3.323 & \href{https://wandb.ai/marin-community/optimizer-scaling/runs/sweep-130m-10B-adamw2151b5lr0.008-wd0.1-minlr0-warmup2000-b10.9--8b96e9}{6} \\
\bottomrule
\end{tabular}
\end{table}

\begin{table}[H]
\centering
\caption{Hyperparameter ablation for AdamW on 130m on 8x Chinchilla Data}
\label{tab:ablation_adamw_130m_on_8x_chinchilla_data}
\begin{tabular}{ccccccccccc}
\toprule
$\beta_1$ & $\beta_2$ & $\epsilon$ & $\eta$ & $\gradnorm$ & $\eta_{min}$ & $\mathrm{BSZ}$ & $\mathrm{warmup}$ & $\lambda$ & Loss & Link \\
\midrule
0.9 & 0.98 & 1e-20 & 0.008 & 1 & 0 & 256 & 1000 & 0.1 & 3.262 & \href{https://wandb.ai/marin-community/optimizer-scaling/runs/sweep-130m-21B-adamwc00703lr0.008-wd0.1-minlr0-warmup1000-b10.9--d73c2c}{0} \\
\midrule
0.95 & -- & -- & -- & -- & -- & -- & -- & -- & 3.273 & \href{https://wandb.ai/marin-community/optimizer-scaling/runs/sweep-130m-21B-adamwd5d639lr0.008-wd0.1-minlr0-warmup1000-b10.95-658eb8}{1} \\
0.98 & -- & -- & -- & -- & -- & -- & -- & -- & 3.430 & \href{https://wandb.ai/marin-community/optimizer-scaling/runs/sweep-130m-21B-adamw0cb2d6lr0.008-wd0.1-minlr0-warmup1000-b10.98-7c952d}{2} \\
-- & 0.9 & -- & -- & -- & -- & -- & -- & -- & 3.272 & \href{https://wandb.ai/marin-community/optimizer-scaling/runs/sweep-130m-21B-adamwe479a6lr0.008-wd0.1-minlr0-warmup1000-b10.9--7570fe}{3} \\
-- & 0.95 & -- & -- & -- & -- & -- & -- & -- & 3.266 & \href{https://wandb.ai/marin-community/optimizer-scaling/runs/sweep-130m-21B-adamw63b609lr0.008-wd0.1-minlr0-warmup1000-b10.9--2da2b5}{4} \\
-- & -- & 1e-25 & -- & -- & -- & -- & -- & -- & 3.262 & \href{https://wandb.ai/marin-community/optimizer-scaling/runs/sweep-130m-21B-adamwc0b69dlr0.008-wd0.1-minlr0-warmup1000-b10.9--2390cc}{5} \\
-- & -- & 1e-15 & -- & -- & -- & -- & -- & -- & 3.263 & \href{https://wandb.ai/marin-community/optimizer-scaling/runs/sweep-130m-21B-adamw9248c1lr0.008-wd0.1-minlr0-warmup1000-b10.9--7ae1a8}{6} \\
\bottomrule
\end{tabular}
\end{table}

\begin{table}[H]
\centering
\caption{Hyperparameter ablation for AdamW on 300m on 1x Chinchilla Data}
\label{tab:ablation_adamw_300m_on_1x_chinchilla_data}
\begin{tabular}{ccccccccccc}
\toprule
$\beta_1$ & $\beta_2$ & $\epsilon$ & $\eta$ & $\gradnorm$ & $\eta_{min}$ & $\mathrm{BSZ}$ & $\mathrm{warmup}$ & $\lambda$ & Loss & Link \\
\midrule
0.9 & 0.98 & 1e-10 & 0.008 & 1 & 0 & 128 & 2000 & 0.1 & 3.264 & \href{https://wandb.ai/marin-community/optimizer-scaling/runs/sweep-300m-6B-adamwa7aafelr0.008-wd0.1-minlr0-warmup2000-b10.9-b-990eda}{0} \\
\midrule
0.95 & -- & -- & -- & -- & -- & -- & -- & -- & 3.271 & \href{https://wandb.ai/marin-community/optimizer-scaling/runs/sweep-300m-6B-adamw55fc1flr0.008-wd0.1-minlr0-warmup2000-b10.95--b7196b}{1} \\
0.98 & -- & -- & -- & -- & -- & -- & -- & -- & 7.351 & \href{https://wandb.ai/marin-community/optimizer-scaling/runs/sweep-300m-6B-adamw3aa56elr0.008-wd0.1-minlr0-warmup2000-b10.98--36a5e7}{2} \\
-- & 0.9 & -- & -- & -- & -- & -- & -- & -- & 3.280 & \href{https://wandb.ai/marin-community/optimizer-scaling/runs/sweep-300m-6B-adamw72a6d2lr0.008-wd0.1-minlr0-warmup2000-b10.9-b-dde1e5}{3} \\
-- & 0.95 & -- & -- & -- & -- & -- & -- & -- & 3.269 & \href{https://wandb.ai/marin-community/optimizer-scaling/runs/sweep-300m-6B-adamw901abclr0.008-wd0.1-minlr0-warmup2000-b10.9-b-de77a3}{4} \\
-- & -- & 1e-25 & -- & -- & -- & -- & -- & -- & 3.265 & \href{https://wandb.ai/marin-community/optimizer-scaling/runs/sweep-300m-6B-adamwfa0d44lr0.008-wd0.1-minlr0-warmup2000-b10.9-b-89eba9}{5} \\
-- & -- & 1e-20 & -- & -- & -- & -- & -- & -- & 3.265 & \href{https://wandb.ai/marin-community/optimizer-scaling/runs/sweep-300m-6B-adamw057b30lr0.008-wd0.1-minlr0-warmup2000-b10.9-b-cff503}{6} \\
\bottomrule
\end{tabular}
\end{table}

\begin{table}[H]
\centering
\caption{Hyperparameter ablation for AdamW on 520m on 1x Chinchilla Data}
\label{tab:ablation_adamw_520m_on_1x_chinchilla_data}
\begin{tabular}{ccccccccccc}
\toprule
$\beta_1$ & $\beta_2$ & $\epsilon$ & $\eta$ & $\gradnorm$ & $\eta_{min}$ & $\mathrm{BSZ}$ & $\mathrm{warmup}$ & $\lambda$ & Loss & Link \\
\midrule
0.9 & 0.98 & 1e-10 & 0.004 & 1 & 0 & 256 & 1000 & 0.2 & 3.110 & \href{https://wandb.ai/marin-community/optimizer-scaling/runs/sweep-520m-10B-adamwf7a9c6lr0.004-wd0.2-minlr0-warmup1000-b10.9--0a09ef}{0} \\
\midrule
0.95 & -- & -- & -- & -- & -- & -- & -- & -- & 3.112 & \href{https://wandb.ai/marin-community/optimizer-scaling/runs/sweep-520m-10B-adamw92fab3lr0.004-wd0.2-minlr0-warmup1000-b10.95-b52d40}{1} \\
0.98 & -- & -- & -- & -- & -- & -- & -- & -- & 3.229 & \href{https://wandb.ai/marin-community/optimizer-scaling/runs/sweep-520m-10B-adamw757dfblr0.004-wd0.2-minlr0-warmup1000-b10.98-82eb00}{2} \\
-- & 0.9 & -- & -- & -- & -- & -- & -- & -- & 3.116 & \href{https://wandb.ai/marin-community/optimizer-scaling/runs/sweep-520m-10B-adamw2399bdlr0.004-wd0.2-minlr0-warmup1000-b10.9--235c9f}{3} \\
-- & 0.95 & -- & -- & -- & -- & -- & -- & -- & 3.111 & \href{https://wandb.ai/marin-community/optimizer-scaling/runs/sweep-520m-10B-adamw49b619lr0.004-wd0.2-minlr0-warmup1000-b10.9--0339de}{4} \\
-- & -- & 1e-25 & -- & -- & -- & -- & -- & -- & 3.116 & \href{https://wandb.ai/marin-community/optimizer-scaling/runs/sweep-520m-10B-adamwe08492lr0.004-wd0.2-minlr0-warmup1000-b10.9--550b0a}{5} \\
-- & -- & 1e-20 & -- & -- & -- & -- & -- & -- & 3.116 & \href{https://wandb.ai/marin-community/optimizer-scaling/runs/sweep-520m-10B-adamw920b54lr0.004-wd0.2-minlr0-warmup1000-b10.9--29bc56}{6} \\
\bottomrule
\end{tabular}
\end{table}

\subsection{Sweeping Results for Cautious}%
\begin{table}[H]
\centering
\caption{Hyperparameter ablation for Cautious on 130m on 1x Chinchilla Data}
\label{tab:ablation_cautious_130m_on_1x_chinchilla_data}
\begin{tabular}{ccccccccccc}
\toprule
$\beta_1$ & $\beta_2$ & $\epsilon$ & $\eta$ & $\gradnorm$ & $\eta_{min}$ & $\mathrm{BSZ}$ & $\mathrm{warmup}$ & $\lambda$ & Loss & Link \\
\midrule
0.95 & 0.98 & 1e-15 & 0.008 & 1 & 0 & 128 & 2000 & 0.1 & 3.535 & \href{https://wandb.ai/marin-community/optimizer-scaling/runs/sweep-130m-2B-cautious7b0572lr0.008-wd0.1-minlr0-warmup2000-b10.-085cb0}{0} \\
\midrule
0.8 & -- & -- & -- & -- & -- & -- & -- & -- & 6.698 & \href{https://wandb.ai/marin-community/optimizer-scaling/runs/sweep-130m-2B-cautious90e2dblr0.008-wd0.1-minlr0-warmup2000-b10.-3a5e65}{1} \\
0.9 & -- & -- & -- & -- & -- & -- & -- & -- & 3.549 & \href{https://wandb.ai/marin-community/optimizer-scaling/runs/sweep-130m-2B-cautious5fbf2flr0.008-wd0.1-minlr0-warmup2000-b10.-8d12f3}{2} \\
0.98 & -- & -- & -- & -- & -- & -- & -- & -- & 3.534 & \href{https://wandb.ai/marin-community/optimizer-scaling/runs/sweep-130m-2B-cautiousd41969lr0.008-wd0.1-minlr0-warmup2000-b10.-7d014c}{3} \\
-- & 0.9 & -- & -- & -- & -- & -- & -- & -- & 3.551 & \href{https://wandb.ai/marin-community/optimizer-scaling/runs/sweep-130m-2B-cautious87fa48lr0.008-wd0.1-minlr0-warmup2000-b10.-dff836}{4} \\
-- & 0.95 & -- & -- & -- & -- & -- & -- & -- & 3.543 & \href{https://wandb.ai/marin-community/optimizer-scaling/runs/sweep-130m-2B-cautious04bad0lr0.008-wd0.1-minlr0-warmup2000-b10.-15c6af}{5} \\
-- & -- & 1e-25 & -- & -- & -- & -- & -- & -- & 3.537 & \href{https://wandb.ai/marin-community/optimizer-scaling/runs/sweep-130m-2B-cautious2a7d1dlr0.008-wd0.1-minlr0-warmup2000-b10.-ff7d5f}{6} \\
\bottomrule
\end{tabular}
\end{table}

\begin{table}[H]
\centering
\caption{Hyperparameter ablation for Cautious on 130m on 2x Chinchilla Data}
\label{tab:ablation_cautious_130m_on_2x_chinchilla_data}
\begin{tabular}{ccccccccccc}
\toprule
$\beta_1$ & $\beta_2$ & $\epsilon$ & $\eta$ & $\gradnorm$ & $\eta_{min}$ & $\mathrm{BSZ}$ & $\mathrm{warmup}$ & $\lambda$ & Loss & Link \\
\midrule
0.98 & 0.98 & 1e-15 & 0.008 & 2 & 0 & 128 & 2000 & 0.1 & 3.403 & \href{https://wandb.ai/marin-community/optimizer-scaling/runs/sweep-130m-5B-cautiousce8123lr0.008-wd0.1-minlr0-warmup2000-b10.-fd1867}{0} \\
\midrule
0.8 & -- & -- & -- & -- & -- & -- & -- & -- & 7.417 & \href{https://wandb.ai/marin-community/optimizer-scaling/runs/sweep-130m-5B-cautious4f3b8flr0.008-wd0.1-minlr0-warmup2000-b10.-ef7ead}{1} \\
0.9 & -- & -- & -- & -- & -- & -- & -- & -- & 3.427 & \href{https://wandb.ai/marin-community/optimizer-scaling/runs/sweep-130m-5B-cautious585d63lr0.008-wd0.1-minlr0-warmup2000-b10.-6158ce}{2} \\
0.95 & -- & -- & -- & -- & -- & -- & -- & -- & 3.413 & \href{https://wandb.ai/marin-community/optimizer-scaling/runs/sweep-130m-5B-cautious5d7675lr0.008-wd0.1-minlr0-warmup2000-b10.-76a83d}{3} \\
-- & 0.9 & -- & -- & -- & -- & -- & -- & -- & >10 & \href{https://wandb.ai/marin-community/optimizer-scaling/runs/sweep-130m-5B-cautious6aa3ablr0.008-wd0.1-minlr0-warmup2000-b10.-752115}{4} \\
-- & 0.95 & -- & -- & -- & -- & -- & -- & -- & 3.408 & \href{https://wandb.ai/marin-community/optimizer-scaling/runs/sweep-130m-5B-cautious35372elr0.008-wd0.1-minlr0-warmup2000-b10.-681c2f}{5} \\
-- & -- & 1e-25 & -- & -- & -- & -- & -- & -- & 3.404 & \href{https://wandb.ai/marin-community/optimizer-scaling/runs/sweep-130m-5B-cautiousedcc52lr0.008-wd0.1-minlr0-warmup2000-b10.-57f831}{6} \\
\bottomrule
\end{tabular}
\end{table}

\begin{table}[H]
\centering
\caption{Hyperparameter ablation for Cautious on 130m on 8x Chinchilla Data}
\label{tab:ablation_cautious_130m_on_8x_chinchilla_data}
\begin{tabular}{ccccccccccc}
\toprule
$\beta_1$ & $\beta_2$ & $\epsilon$ & $\eta$ & $\gradnorm$ & $\eta_{min}$ & $\mathrm{BSZ}$ & $\mathrm{warmup}$ & $\lambda$ & Loss & Link \\
\midrule
0.98 & 0.98 & 1e-15 & 0.008 & 2 & 0 & 256 & 2000 & 0.1 & 3.253 & \href{https://wandb.ai/marin-community/optimizer-scaling/runs/sweep-130m-21B-cautious3e1de9lr0.008-wd0.1-minlr0-warmup2000-b10-8f1bb6}{0} \\
\midrule
0.8 & -- & -- & -- & -- & -- & -- & -- & -- & 7.395 & \href{https://wandb.ai/marin-community/optimizer-scaling/runs/sweep-130m-21B-cautious903025lr0.008-wd0.1-minlr0-warmup2000-b10-c9b0ed}{1} \\
0.9 & -- & -- & -- & -- & -- & -- & -- & -- & 3.271 & \href{https://wandb.ai/marin-community/optimizer-scaling/runs/sweep-130m-21B-cautiousdb200alr0.008-wd0.1-minlr0-warmup2000-b10-d34aeb}{2} \\
0.95 & -- & -- & -- & -- & -- & -- & -- & -- & 3.259 & \href{https://wandb.ai/marin-community/optimizer-scaling/runs/sweep-130m-21B-cautious1b41e3lr0.008-wd0.1-minlr0-warmup2000-b10-e60376}{3} \\
-- & 0.9 & -- & -- & -- & -- & -- & -- & -- & >10 & \href{https://wandb.ai/marin-community/optimizer-scaling/runs/sweep-130m-21B-cautious918292lr0.008-wd0.1-minlr0-warmup2000-b10-43c09c}{4} \\
-- & 0.95 & -- & -- & -- & -- & -- & -- & -- & 3.253 & \href{https://wandb.ai/marin-community/optimizer-scaling/runs/sweep-130m-21B-cautious8370eflr0.008-wd0.1-minlr0-warmup2000-b10-beca91}{5} \\
-- & -- & 1e-25 & -- & -- & -- & -- & -- & -- & 3.251 & \href{https://wandb.ai/marin-community/optimizer-scaling/runs/sweep-130m-21B-cautious40f0d8lr0.008-wd0.1-minlr0-warmup2000-b10-5be279}{6} \\
\bottomrule
\end{tabular}
\end{table}

\begin{table}[H]
\centering
\caption{Hyperparameter ablation for Cautious on 300m on 1x Chinchilla Data}
\label{tab:ablation_cautious_300m_on_1x_chinchilla_data}
\begin{tabular}{ccccccccccc}
\toprule
$\beta_1$ & $\beta_2$ & $\epsilon$ & $\eta$ & $\gradnorm$ & $\eta_{min}$ & $\mathrm{BSZ}$ & $\mathrm{warmup}$ & $\lambda$ & Loss & Link \\
\midrule
0.98 & 0.98 & 1e-25 & 0.008 & 2 & 0 & 128 & 2000 & 0.1 & 3.260 & \href{https://wandb.ai/marin-community/optimizer-scaling/runs/sweep-300m-6B-cautiousc8346clr0.008-wd0.1-minlr0-warmup2000-b10.-852297}{0} \\
\midrule
0.9 & -- & -- & -- & -- & -- & -- & -- & -- & 3.286 & \href{https://wandb.ai/marin-community/optimizer-scaling/runs/sweep-300m-6B-cautiousfed852lr0.008-wd0.1-minlr0-warmup2000-b10.-4b06ef}{1} \\
0.95 & -- & -- & -- & -- & -- & -- & -- & -- & 3.271 & \href{https://wandb.ai/marin-community/optimizer-scaling/runs/sweep-300m-6B-cautious7244dblr0.008-wd0.1-minlr0-warmup2000-b10.-5fd975}{2} \\
-- & 0.9 & -- & -- & -- & -- & -- & -- & -- & >10 & \href{https://wandb.ai/marin-community/optimizer-scaling/runs/sweep-300m-6B-cautiousd8e2a3lr0.008-wd0.1-minlr0-warmup2000-b10.-3812f2}{3} \\
-- & 0.95 & -- & -- & -- & -- & -- & -- & -- & >10 & \href{https://wandb.ai/marin-community/optimizer-scaling/runs/sweep-300m-6B-cautious69dafalr0.008-wd0.1-minlr0-warmup2000-b10.-5e0115}{4} \\
-- & -- & 1e-25 & -- & -- & -- & -- & -- & -- & 3.260 & \href{https://wandb.ai/marin-community/optimizer-scaling/runs/sweep-300m-6B-cautiousc8346clr0.008-wd0.1-minlr0-warmup2000-b10.-852297}{5} \\
-- & -- & 1e-15 & -- & -- & -- & -- & -- & -- & 3.260 & \href{https://wandb.ai/marin-community/optimizer-scaling/runs/sweep-300m-6B-cautious56bf47lr0.008-wd0.1-minlr0-warmup2000-b10.-19a98d}{6} \\
\bottomrule
\end{tabular}
\end{table}

\begin{table}[H]
\centering
\caption{Hyperparameter ablation for Cautious on 520m on 1x Chinchilla Data}
\label{tab:ablation_cautious_520m_on_1x_chinchilla_data}
\begin{tabular}{ccccccccccc}
\toprule
$\beta_1$ & $\beta_2$ & $\epsilon$ & $\eta$ & $\gradnorm$ & $\eta_{min}$ & $\mathrm{BSZ}$ & $\mathrm{warmup}$ & $\lambda$ & Loss & Link \\
\midrule
0.98 & 0.98 & 1e-25 & 0.008 & 1 & 0 & 256 & 2000 & 0.1 & 3.100 & \href{https://wandb.ai/marin-community/optimizer-scaling/runs/sweep-520m-10B-cautiouscda486lr0.008-wd0.1-minlr0-warmup2000-b10-1d8924}{0} \\
\midrule
0.8 & -- & -- & -- & -- & -- & -- & -- & -- & >10 & \href{https://wandb.ai/marin-community/optimizer-scaling/runs/sweep-520m-10B-cautioush937574lr0.008-wd0.1-minlr0-warmup2000-b1-288d18}{1} \\
0.9 & -- & -- & -- & -- & -- & -- & -- & -- & 7.264 & \href{https://wandb.ai/marin-community/optimizer-scaling/runs/sweep-520m-10B-cautiousfe5526lr0.008-wd0.1-minlr0-warmup2000-b10-a0ea38}{2} \\
0.95 & -- & -- & -- & -- & -- & -- & -- & -- & 3.108 & \href{https://wandb.ai/marin-community/optimizer-scaling/runs/sweep-520m-10B-cautiousb9bea7lr0.008-wd0.1-minlr0-warmup2000-b10-4e055f}{3} \\
-- & 0.9 & -- & -- & -- & -- & -- & -- & -- & >10 & \href{https://wandb.ai/marin-community/optimizer-scaling/runs/sweep-520m-10B-cautiousfa5ab7lr0.008-wd0.1-minlr0-warmup2000-b10-fc260f}{4} \\
-- & 0.95 & -- & -- & -- & -- & -- & -- & -- & 3.105 & \href{https://wandb.ai/marin-community/optimizer-scaling/runs/sweep-520m-10B-cautious06f834lr0.008-wd0.1-minlr0-warmup2000-b10-e18beb}{5} \\
-- & -- & 1e-25 & -- & -- & -- & -- & -- & -- & 3.100 & \href{https://wandb.ai/marin-community/optimizer-scaling/runs/sweep-520m-10B-cautiouscda486lr0.008-wd0.1-minlr0-warmup2000-b10-1d8924}{6} \\
\bottomrule
\end{tabular}
\end{table}

\subsection{Sweeping Results for Lion}%
\begin{table}[H]
\centering
\caption{Hyperparameter ablation for Lion on 130m on 1x Chinchilla Data}
\label{tab:ablation_lion_130m_on_1x_chinchilla_data}
\begin{tabular}{cccccccccc}
\toprule
$\beta_1$ & $\beta_2$ & $\eta$ & $\gradnorm$ & $\eta_{min}$ & $\mathrm{BSZ}$ & $\mathrm{warmup}$ & $\lambda$ & Loss & Link \\
\midrule
0.9 & 0.95 & 0.002 & 1 & 0 & 128 & 2000 & 0.7 & 3.552 & \href{https://wandb.ai/marin-community/optimizer-scaling/runs/sweep-130m-2B-lion3532d0lr0.002-wd0.7-minlr0-warmup2000-b10.9-b2-2f88e4}{0} \\
\midrule
0.8 & -- & -- & -- & -- & -- & -- & -- & 3.575 & \href{https://wandb.ai/marin-community/optimizer-scaling/runs/sweep-130m-2B-lion8b138flr0.002-wd0.7-minlr0-warmup2000-b10.8-b2-9d739e}{1} \\
0.95 & -- & -- & -- & -- & -- & -- & -- & 3.557 & \href{https://wandb.ai/marin-community/optimizer-scaling/runs/sweep-130m-2B-lion49977alr0.002-wd0.7-minlr0-warmup2000-b10.95-b-0c8700}{2} \\
0.98 & -- & -- & -- & -- & -- & -- & -- & 7.644 & \href{https://wandb.ai/marin-community/optimizer-scaling/runs/sweep-130m-2B-lion584608lr0.002-wd0.7-minlr0-warmup2000-b10.98-b-bb0378}{3} \\
-- & 0.9 & -- & -- & -- & -- & -- & -- & 3.550 & \href{https://wandb.ai/marin-community/optimizer-scaling/runs/sweep-130m-2B-lionf88c83lr0.002-wd0.7-minlr0-warmup2000-b10.9-b2-3f2e17}{4} \\
-- & 0.98 & -- & -- & -- & -- & -- & -- & 3.630 & \href{https://wandb.ai/marin-community/optimizer-scaling/runs/sweep-130m-2B-lionfc7ef4lr0.002-wd0.7-minlr0-warmup2000-b10.9-b2-27db87}{5} \\
-- & -- & 0.0005 & -- & -- & -- & -- & -- & 3.579 & \href{https://wandb.ai/marin-community/optimizer-scaling/runs/sweep-130m-2B-liony630ce4lr0.0005-wd0.7-minlr0-warmup2000-b10.9--ceaedd}{6} \\
\bottomrule
\end{tabular}
\end{table}

\begin{table}[H]
\centering
\caption{Hyperparameter ablation for Lion on 130m on 2x Chinchilla Data}
\label{tab:ablation_lion_130m_on_2x_chinchilla_data}
\begin{tabular}{cccccccccc}
\toprule
$\beta_1$ & $\beta_2$ & $\eta$ & $\gradnorm$ & $\eta_{min}$ & $\mathrm{BSZ}$ & $\mathrm{warmup}$ & $\lambda$ & Loss & Link \\
\midrule
0.9 & 0.95 & 0.001 & 1 & 0 & 128 & 2000 & 0.7 & 3.414 & \href{https://wandb.ai/marin-community/optimizer-scaling/runs/sweep-130m-5B-lione0352dlr0.001-wd0.7-minlr0-warmup2000-b10.9-b2-89e6a1}{0} \\
\midrule
0.8 & -- & -- & -- & -- & -- & -- & -- & 3.436 & \href{https://wandb.ai/marin-community/optimizer-scaling/runs/sweep-130m-5B-lion50b543lr0.001-wd0.7-minlr0-warmup2000-b10.8-b2-b6925d}{1} \\
0.95 & -- & -- & -- & -- & -- & -- & -- & 3.418 & \href{https://wandb.ai/marin-community/optimizer-scaling/runs/sweep-130m-5B-liondeda64lr0.001-wd0.7-minlr0-warmup2000-b10.95-b-005903}{2} \\
0.98 & -- & -- & -- & -- & -- & -- & -- & 7.313 & \href{https://wandb.ai/marin-community/optimizer-scaling/runs/sweep-130m-5B-lion2ccccblr0.001-wd0.7-minlr0-warmup2000-b10.98-b-ce81dd}{3} \\
-- & 0.9 & -- & -- & -- & -- & -- & -- & 3.433 & \href{https://wandb.ai/marin-community/optimizer-scaling/runs/sweep-130m-5B-lion9321cblr0.001-wd0.7-minlr0-warmup2000-b10.9-b2-f29f78}{4} \\
-- & 0.98 & -- & -- & -- & -- & -- & -- & 3.409 & \href{https://wandb.ai/marin-community/optimizer-scaling/runs/sweep-130m-5B-lion5d07e9lr0.001-wd0.7-minlr0-warmup2000-b10.9-b2-6d65d0}{5} \\
-- & -- & 0.0005 & -- & -- & -- & -- & -- & 3.435 & \href{https://wandb.ai/marin-community/optimizer-scaling/runs/sweep-130m-5B-lionb0bcd6lr0.0005-wd0.7-minlr0-warmup2000-b10.9-b-409f2d}{6} \\
\bottomrule
\end{tabular}
\end{table}

\begin{table}[H]
\centering
\caption{Hyperparameter ablation for Lion on 130m on 4x Chinchilla Data}
\label{tab:ablation_lion_130m_on_4x_chinchilla_data}
\begin{tabular}{cccccccccc}
\toprule
$\beta_1$ & $\beta_2$ & $\eta$ & $\gradnorm$ & $\eta_{min}$ & $\mathrm{BSZ}$ & $\mathrm{warmup}$ & $\lambda$ & Loss & Link \\
\midrule
0.9 & 0.95 & 0.001 & 1 & 0 & 128 & 2000 & 0.7 & 3.331 & \href{https://wandb.ai/marin-community/optimizer-scaling/runs/sweep-130m-10B-liondd8061alr0.001-wd0.7-minlr0-warmup2000-b10.9--72b789}{0} \\
\midrule
0.8 & -- & -- & -- & -- & -- & -- & -- & 3.345 & \href{https://wandb.ai/marin-community/optimizer-scaling/runs/sweep-130m-10B-lion6ebc4dlr0.001-wd0.7-minlr0-warmup2000-b10.8-b-9ccc4b}{1} \\
0.95 & -- & -- & -- & -- & -- & -- & -- & 3.335 & \href{https://wandb.ai/marin-community/optimizer-scaling/runs/sweep-130m-10B-lionk6ee825lr0.001-wd0.7-minlr0-warmup2000-b10.95-4aeb28}{2} \\
0.98 & -- & -- & -- & -- & -- & -- & -- & 7.330 & \href{https://wandb.ai/marin-community/optimizer-scaling/runs/sweep-130m-10B-lionk98635alr0.001-wd0.7-minlr0-warmup2000-b10.98-20924d}{3} \\
-- & 0.9 & -- & -- & -- & -- & -- & -- & 3.345 & \href{https://wandb.ai/marin-community/optimizer-scaling/runs/sweep-130m-10B-lioncda5edlr0.001-wd0.7-minlr0-warmup2000-b10.9-b-6a3aec}{4} \\
-- & 0.98 & -- & -- & -- & -- & -- & -- & 3.340 & \href{https://wandb.ai/marin-community/optimizer-scaling/runs/sweep-130m-10B-lion489f64lr0.001-wd0.7-minlr0-warmup2000-b10.9-b-e5b223}{5} \\
-- & -- & 0.0005 & -- & -- & -- & -- & -- & 3.338 & \href{https://wandb.ai/marin-community/optimizer-scaling/runs/sweep-130m-10B-lionfc33e8lr0.0005-wd0.7-minlr0-warmup2000-b10.9--abc4e2}{6} \\
\bottomrule
\end{tabular}
\end{table}

\begin{table}[H]
\centering
\caption{Hyperparameter ablation for Lion on 130m on 8x Chinchilla Data}
\label{tab:ablation_lion_130m_on_8x_chinchilla_data}
\begin{tabular}{cccccccccc}
\toprule
$\beta_1$ & $\beta_2$ & $\eta$ & $\gradnorm$ & $\eta_{min}$ & $\mathrm{BSZ}$ & $\mathrm{warmup}$ & $\lambda$ & Loss & Link \\
\midrule
0.9 & 0.98 & 0.001 & 1 & 0 & 128 & 2000 & 0.7 & 3.252 & \href{https://wandb.ai/marin-community/optimizer-scaling/runs/sweep-130m-21B-lion85b813lr0.001-wd0.7-minlr0-warmup2000-b10.9-b-eef545}{0} \\
\midrule
0.8 & -- & -- & -- & -- & -- & -- & -- & 3.287 & \href{https://wandb.ai/marin-community/optimizer-scaling/runs/sweep-130m-21B-lionje7bccblr0.001-wd0.7-minlr0-warmup2000-b10.8--46199a}{1} \\
0.95 & -- & -- & -- & -- & -- & -- & -- & 3.264 & \href{https://wandb.ai/marin-community/optimizer-scaling/runs/sweep-130m-21B-lionjc17beflr0.001-wd0.7-minlr0-warmup2000-b10.95-bd5017}{2} \\
0.98 & -- & -- & -- & -- & -- & -- & -- & 3.286 & \href{https://wandb.ai/marin-community/optimizer-scaling/runs/sweep-130m-21B-liona1354cclr0.001-wd0.7-minlr0-warmup2000-b10.98-b74486}{3} \\
-- & 0.9 & -- & -- & -- & -- & -- & -- & 3.277 & \href{https://wandb.ai/marin-community/optimizer-scaling/runs/sweep-130m-21B-liona9ff1b8lr0.001-wd0.7-minlr0-warmup2000-b10.9--9c346c}{4} \\
-- & 0.95 & -- & -- & -- & -- & -- & -- & 3.263 & \href{https://wandb.ai/marin-community/optimizer-scaling/runs/sweep-130m-21B-lionje4e710lr0.001-wd0.7-minlr0-warmup2000-b10.9--f5263e}{5} \\
-- & -- & 0.0005 & -- & -- & -- & -- & -- & 3.254 & \href{https://wandb.ai/marin-community/optimizer-scaling/runs/sweep-130m-21B-lionja6ac62lr0.0005-wd0.7-minlr0-warmup2000-b10.9-39f7c0}{6} \\
\bottomrule
\end{tabular}
\end{table}

\begin{table}[H]
\centering
\caption{Hyperparameter ablation for Lion on 300m on 1x Chinchilla Data}
\label{tab:ablation_lion_300m_on_1x_chinchilla_data}
\begin{tabular}{cccccccccc}
\toprule
$\beta_1$ & $\beta_2$ & $\eta$ & $\gradnorm$ & $\eta_{min}$ & $\mathrm{BSZ}$ & $\mathrm{warmup}$ & $\lambda$ & Loss & Link \\
\midrule
0.9 & 0.95 & 0.001 & 1 & 0 & 128 & 2000 & 0.6 & 3.268 & \href{https://wandb.ai/marin-community/optimizer-scaling/runs/sweep-300m-6B-lion00979clr0.001-wd0.6-minlr0-warmup2000-b10.9-b2-81fdf3}{0} \\
\midrule
0.8 & -- & -- & -- & -- & -- & -- & -- & 3.283 & \href{https://wandb.ai/marin-community/optimizer-scaling/runs/sweep-300m-6B-lionb6c8celr0.001-wd0.6-minlr0-warmup2000-b10.8-b2-b954bf}{1} \\
0.95 & -- & -- & -- & -- & -- & -- & -- & 3.271 & \href{https://wandb.ai/marin-community/optimizer-scaling/runs/sweep-300m-6B-lion11c981lr0.001-wd0.6-minlr0-warmup2000-b10.95-b-4fef8e}{2} \\
0.98 & -- & -- & -- & -- & -- & -- & -- & 7.690 & \href{https://wandb.ai/marin-community/optimizer-scaling/runs/sweep-300m-6B-lion5c8de5lr0.001-wd0.6-minlr0-warmup2000-b10.98-b-1d6304}{3} \\
-- & 0.9 & -- & -- & -- & -- & -- & -- & 3.283 & \href{https://wandb.ai/marin-community/optimizer-scaling/runs/sweep-300m-6B-lionf81910lr0.001-wd0.6-minlr0-warmup2000-b10.9-b2-3a8ce5}{4} \\
-- & 0.98 & -- & -- & -- & -- & -- & -- & 3.271 & \href{https://wandb.ai/marin-community/optimizer-scaling/runs/sweep-300m-6B-lion062393lr0.001-wd0.6-minlr0-warmup2000-b10.9-b2-aaa1ff}{5} \\
-- & -- & 0.0005 & -- & -- & -- & -- & -- & 3.286 & \href{https://wandb.ai/marin-community/optimizer-scaling/runs/sweep-300m-6B-lionfb90dflr0.0005-wd0.6-minlr0-warmup2000-b10.9-b-8dee1b}{6} \\
\bottomrule
\end{tabular}
\end{table}

\begin{table}[H]
\centering
\caption{Hyperparameter ablation for Lion on 520m on 1x Chinchilla Data}
\label{tab:ablation_lion_520m_on_1x_chinchilla_data}
\begin{tabular}{cccccccccc}
\toprule
$\beta_1$ & $\beta_2$ & $\eta$ & $\gradnorm$ & $\eta_{min}$ & $\mathrm{BSZ}$ & $\mathrm{warmup}$ & $\lambda$ & Loss & Link \\
\midrule
0.9 & 0.95 & 0.001 & 1 & 0 & 128 & 2000 & 0.7 & 3.108 & \href{https://wandb.ai/marin-community/optimizer-scaling/runs/sweep-520m-10B-liond8061alr0.001-wd0.7-minlr0-warmup2000-b10.9-b-942db7}{0} \\
\midrule
0.8 & -- & -- & -- & -- & -- & -- & -- & 3.117 & \href{https://wandb.ai/marin-community/optimizer-scaling/runs/sweep-520m-10B-lion6ebc4dlr0.001-wd0.7-minlr0-warmup2000-b10.8-b-edd66c}{1} \\
0.95 & -- & -- & -- & -- & -- & -- & -- & 3.121 & \href{https://wandb.ai/marin-community/optimizer-scaling/runs/sweep-520m-10B-lion6ee825lr0.001-wd0.7-minlr0-warmup2000-b10.95--60d07d}{2} \\
0.98 & -- & -- & -- & -- & -- & -- & -- & 7.577 & \href{https://wandb.ai/marin-community/optimizer-scaling/runs/sweep-520m-10B-lion98635alr0.001-wd0.7-minlr0-warmup2000-b10.98--391c82}{3} \\
-- & 0.9 & -- & -- & -- & -- & -- & -- & 3.119 & \href{https://wandb.ai/marin-community/optimizer-scaling/runs/sweep-520m-10B-lioncda5edlr0.001-wd0.7-minlr0-warmup2000-b10.9-b-2427cd}{4} \\
-- & 0.98 & -- & -- & -- & -- & -- & -- & 3.208 & \href{https://wandb.ai/marin-community/optimizer-scaling/runs/sweep-520m-10B-lion489f64lr0.001-wd0.7-minlr0-warmup2000-b10.9-b-5ffcf2}{5} \\
-- & -- & 0.0005 & -- & -- & -- & -- & -- & 3.113 & \href{https://wandb.ai/marin-community/optimizer-scaling/runs/sweep-520m-10B-lionfc33e8lr0.0005-wd0.7-minlr0-warmup2000-b10.9--6dcec6}{6} \\
\bottomrule
\end{tabular}
\end{table}

\subsection{Sweeping Results for Mars}%
\begin{table}[H]
\centering
\caption{Hyperparameter ablation for Mars on 130m on 1x Chinchilla Data}
\label{tab:ablation_mars_130m_on_1x_chinchilla_data}
\begin{tabular}{cccccccccccc}
\toprule
$\beta_1$ & $\beta_2$ & $\epsilon$ & $\gamma$ & $\eta$ & $\gradnorm$ & $\eta_{min}$ & $\mathrm{BSZ}$ & $\mathrm{warmup}$ & $\lambda$ & Loss & Link \\
\midrule
0.9 & 0.95 & 1e-25 & 0.025 & 0.016 & 1 & 0 & 128 & 2000 & 0.1 & 3.537 & \href{https://wandb.ai/marin-community/optimizer-scaling/runs/sweep-130m-2B-mars692316lr0.016-wd0.1-minlr0-warmup2000-b10.9-b2-93e276}{0} \\
\midrule
0.8 & -- & -- & -- & -- & -- & -- & -- & -- & -- & 3.568 & \href{https://wandb.ai/marin-community/optimizer-scaling/runs/sweep-130m-2B-mars429f19lr0.016-wd0.1-minlr0-warmup2000-b10.8-b2-8e6b1f}{1} \\
0.95 & -- & -- & -- & -- & -- & -- & -- & -- & -- & 3.548 & \href{https://wandb.ai/marin-community/optimizer-scaling/runs/sweep-130m-2B-mars07f867lr0.016-wd0.1-minlr0-warmup2000-b10.95-b-1c9da8}{2} \\
0.98 & -- & -- & -- & -- & -- & -- & -- & -- & -- & 3.586 & \href{https://wandb.ai/marin-community/optimizer-scaling/runs/sweep-130m-2B-mars32fa0blr0.016-wd0.1-minlr0-warmup2000-b10.98-b-f2c1f5}{3} \\
-- & 0.9 & -- & -- & -- & -- & -- & -- & -- & -- & 3.548 & \href{https://wandb.ai/marin-community/optimizer-scaling/runs/sweep-130m-2B-mars2816c1lr0.016-wd0.1-minlr0-warmup2000-b10.9-b2-ea14d0}{4} \\
-- & 0.98 & -- & -- & -- & -- & -- & -- & -- & -- & 3.536 & \href{https://wandb.ai/marin-community/optimizer-scaling/runs/sweep-130m-2B-marse45e36lr0.016-wd0.1-minlr0-warmup2000-b10.9-b2-2de81b}{5} \\
-- & 0.99 & -- & -- & -- & -- & -- & -- & -- & -- & 3.562 & \href{https://wandb.ai/marin-community/optimizer-scaling/runs/sweep-130m-2B-mars2f8911lr0.016-wd0.1-minlr0-warmup2000-b10.9-b2-172044}{6} \\
\bottomrule
\end{tabular}
\end{table}

\begin{table}[H]
\centering
\caption{Hyperparameter ablation for Mars on 130m on 2x Chinchilla Data}
\label{tab:ablation_mars_130m_on_2x_chinchilla_data}
%
\end{table}

\begin{table}[H]
\centering
\caption{Hyperparameter ablation for Mars on 130m on 4x Chinchilla Data}
\label{tab:ablation_mars_130m_on_4x_chinchilla_data}
%
\end{table}

\begin{table}[H]
\centering
\caption{Hyperparameter ablation for Mars on 130m on 8x Chinchilla Data}
\label{tab:ablation_mars_130m_on_8x_chinchilla_data}
%
\end{table}

\subsection{Sweeping Results for NAdamW}%
\begin{table}[H]
\centering
\caption{Hyperparameter ablation for NAdamW on 130m on 1x Chinchilla Data}
\label{tab:ablation_nadamw_130m_on_1x_chinchilla_data}
%
\end{table}

\begin{table}[H]
\centering
\caption{Hyperparameter ablation for NAdamW on 130m on 2x Chinchilla Data}
\label{tab:ablation_nadamw_130m_on_2x_chinchilla_data}
%
\end{table}

\begin{table}[H]
\centering
\caption{Hyperparameter ablation for NAdamW on 130m on 4x Chinchilla Data}
\label{tab:ablation_nadamw_130m_on_4x_chinchilla_data}
%
\end{table}

\begin{table}[H]
\centering
\caption{Hyperparameter ablation for NAdamW on 130m on 8x Chinchilla Data}
\label{tab:ablation_nadamw_130m_on_8x_chinchilla_data}
%
\end{table}

\subsection{Sweeping Results for Adam-Mini}%
\begin{table}[H]
\centering
\caption{Hyperparameter ablation for Adam-Mini on 130m on 1x Chinchilla Data}
\label{tab:ablation_adam-mini_130m_on_1x_chinchilla_data}
%
\end{table}

\begin{table}[H]
\centering
\caption{Hyperparameter ablation for Adam-Mini on 130m on 2x Chinchilla Data}
\label{tab:ablation_adam-mini_130m_on_2x_chinchilla_data}
%
\end{table}

\begin{table}[H]
\centering
\caption{Hyperparameter ablation for Adam-Mini on 130m on 4x Chinchilla Data}
\label{tab:ablation_adam-mini_130m_on_4x_chinchilla_data}
%
\end{table}

\begin{table}[H]
\centering
\caption{Hyperparameter ablation for Adam-Mini on 130m on 8x Chinchilla Data}
\label{tab:ablation_adam-mini_130m_on_8x_chinchilla_data}
%
\end{table}

\subsection{Sweeping Results for Kron}%
\begin{table}[H]
\centering
\caption{Hyperparameter ablation for Kron on 130m on 1x Chinchilla Data}
\label{tab:ablation_kron_130m_on_1x_chinchilla_data}
%
\end{table}

\begin{table}[H]
\centering
\caption{Hyperparameter ablation for Kron on 130m on 2x Chinchilla Data}
\label{tab:ablation_kron_130m_on_2x_chinchilla_data}
%
\end{table}

\begin{table}[H]
\centering
\caption{Hyperparameter ablation for Kron on 130m on 4x Chinchilla Data}
\label{tab:ablation_kron_130m_on_4x_chinchilla_data}
%
\end{table}

\begin{table}[H]
\centering
\caption{Hyperparameter ablation for Kron on 130m on 8x Chinchilla Data}
\label{tab:ablation_kron_130m_on_8x_chinchilla_data}
%
\end{table}

\subsection{Sweeping Results for Soap}%
\begin{table}[H]
\centering
\caption{Hyperparameter ablation for Soap on 130m on 1x Chinchilla Data}
\label{tab:ablation_soap_130m_on_1x_chinchilla_data}
%
\end{table}

\begin{table}[H]
\centering
\caption{Hyperparameter ablation for Soap on 130m on 2x Chinchilla Data}
\label{tab:ablation_soap_130m_on_2x_chinchilla_data}
%
\end{table}

\begin{table}[H]
\centering
\caption{Hyperparameter ablation for Soap on 130m on 4x Chinchilla Data}
\label{tab:ablation_soap_130m_on_4x_chinchilla_data}
%
\end{table}

\begin{table}[H]
\centering
\caption{Hyperparameter ablation for Soap on 130m on 8x Chinchilla Data}
\label{tab:ablation_soap_130m_on_8x_chinchilla_data}
%
\end{table}

\subsection{Sweeping Results for Muon}%
\begin{table}[H]
\centering
\caption{Hyperparameter ablation for Muon on 130m on 1x Chinchilla Data}
\label{tab:ablation_muon_130m_on_1x_chinchilla_data}
%
\end{table}

\begin{table}[H]
\centering
\caption{Hyperparameter ablation for Muon on 130m on 2x Chinchilla Data}
\label{tab:ablation_muon_130m_on_2x_chinchilla_data}
%
\end{table}

\begin{table}[H]
\centering
\caption{Hyperparameter ablation for Muon on 130m on 4x Chinchilla Data}
\label{tab:ablation_muon_130m_on_4x_chinchilla_data}
%
\end{table}

\begin{table}[H]
\centering
\caption{Hyperparameter ablation for Muon on 130m on 8x Chinchilla Data}
\label{tab:ablation_muon_130m_on_8x_chinchilla_data}
%
\end{table}

\subsection{Sweeping Results for Scion}%
\begin{table}[H]
\centering
\caption{Hyperparameter ablation for Scion on 130m on 1x Chinchilla Data}
\label{tab:ablation_scion_130m_on_1x_chinchilla_data}
%
\end{table}

\begin{table}[H]
\centering
\caption{Hyperparameter ablation for Scion on 130m on 2x Chinchilla Data}
\label{tab:ablation_scion_130m_on_2x_chinchilla_data}
%
\end{table}

\begin{table}[H]
\centering
\caption{Hyperparameter ablation for Scion on 130m on 4x Chinchilla Data}
\label{tab:ablation_scion_130m_on_4x_chinchilla_data}
%
\end{table}

\begin{table}[H]
\centering
\caption{Hyperparameter ablation for Scion on 130m on 8x Chinchilla Data}
\label{tab:ablation_scion_130m_on_8x_chinchilla_data}
%
\end{table}

\subsection{Sweeping Results for Sophia}%
\begin{table}[H]
\centering
\caption{Hyperparameter ablation for Sophia on 130m on 1x Chinchilla Data}
\label{tab:ablation_sophia_130m_on_1x_chinchilla_data}
\begin{tabular}{cccccccccc}
\toprule
$\beta_1$ & $\beta_2$ & $\epsilon$ & $\gamma$ & $\eta$ & $\mathrm{BSZ}$ & $\mathrm{warmup}$ & $\lambda$ & Loss & Link \\
\midrule
0.95 & 0.9 & 1e-07 & 0.0125 & 0.004 & 128 & 4000 & 0 & 3.544 & \href{https://wandb.ai/marin-community/optimizer-scaling/runs/sweep-130m-2B-sophiag67fd60lr0.004-wd0-minlr0-warmup4000-b10.95--7bdb33}{0} \\
\midrule
0.8 & -- & -- & -- & -- & -- & -- & -- & 3.581 & \href{https://wandb.ai/marin-community/optimizer-scaling/runs/sweep-130m-2B-sophiagf9d750lr0.004-wd0-minlr0-warmup4000-b10.8-b-1d728f}{1} \\
0.9 & -- & -- & -- & -- & -- & -- & -- & 3.546 & \href{https://wandb.ai/marin-community/optimizer-scaling/runs/sweep-130m-2B-sophiau81957flr0.004-wd0-minlr0-warmup4000-b10.9-b-743aad}{2} \\
0.98 & -- & -- & -- & -- & -- & -- & -- & 3.624 & \href{https://wandb.ai/marin-community/optimizer-scaling/runs/sweep-130m-2B-sophiagcf99e7lr0.004-wd0-minlr0-warmup4000-b10.98--b5b74a}{3} \\
-- & 0.95 & -- & -- & -- & -- & -- & -- & 3.546 & \href{https://wandb.ai/marin-community/optimizer-scaling/runs/sweep-130m-2B-sophiaud75557lr0.004-wd0-minlr0-warmup4000-b10.95--9bc769}{1} \\
-- & 0.98 & -- & -- & -- & -- & -- & -- & 3.553 & \href{https://wandb.ai/marin-community/optimizer-scaling/runs/sweep-130m-2B-sophiag5970d9lr0.004-wd0-minlr0-warmup4000-b10.95--d5a961}{2} \\
-- & 0.99 & -- & -- & -- & -- & -- & -- & 3.563 & \href{https://wandb.ai/marin-community/optimizer-scaling/runs/sweep-130m-2B-sophiagceaccflr0.004-wd0-minlr0-warmup4000-b10.95--4fb41c}{3} \\
\bottomrule
\end{tabular}
\end{table}

\begin{table}[H]
\centering
\caption{Hyperparameter ablation for Sophia on 130m on 2x Chinchilla Data}
\label{tab:ablation_sophia_130m_on_2x_chinchilla_data}
%
\end{table}

\begin{table}[H]
\centering
\caption{Hyperparameter ablation for Sophia on 130m on 4x Chinchilla Data}
\label{tab:ablation_sophia_130m_on_4x_chinchilla_data}
%
\end{table}

\begin{table}[H]
\centering
\caption{Hyperparameter ablation for Sophia on 130m on 8x Chinchilla Data}
\label{tab:ablation_sophia_130m_on_8x_chinchilla_data}
%
\end{table}

\section{Hyperparameter Ablation in Phase II}
\label{app:detailed_experiments_2}

We reported the results for the optimizers we swept in Phase II. The result is formulated in the same way as in Phase I.

\subsection{Sweeping Results for AdamW}%
\begin{table}[H]
\centering
\caption{Hyperparameter ablation for AdamW on 300m on 2x Chinchilla Data}
\label{tab:ablation_adamw_300m_on_2x_chinchilla_data}
%
\end{table}

\subsection{Sweeping Results for Cautious}%
\begin{table}[H]
\centering
\caption{Hyperparameter ablation for Cautious on 300m on 2x Chinchilla Data}
\label{tab:ablation_cautious_300m_on_2x_chinchilla_data}
%
\end{table}

\subsection{Sweeping Results for Lion}%
\begin{table}[H]
\centering
\caption{Hyperparameter ablation for Lion on 300m on 2x Chinchilla Data}
\label{tab:ablation_lion_300m_on_2x_chinchilla_data}
%
\end{table}

\subsection{Sweeping Results for Mars}%
\begin{table}[H]
\centering
\caption{Hyperparameter ablation for Mars on 300m on 2x Chinchilla Data}
\label{tab:ablation_mars_300m_on_2x_chinchilla_data}
%
\end{table}

\subsection{Sweeping Results for NAdamW}%
\begin{table}[H]
\centering
\caption{Hyperparameter ablation for NAdamW on 300m on 2x Chinchilla Data}
\label{tab:ablation_nadamw_300m_on_2x_chinchilla_data}
%
\end{table}

\subsection{Sweeping Results for Adam-Mini}%
\begin{table}[H]
\centering
\caption{Hyperparameter ablation for Adam-Mini on 300m on 2x Chinchilla Data}
\label{tab:ablation_adam-mini_300m_on_2x_chinchilla_data}
%
\end{table}

\subsection{Sweeping Results for Muon}%
\begin{table}[H]
\centering
\caption{Hyperparameter ablation for Muon on 300m on 2x Chinchilla Data}
\label{tab:ablation_muon_300m_on_2x_chinchilla_data}
%
\end{table}

\subsection{Sweeping Results for Scion}%
\begin{table}[H]
\centering
\caption{Hyperparameter ablation for Scion on 300m on 2x Chinchilla Data}
\label{tab:ablation_scion_300m_on_2x_chinchilla_data}
%
\end{table}

\subsection{Sweeping Results for Kron}%
\begin{table}[H]
\centering
\caption{Hyperparameter ablation for Kron on 300m on 2x Chinchilla Data}
\label{tab:ablation_kron_300m_on_2x_chinchilla_data}
%
\end{table}

\subsection{Sweeping Results for Soap}%
\begin{table}[H]
\centering
\caption{Hyperparameter ablation for Soap on 300m on 2x Chinchilla Data}
\label{tab:ablation_soap_300m_on_2x_chinchilla_data}
%
\end{table}

\section{Hyperparameter Ablation in Phase III 1.2B experiments}
\label{app:detailed_experiments_3}

We reported the results for the optimizers we swept in Phase III. The result is formulated in the same way as in Phase I.

\subsection{Sweeping Results for AdamW}%
\begin{table}[H]
\centering
\caption{Hyperparameter ablation for AdamW on 1.2b on 1x Chinchilla Data}
\label{tab:ablation_adamw_1.2b_on_1x_chinchilla_data}
%
\end{table}

\begin{table}[H]
\centering
\caption{Hyperparameter ablation for AdamW on 1.2b on 2x Chinchilla Data}
\label{tab:ablation_adamw_1.2b_on_2x_chinchilla_data}
%
\end{table}

\begin{table}[H]
\centering
\caption{Hyperparameter ablation for AdamW on 1.2b on 4x Chinchilla Data}
\label{tab:ablation_adamw_1.2b_on_4x_chinchilla_data}
%
\end{table}

\begin{table}[H]
\centering
\caption{Hyperparameter ablation for AdamW on 1.2b on 8x Chinchilla Data}
\label{tab:ablation_adamw_1.2b_on_8x_chinchilla_data}
%
\end{table}

\subsection{Sweeping Results for NAdamW}%
\begin{table}[H]
\centering
\caption{Hyperparameter ablation for NAdamW on 1.2b on 1x Chinchilla Data}
\label{tab:ablation_nadamw_1.2b_on_1x_chinchilla_data}
%
\end{table}

\begin{table}[H]
\centering
\caption{Hyperparameter ablation for NAdamW on 1.2b on 2x Chinchilla Data}
\label{tab:ablation_nadamw_1.2b_on_2x_chinchilla_data}
%
\end{table}

\begin{table}[H]
\centering
\caption{Hyperparameter ablation for NAdamW on 1.2b on 4x Chinchilla Data}
\label{tab:ablation_nadamw_1.2b_on_4x_chinchilla_data}
%
\end{table}

\begin{table}[H]
\centering
\caption{Hyperparameter ablation for NAdamW on 1.2b on 8x Chinchilla Data}
\label{tab:ablation_nadamw_1.2b_on_8x_chinchilla_data}
%
\end{table}

\subsection{Sweeping Results for Soap}%
\begin{table}[H]
\centering
\caption{Hyperparameter ablation for Soap on 1.2b on 1x Chinchilla Data}
\label{tab:ablation_soap_1.2b_on_1x_chinchilla_data}
%
\end{table}

\begin{table}[H]
\centering
\caption{Hyperparameter ablation for Soap on 1.2b on 2x Chinchilla Data}
\label{tab:ablation_soap_1.2b_on_2x_chinchilla_data}
%
\end{table}

\begin{table}[H]
\centering
\caption{Hyperparameter ablation for Soap on 1.2b on 4x Chinchilla Data}
\label{tab:ablation_soap_1.2b_on_4x_chinchilla_data}
%
\end{table}

\begin{table}[H]
\centering
\caption{Hyperparameter ablation for Soap on 1.2b on 8x Chinchilla Data}
\label{tab:ablation_soap_1.2b_on_8x_chinchilla_data}
%
\end{table}

\subsection{Sweeping Results for Muon}%
\begin{table}[H]
\centering
\caption{Hyperparameter ablation for Muon on 1.2b on 1x Chinchilla Data}
\label{tab:ablation_muon_1.2b_on_1x_chinchilla_data}
%
\end{table}

\begin{table}[H]
\centering
\caption{Hyperparameter ablation for Muon on 1.2b on 2x Chinchilla Data}
\label{tab:ablation_muon_1.2b_on_2x_chinchilla_data}
%
\end{table}

\begin{table}[H]
\centering
\caption{Hyperparameter ablation for Muon on 1.2b on 4x Chinchilla Data}
\label{tab:ablation_muon_1.2b_on_4x_chinchilla_data}
%
\end{table}

\begin{table}[H]
\centering
\caption{Hyperparameter ablation for Muon on 1.2b on 8x Chinchilla Data}
\label{tab:ablation_muon_1.2b_on_8x_chinchilla_data}
%
\end{table}

\section{Hyperparameter Ablation in Phase III 16x Chinchilla experiments}
\label{app:detailed_experiments_4}

\subsection{Sweeping Results for AdamW}%
\begin{table}[H]
\centering
\caption{Hyperparameter ablation for AdamW on 130m on 16x Chinchilla Data}
\label{tab:ablation_adamw_130m_on_16x_chinchilla_data}
%
\end{table}

\subsection{Sweeping Results for NAdamW}%
\begin{table}[H]
\centering
\caption{Hyperparameter ablation for NAdamW on 130m on 16x Chinchilla Data}
\label{tab:ablation_nadamw_130m_on_16x_chinchilla_data}
%
\end{table}

\subsection{Sweeping Results for Soap}%
\begin{table}[H]
\centering
\caption{Hyperparameter ablation for Soap on 130m on 16x Chinchilla Data}
\label{tab:ablation_soap_130m_on_16x_chinchilla_data}
%
\end{table}

\subsection{Sweeping Results for Muon}%
\begin{table}[H]
\centering
\caption{Hyperparameter ablation for Muon on 130m on 16x Chinchilla Data}
\label{tab:ablation_muon_130m_on_16x_chinchilla_data}
%
\end{table}

\section{Comparison with Prior Work}
\label{sec:comparison-with-prior}

This paper benchmarks the performance of 11 proposed optimizers and show vastly different speed-up ratio than prior works reported. In this section, we will compare the setup of our experiments with prior works with the hope of understanding the difference.

\begin{enumerate}[leftmargin=*]
    \item Sophia~\cite{liu2024sophiascalablestochasticsecondorder} (2$\times$) This work utilizes a small peak learning rate of learning rate smaller than 6e-4 (similar to the one shown in~\Cref{fig:motivation} Top Left). The reason for the small peak learning rate is likely 2-fold: (i) the authors are training on a pretraining dataset PILE with lower quality compared to the current pretraining dataset and (ii) in the implementation of Levanter that the authors used, the data shuffling is not completely random and instead is correlated on every compute node. Upon reproducing the results, we note that this difference can significantly impact the stability of the training process and a complete random shuffling is crucial for the usage of a large learning rate.
    \item  MARS~\cite{yuan2025mars} (2$\times$) This papers considers a similar setup as Sophia and uses a similar AdamW baseline. We note that in the first version of the paper, the authors also reported that increasing the learning rate of AdamW to 3e-3 can significantly improve the performance of AdamW (see Figure 6 of~\cite{yuan2025mars} arxiv version 1).
    \item  Soap~\cite{vyas2025soapimprovingstabilizingshampoo} (1.4$\times$) The actual speedup of Soap on 300M and 520M models are 1.2 to 1.3$\times$, which is only slightly lower than the claimed 1.4$\times$ speedup. We note that our implementation of Soap is slightly different from the one used in the paper that we performs blocking of weight in order to reduce the memory footprint and further uses bfloat16 for the momentum in the 1.2B experiments. Both modifications may lead to slightly lower step-wise performance.
    \item Muon~\cite{jordan2024muon,liu2025muonscalablellmtraining} (2$\times$) The speedup of Muon reported in different works are vastly different. In the original Nanogpt speedrun, Muon achieves 1.3$\times$ speedup over AdamW. Later the reproduction of Kimi reported a much higher speedup of 2$\times$. We note that the Kimi version utilizes a notably low learning rate for AdamW (8e-4 to 9e-4 for model between 400M to 1.5B) in the scaling experiments. We also note that the smaller learning rate is important for hyperparameter transfer from AdamW when roughly matching the update norm and AdamW can perform better higher learning rate in our experiments.
    Further, their comparison of Muon and AdamW on the MoE experiments compare two models with not fully decayed learning rate and this may significantly favors Muon, as shown in~\Cref{fig:necessity}.
    It is later shown independently in the work of Essential AI~\cite{ai2025practicalefficiencymuonpretraining} that Muon's token efficiency compared to AdamW is only 1.1 to 1.2$\times$.
    \item Cautious~\cite{liang2025cautious}, Block-wise Learning Rate Adam~\cite{wang2025sharpnessdisparityprincipletransformers}, FOCUS~\cite{liu2025focusorderconcentratedupdating} report 2$\times$ speedup over AdamW. These papers use a similar baseline as Sophia and MARS.
    \item SWAN~\cite{ma2025swansgdnormalizationwhitening} and DION~\cite{ahn2025diondistributedorthonormalizedupdates} report 2-3$\times$ speedup over AdamW. The comparison between these two optimizers and AdamW is carried out on a smaller than 1$\times$ Chinchilla regime, where the speed-up of matrix-based optimizer may be larger. We also note that we didn't consider the communication cost, which is the main focus of DION.
    \item SPlus~\cite{frans2025stablewhiteningoptimizerefficient} reports a 2$\times$ speedup over AdamW. This work considers an atypical setup where the model is trained with a constant learning rate.
\end{enumerate}

\end{document}